%% file: main.tex
\pdfoutput=1
\documentclass[11pt]{article}

\usepackage[final]{coling}

\usepackage{times}
\usepackage{latexsym}

\usepackage[T1]{fontenc}
\usepackage[utf8]{inputenc}

\usepackage{microtype}

\usepackage{inconsolata}

\usepackage{graphicx}
\usepackage{booktabs}
\usepackage{soul}
\usepackage{color}
\usepackage{pifont}
\usepackage{multirow}
\usepackage{multicol}
\usepackage{enumitem,kantlipsum}
\usepackage[normalem]{ulem}
\usepackage{color,colortbl}
\usepackage{todonotes} 
\usepackage{soul}
\usepackage{threeparttable, makecell}
\usepackage{makecell}
\usepackage{xcolor}
\usepackage{array}
\usepackage{tcolorbox}

\usepackage{comment}
\usepackage{caption}
\usepackage{subcaption}
\usepackage{listings} % Add this for using the listings package

\usepackage{arabtex}
\usepackage{utf8}
\setcode{utf8}
\usepackage{tikz}  

\definecolor{darksky}{RGB}{33, 150, 243} % Adjust RGB values for custom colors

\usepackage{url}
\usepackage{soul}
\usepackage{xcolor} % Required for custom colors
\usepackage{todonotes}
\definecolor{lightskyblue}{rgb}{0.53, 0.81, 0.98}
\usepackage{graphicx}
\definecolor{blue}{rgb}{0,0, 0.6}
\definecolor{dkgreen}{rgb}{0,0.6,0}
\definecolor{gray}{rgb}{0.5,0.5,0.5}
\definecolor{mauve}{rgb}{0.58,0,0.82}
\definecolor{mauve}{rgb}{0,0,0}
\definecolor{black}{rgb}{0,0,0}
\definecolor{tri}{rgb}{.25,.88,.82}
\definecolor{lilac}{rgb}{0.85,0.64,0.85}
\definecolor{lightblue}{rgb}{0.53, 0.81, 0.98}

\newcommand{\ds }{\emph{AraDiCE}}
\newcommand{\dsc }{\emph{AraDiCE-Culture}}
\definecolor{darkred}{rgb}{0.6, 0.2, 0.2}
\definecolor{darksky}{rgb}{0.1, 0.3, 0.8}
\title{\textbf{\textsc{\textcolor{darksky}{AraDiCE}}}: Benchmarks for Dialectal and Cultural Capabilities in LLMs}
\author{
    Basel Mousi,
    Nadir Durrani,
    Fatema Ahmad,
    Md. Arid Hasan,\textsuperscript{${\dagger}$} 
    Maram Hasanain,\\
    {\bf        
        Tameem Kabbani,\textsuperscript{${\ast}$}    
        Fahim Dalvi,
        Shammur Absar Chowdhury,
        Firoj Alam
    }\\
    Qatar Computing Research Institute, Qatar, 
    \textsuperscript{$^{\dagger}$}University of New Brunswick, Canada \\
    \textsuperscript{$^{\ast}$}American University of Sharjah, UAE \\
    \{bmousi, ndurrani, fialam\}@hbku.edu.qa \\
}
\begin{document}
\maketitle
\begin{abstract}
%\todo[inline]{TO DO}
% \begin{list}
% We benchmarked four existing Arabic datasets for understanding and generation tasks, seven datasets translated into MSA and dialects (Levantine and Egyptian) and post-edited, as well as an in-house developed cultural dataset. 

Arabic, with its rich diversity of dialects, remains significantly underrepresented in Large Language Models, particularly in dialectal variations. We address this gap by introducing seven synthetic datasets in dialects alongside Modern Standard Arabic (MSA), created using Machine Translation (MT) combined with human post-editing. We present \textbf{AraDiCE}, a benchmark for \textbf{Ara}bic \textbf{Di}alect and \textbf{C}ultural \textbf{E}valuation.
%for evaluating 
We evaluate LLMs on dialect comprehension and generation, focusing specifically on low-resource Arabic dialects. Additionally, we introduce the first-ever fine-grained benchmark designed to evaluate cultural awareness across the Gulf, Egypt, and Levant regions, providing a novel dimension to LLM evaluation. Our findings demonstrate that while Arabic-specific models like Jais and AceGPT outperform multilingual models on dialectal tasks, significant challenges persist in dialect identification, generation, and translation. This work contributes $\approx$45K post-edited samples, a cultural benchmark, and highlights the importance of tailored training to improve LLM performance in capturing the nuances of diverse Arabic dialects and cultural contexts. We have released the dialectal translation models and benchmarks developed %curated 
in this study.\footnote{\url{https://huggingface.co/datasets/QCRI/AraDiCE}} 

\end{list}
\end{abstract}

\input{sections/intro}
\input{sections/dataset}

\input{sections/translation}
\input{sections/experiments}
\input{sections/results}

\input{sections/related_work}
\input{sections/conclusions}

\section{Limitations}
\label{sec:limitations}

\begin{itemize}[noitemsep,topsep=0pt,leftmargin=*,labelsep=.5em]
    \item Post-editing machine translation outputs is a tedious process, and the translator's choice of words can significantly impact translation quality. We provided clear instructions, conducted thorough training, and reviewed random samples to offer feedback. However, due to the size and diversity of the datasets, we could not review all datasets comprehensively.

    \item Another issue to highlight is that most datasets, except ArabicMMLU, are adapted from English and thus are influenced by Western culture. While we made efforts in annotation guidelines and post-editing to address these cultural biases, the subjective nature of sensitivity means that some samples may still be considered sensitive by different individuals or communities.

    \item While our study primarily focuses on Arabic dialects, it is limited in its coverage of the diverse dialects spoken across the Arab region. We mainly addressed Levantine, Gulf and Egyptian Arabic, but left out dialects such as Maghrebi and Sudanese. Future work should aim to fill these gaps by expanding coverage to a broader range of dialects, providing a more comprehensive evaluation of Arabic language models.

    \item Due to resource limitations, we only evaluated our benchmarks using models up to 13B parameters. As a one-off experiment, we tested the ArabicMMLU task with a larger  Jais 30B,\footnote{\url{https://huggingface.co/inceptionai/jais-30b-chat-v1}} model. We found (See results in Appendix \ref{sec:app_world_knowledge}, Figure \ref{fig:avg-scores-arabic-mmlu-ar-jais30B}) Jais 30B to perform similarly to Jais 13B indicating that larger models do not necessarily show significant improvements in this case. Due to hardware limitations, we could not run Jais 70B models, but it would be interesting to compare the higher-scale Jais and Llama models to see if their increased scale can compensate for the lack of dialect-specific training.

    % \item Due to resource limitations, we only evaluated our benchmarks using models up to 13B parameters. As a one-off experiment, we tested the ArabicMMLU task with larger models, specifically Llama 3 70B\footnote{\url{https://huggingface.co/meta-llama/Meta-Llama-3-70B}} and Jais 30B,\footnote{\url{https://huggingface.co/inceptionai/jais-30b-chat-v1}. We were unable to run Jais 70B model.} as shown Appendix \ref{sec:app_world_knowledge}, Figure \ref{fig:avg-scores-arabic-mmlu-ar-big-models}. These results suggest that while Arabic-centric models like JAIS and AceGPT outperform general models like Llama and Mistral of similar size, the performance of Llama 3 70B surpasses that of our Arabic-centric models. This indicates that larger, more general models have the potential to achieve superior results. If adapted specifically for Arabic, these larger models could potentially perform even better, highlighting the need for further fine-tuning and adaptation to fully leverage their capabilities for Arabic dialects. 

\end{itemize}

\section*{Ethical Consideration}
\label{sec:ethics}
We do not anticipate any ethical issues in this study. We extended publicly available datasets through the PEMT process. Additionally, our 
% in-house developed 
culturally aware \dsc{} dataset, consists of question-answer pairs based on commonsense knowledge and does not include information related to individual or organizational identities. Therefore, we do not foresee any potential risks arising from the outcomes of our study.

%\section*{Acknowledgments}
%The contributions of M. Hasanain were funded by the NPRP grant 14C-0916-210015, which is provided by the Qatar National Research Fund (a member of Qatar Foundation).

\bibliography{bib/bibliography}

\clearpage
\newpage
\section*{Appendix}
\label{sec:appendix}
\appendix
\input{sections/appendices}
\end{document}

%% file: sections/intro.tex
\section{Introduction}
\label{sec:introduction}

Large Language Models (LLMs) have consistently pushed the boundaries of Natural Language Processing (NLP), achieving state-of-the-art performance on a wide range of tasks. These models have excelled in areas such as machine translation, summarization, sentiment analysis, and even more complex applications like legal document analysis and creative writing \cite{openai2023gpt4,touvron2023llama,bubeck2023sparks}. Their remarkable ability to extract, reason, and generalize knowledge is fueled by training on vast amounts of data covering diverse topics and domains.

\begin{figure}
    \centering
    \includegraphics[width=0.95\linewidth]{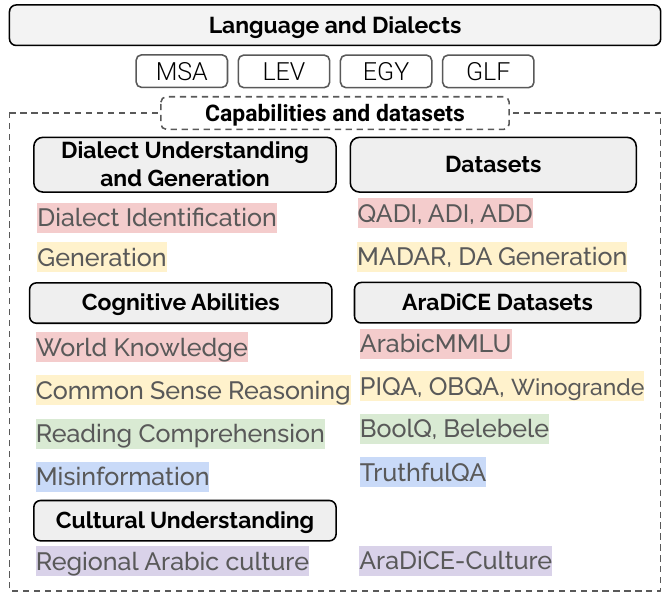}
    \caption{Capabilities and associated datasets for benchmarking, evaluated on different dialects.
    %\firoj{Com: Figures that give an overview of something are appreciated, but I'd suggest to think about Figure 1 a bit more. Structurally, it's a table, and it took me a while to get that.}
    % \firoj{feel free to improve it}    %https://docs.google.com/presentation/d/1sOkFSHN6GtzDI0WN27DSulPO_zvBtzFpZpufCFfQIqw/edit?usp=sharing
    }
    \label{fig:benchmarking_tasks_datasets}
    \vspace{-0.35cm}
\end{figure}
However, the success of these models is heavily skewed towards languages with abundant resources, such as English \cite{bang2023multitask, ahuja2023mega}. Low-resource languages, including Arabic and its various dialects, are significantly underrepresented in the datasets used to train these models. This disparity poses a substantial challenge, as LLMs require extensive and diverse data to perform effectively. Consequently, speakers of low-resource languages are at a disadvantage, unable to fully benefit from the advancements in NLP technologies. Recent efforts have been made to train Arabic LLMs \cite{sengupta2023jais, bari2024allamlargelanguagemodels,fanar2024} and to adapt multilingual models for Arabic \cite{touvron2023llama,huang2024acegpt}, but these models are predominantly tailored to MSA, leaving them less effective in handling dialectal Arabic (DA). To formalize this observation, we conduct a systematic study to benchmark Arabic and multilingual models in their performance on \textbf{Dialectal Arabic}. More specifically, we aim to answer the following questions:

\begin{itemize}  [noitemsep,topsep=0pt,leftmargin=*,labelsep=.5em]
    \item Can LLMs effectively perform basic NLP tasks in dialects? \textbf{(Understanding and Generation)} 
    \item Can LLMs demonstrate reasoning, comprehension, and handle knowledge and misinformation in dialects? \textbf{(Cognitive Abilities)}
    \item Are they aware of Arabic cultural knowledge? \textbf{(Cultural Understanding)}
\end{itemize}

\noindent To this end, we compiled a comprehensive suite of both existing and newly developed benchmarks to assess the capabilities of these models. We primarily evaluate their fundamental NLP abilities in encoding dialects through dialect identification (e.g. \newcite{el-haj-etal-2018-arabic}) and dialectal machine translation (e.g. \newcite{abdul-mageed-etal-2023-nadi}) tasks. 
%s %(ADI\footnote{\url{https://arabicspeech.org/adi_resources/mgb3}} and QADI \cite{abdelali2020arabic}) and in generation through machine translation (). 
To assess broader cognitive abilities, %understanding capabilities, 
we include benchmarks that gauge \textit{World Knowledge} 
% (MMLU~\cite{hendrycks2020measuring}, 
(ArabicMMLU~\cite{koto2024arabicmmlu}),
\textit{Common Sense Reasoning} (PIQA~\cite{bisk2020piqa}, OBQA~\cite{mihaylov-etal-2018-suit}, Winogrande~\cite{sakaguchi2021winogrande}), \textit{Reading Comprehension}, BoolQ~\cite{clark2019boolq}, Belebele~\cite{bandarkar2023belebele}), and the ability to handle \textit{Misinformation} (TruthfulQA~\cite{lin-etal-2022-truthfulqa}). 

These benchmarks are primarily available in English, but researchers and practitioners have increasingly turned to synthetic data creation methods to address the gap in low-resource languages \cite{long2024llms}. One promising approach involves utilizing MT to generate synthetic datasets for these languages. This method harnesses the capabilities of existing translation models to create large-scale, high-quality synthetic data, which can then be used to train LLMs.

In this paper, we introduce a comprehensive approach
% framework 
that leverages MT, specifically from English to MSA and MSA to dialects, combined with human post-editing, to develop synthetic benchmarks for low-resource DA.
% dialectal Arabic. 
We concentrate on the Levantine (LEV), and Egyptian (EGY) dialects. 
%This is a pivotal step towards training dialect-aware Arabic LLMs. 
Our primary goal is to curate benchmarks that evaluate the performance of LLMs across underrepresented language and dialects. 

% In addition to assessing dialect comprehension and general cognitive abilities through NLU tasks, 
Moreover, we introduce a third and novel benchmark, \textbf{AraDiCE-Culture}, focused on cultural awareness across the Levantine, Egyptian, and Gulf regions. This benchmark evaluates whether LLMs grasp regional cultural nuances beyond language. Our dataset includes questions on public holidays, food, geography, history, public figures, traditional clothing, and more. 
We probe cultural specifics to assess if the LLMs can differentiate between these regions, emphasizing the importance of models to understand both dialects and their cultural\footnote{We follow the definition of culture from \citet{alkhamissi-etal-2024-investigating}, which describes it as ``a multi-faceted inquiry reflecting diversity across worldviews and belief systems.''} contexts.
% \firoj{main weakness of the paper is the lack of an understanding what "culture" actually is. }
%\textcolor{blue}{Culture is a complex construct and to define ``culture'' we adopt the definition of it as ``a multi-faceted inquiry reflecting diversity across worldviews and belief systems'' \cite{alkhamissi-etal-2024-investigating}.}
% By probing cultural specifics, we assess if LLMs can differentiate between these regions, highlighting the need for models to understand not just dialects, but the cultural contexts in which they are used.
% To address this gap, researchers and practitioners have increasingly turned to synthetic data creation methods. One promising approach involves utilizing Machine Translation (MT) to generate synthetic datasets for low-resource languages. This method leverages the capabilities of existing translation models to create large-scale, high-quality synthetic data, which can then be used to train LLMs.
% In this paper, we present a comprehensive framework for leveraging Machine Translation to create synthetic benchmarking with post-editing for low-resource dialectal Arabic as a step towards training dialect aware Arabic LLMs. Our approach focuses on creating high-quality, diverse datasets that can enhance the performance of LLMs in underrepresented languages and dialects. We provide an in-depth analysis of the challenges involved in this process and propose solutions to optimize the quality and utility of the synthetic data. 
Our contributions in this work include:
\begin{itemize}[noitemsep,topsep=0pt,leftmargin=*,labelsep=.5em]
    % \item We present an approach 
    % for translating and post-editing dialectal data to curate benchmarks.
    % \item We benchmark the capabilities of LLMs for cognitive abilities, and various understanding and generation tasks in both MSA and DA.
    \item We develop benchmarks through the translation and post-editing of dialectal data, and assess LLMs' performance in cognitive abilities, understanding, and generation tasks across MSA and Dialectal Arabic.
    %dialectal Arabic.
    % that encompass commonsense reasoning, world knowledge, and various understanding and generation tasks in MSA and dialectal Arabic. %, specifically for Egyptian and Levantine dialects.
    \item We create the first benchmark for region-wise cultural evaluation in the Gulf, Egyptian, and Levant regions. %Egypt, and Levantine regions.
    \item We present a comparative analysis of Arabic-focused LLMs, such as Jais \cite{sengupta2023jais} and AceGPT \cite{huang2024acegpt}, alongside state-of-the-art models Llama 3 \cite{touvron2023llama} and Mistral \cite{jiang2023mistral7b}.
    \item To our knowledge, this is the first effort to develop dialectal benchmarks for Arabic LLMs. %, providing approximately 48,000 post-edited samples to the community. 
    \item We have released the dialectal translation models and benchmarks curated in this study.
    % \item Translated and post-edited benchmark datasets consisting of various evaluation capabilities, including commonsense reasoning, world knowledge, natural language understanding (NLU), and tasks related to misinformation.
    % \item Assessment of open-source LLMs using both versions of the datasets.
    % \item A comparative analysis highlighting the pros and cons of automatically translated datasets for benchmarking. 
\end{itemize}
% \noindent To our knowledge, this is the first effort to develop dialectal benchmarks for Arabic LLMs. We will make these efforts publicly available, releasing the dialectal translation models and benchmarks curated in this study. 
A summary of our findings is as follows: 
\begin{description}[noitemsep,topsep=0pt,leftmargin=*,labelsep=.5em]
 \item [Understanding and Generation] LLMs generally struggle with dialect identification, generation, and translation tasks. Although Arabic-centric models perform better on dialectal tasks, their performance still lags compared to MSA or English. These models often rely on MSA knowledge for distinguishing between dialects. They are better at encoding and understanding dialects than at generating or translating them, as evidenced in dialect generation and translation tasks.
 % % % We observed that the 
 % % These models tend to fall back on MSA knowledge when distinguishing between the dialects. Moreover, 
 % % % we observed that the
 % These models often rely on MSA knowledge when distinguishing between dialects and are better at encoding and understanding dialects than at generating them.
 % LLMs are better at encoding and understanding dialects than the generation itself, as evident in dialect generation and translation tasks.
 \item [Cognitive Abilities] %Our study reflects that 
 % Arabic-centric LLMs are more equipped to handle dialects, in addition to MSA; whereas multilingual LLMs fail drastically indicating the current gap in linguistic diversity adaptation of these models (Llama3 and Mistral) in the scope of DA. %Arabic dialects. %Furthermore, these results underscores 
 Arabic-centric LLMs, such as Jais and AceGPT are better equipped to handle both MSA and its dialects. In contrast, multilingual LLMs, such as Llama3 and Mistral, show a substantial deficiency in adapting to Arabic dialects, highlighting a significant gap in their ability to handle linguistic diversity in this context.
 \item [Cultural Understanding] Arabic-centric models also demonstrate a superior understanding of cultural nuances compared to general multilingual LLMs. This highlights the need for a more specialized training regimen to effectively address regional linguistic and cultural variations.
 \item [Dialectal Variability] The effectiveness of Arabic-centric models varies notably across dialects, with stronger performance in generating Gulf responses, likely due to the dialect's closer resemblance to MSA. Our findings highlight the importance of \ds{} in assessing dialectal capabilities and identifying gaps in LLM models.
\end{description}

%% file: sections/dataset.tex
\section{Datasets}
\label{sec:datasets}

% First, we select a number of datasets that have been used for standard LLM benchmarking, including Arabic. The capabilities assessed on these datasets include \textit{understanding and generation}, \textit{world knowledge}, \textit{commonsense reasoning}, \textit{reading comprehension}, \textit{misinformation}, and \textit{cultural understanding}, as listed in Figure \ref{fig:benchmarking_tasks_datasets}. 
% 

We begin by selecting a range of datasets commonly used for standard LLM benchmarking, including specific to Arabic. These datasets evaluate capabilities such as \textit{understanding and generation}, \textit{world knowledge}, \textit{commonsense reasoning}, \textit{reading comprehension}, \textit{misinformation}, and \textit{cultural understanding}, as outlined in Figure \ref{fig:benchmarking_tasks_datasets}.

We selected the datasets with a focus on diversity and linguistic compatibility.\footnote{By linguistic compatibility, we mean native speakers may not use dialectal Arabic for queries related to programming and coding.} These datasets include tasks involving generation and multiple-choice questions. 
% Below we provide a brief description of the datasets. 
%As shown in Figure \ref{fig:benchmarking_tasks_datasets}, 
The datasets used in this study include: \textit{(i)} four existing Arabic datasets for understanding and generation:  \textit{Arabic Dialects Dataset (ADD)} \cite{el-haj-etal-2018-arabic}, \textit{ADI}~\cite{abdelali-etal-2024-larabench},
% \footnote{\url{https://arabicspeech.org/adi_resources/mgb5}} 
and \textit{QADI} \citep{abdelali-etal-2021-qadi}, along with a dialectal response generation dataset \cite{naous2022acmtallip}, and \textit{MADAR} \cite{bouamor-etal-2018-madar}; \textit{(ii)} seven datasets translated into MSA and dialects (Levantine and Egyptian),\footnote{Except for ArabicMMLU, which is already in MSA.} which include \textit{ArabicMMLU}~\cite{koto2024arabicmmlu} \textit{BoolQ}~\cite{clark2019boolq}, \textit{PIQA} \cite{bisk2020piqa}, \textit{OBQA}~\cite{mihaylov-etal-2018-suit}, \textit{Winogrande} \cite{sakaguchi2021winogrande}, \textit{Belebele} \cite{bandarkar2023belebele}, and \textit{TruthfulQA} \cite{lin-etal-2022-truthfulqa}; and \textit{(iii)} AraDiCE-Culture, an in-house developed regional Arabic cultural understanding dataset. A detailed description of each dataset can be found in Appendix \ref{sec:appendix:dataset}.
% Despite the success of LLMs in many applications, these giant models are often criticized for their lack of factual knowledge, reasoning capabilities among others, and well known for experiencing hallucinations. Validating the knowledge embedded in these pretrained models become crucial especially when such models are applied to high-stakes scenarios, such as medical diagnosis and legal judgments. Hence, we extend our extensive evaluation to the following capabilities, tasks and datasets, as also highlighted in Figure \ref{fig:benchmarking_tasks_datasets}: 

% \nd{Winogrande and TruthfulQA in Figure 1 has redlines underneath}

% figure preparing - 

\paragraph{Cultural Dataset:} We curated a dataset of 180 culturally specific questions by hiring native annotators from the Gulf, Egypt, and the Levant regions to generate seed questions centered on cultural and country-specific themes. After reviewing all submitted questions, we selected those with varying answers across regions. To verify these differences, we appended a country name to each question and used Google Search to examine the top 5 search results. Only questions with distinct answers across countries were retained, resulting in a final set of 30 unique questions. These questions were then translated into the  dialects of six countries within the targeted three regions (Gulf, Levant, and Egypt). %, creating a complete dataset of 180 questions 
The questions span various categories, including public holidays, food, geography, and religion. For gold-standard answers, we followed the NativeQA framework \cite{hasan2024nativqa}. Annotators reviewed both the questions and the top 5 %Google 
search results, combining web data with their cultural knowledge to provide the most accurate answers. Detailed guidelines and sample question-answer pairs can be found in Appendices \ref{sec-app:culture-understanding}, \ref{sec:appendix_anno_guideline_cultural_dataset} and Table~\ref{tab-appx-cultural-responses}.

% We curated a dataset of 180 culturally specific questions by hiring annotators native to Gulf, Egypt, and the Levantine region to generate seed questions based on cultural and country-specific themes. We then reviewed all the submitted questions, selecting only those where the answers varied across countries. To verify the differences, we concatenated each question with a country name and submitted it to Google, inspecting the top 5 search results. Only questions with distinct answers for each country were retained, resulting in a final set of 30 unique questions. These questions were then automatically translated into the target dialects (Gulf, Levantine, and Egyptian) using GPT-4, producing a full dataset of 180 questions. The questions cover a range of categories, including public holidays, food, geography, history, literature, and religion. To generate gold-standard answers, we followed the NativeQA framework  \cite{hasan2024nativqa}. Annotators were shown both the questions and the top 5 Google search results for each, combining web data with their own cultural knowledge to provide the most accurate answer. Detailed guidelines and sample question-answer pairs can be found in the appendix \ref{sec:appendix_anno_guideline_cultural_dataset}.

%% file: sections/translation.tex
%\section{MT and Post-Editing Machine Translation (PEMT)}
\section{Machine Translation and Post-Editing}
\label{sec:translation}

Machine Translation is crucial for developing resources for low-resource languages with multiple dialects, like Arabic, which has MSA for formal use and diverse dialects like Egyptian and Levantine for everyday communication \cite{DurraniAI14}. These dialects differ significantly from MSA and from each other, presenting unique challenges for NLP tasks.

For MSA, several studies have released datasets, including MMLU, HellaSwag, and Arc-Challenge by Okapi \cite{lai2023okapi}, as well as MMLU by AceGPT \cite{huang2024acegpt}. Due to inconsistencies in the number of samples between the original datasets, we refrained from adopting them directly. Instead, we translated these datasets into MSA using Google Translate. %\footnote{\url{https://pypi.org/project/deep-translator/}} 
Currently, no publicly available translation systems support direct translation between Arabic dialects or between English and specific Arabic dialects; most MT systems focus on MSA-English translation. To address this gap, we trained models to translate from MSA to various Arabic dialects. Translated datasets were manually post-edited for fluency and accuracy. Details of our dialectal MT and post-editing framework are provided below.

\subsection{Dialectal MT}
\label{sec:diaMT}

% Our work focuses on developing machine translation (MT) models specifically tailored for Arabic dialects. We have trained MT models to translate between MSA and two widely spoken dialects: Egyptian Arabic and Levantine Arabic. By translating MSA benchmarks, such as ArabicMMLU, into these dialects, our models create valuable resources for evaluating LLMs in their understanding of dialectal Arabic.

We develop MT models translating between MSA and two major dialects: Egyptian  and Levantine Arabic. By translating MSA benchmarks like ArabicMMLU, we create resources for evaluating LLMs' understanding of dialectal Arabic.

%\subsubsection{Data}

\paragraph{Data:}  For training our translation systems, we utilized several datasets, including the MADAR 
%5th orpus 
\cite{bouamor-etal-2018-madar}, UFAL 
% corpus 
\cite{krubinski-etal-2023-multi}, LDC
% corpus, \footnote{\url{https://catalog.ldc.upenn.edu/byproject}} DIA2MSA %dataset 
\cite{mubarak2018dial2msa}, ArzEn-MultiGenre 
% dataset 
\cite{10023181}, and the SADID 
% dataset 
\cite{abid-2020-sadid}. Please see Appendix \ref{sec:appendix:data} for more details.

%\subsubsection{Models}

\paragraph{Models:}  We fine-tuned two robust machine translation models: \textbf{AraT5} \cite{nagoudi-etal-2022-arat5} and \textbf{NLLB} \cite{nllbteam2022language}. We experimented with several variants of these models, with sizes ranging from 600M to 3.3B parameters. AraT5 is an Arabic language model based on the T5 (Text-to-Text Transfer Transformer) architecture. In our preliminary experiments, we found the NLLB 3.3B model to surpass AraT5 and its smaller variants, post fine-tuning with dialectal data.  We carried out ablation studies using different data mixtures on the NLLB 3.3B model. We shortlisted three systems per dialect using BLEU scores as the primary criterion (see Tables \ref{tab:lev-model-scores} and \ref{tab:egy-model-scores} in Appendix \ref{ssec:appendix_mt_results}) and conducted human evaluation to select the best system for each dialect. Please see Appendix \ref{sec:appendix:MT} for details on our exploration of Dialectal MT.

%We shortlisted three systems per dialect using BLEU scores as our primary criterion (See Tables \ref{tab:lev-model-scores}, \ref{tab:egy-model-scores} in Appendix \ref{ssec:appendix_mt_results}) and used human evaluation to select the best system for each dialect. Please see Appendix \ref{sec:appendix:MT} for details on our exploration of Dialectal MT.

\subsection{Post-Editing of MT (PEMT)}
%\subsection{PEMT}

The translated datasets were manually post-edited to ensure fluency and adequacy. 
% We focused post-editing efforts on the test sets., particularly for datasets like BoolQ and TruthfulQA. 
% We mostly post-edited the whole test/val sets (see Table Table \ref{tab:data_stat}) except for BoolQ and TruthfulQA. 
We post-edited the majority of the test sets (see Table \ref{tab:data_stat}), with the exception of BoolQ and TruthfulQA.
For these two datasets, we chose specific samples for post-editing based on the following criteria: \textit{(i)} shorter text length to minimize post-editing effort, \textit{(ii)} cultural and religious compatibility with the Arab region, and \textit{(iii)} linguistic compatibility with Arabic. We excluded language-specific samples, such as those asking for the origin of an English word. In Table \ref{tab:data_stat}, we provide statistics of the datasets along with their translation and post-editing status.

\begin{table}[]
\centering
\setlength{\tabcolsep}{2.5pt}
\scalebox{0.85}{
\begin{tabular}{@{}llccr@{}}
\toprule
\multicolumn{1}{c}{\textbf{Data}} & \multicolumn{1}{c}{\textbf{Set}} & \multicolumn{1}{c}{\textbf{Trans.}} & \multicolumn{1}{c}{\textbf{Edited}} & \multicolumn{1}{c}{\textbf{Size}} \\ \midrule
\multicolumn{5}{c}{\textbf{Understanding and Generation}} \\  \midrule
QADI & Test &\ding{55} & \ding{55} & 2,597 \\
ADI & Test & \ding{55} & \ding{55} & 550 \\
MADAR & Test & \ding{55} & \ding{55} & 37,550 \\
DA Response & Test & \ding{55} & \ding{55} & 1,000 \\ \midrule
\multicolumn{5}{c}{\textbf{Cognitive Abilities}} \\  \midrule
ArabicMMLU & Test & \ding{51} & \ding{51} & 14,459 \\
PIQA & Val & \ding{51} & \ding{51} & 1838 \\
OBQA & Test &\ding{51} & \ding{51} & 500 \\
Belebele & Test & \ding{55} & \ding{55} & 3,600 \\
Winogrande & Val & \ding{51} & \ding{51} & 1,267 \\
TruthfulQA & Test & \ding{51} & \ding{51} & 780 \\
BoolQ & Val &\ding{51} & \ding{51} & 892 \\
\midrule
\multicolumn{5}{c}{\textbf{Cultural Understanding}} \\  \midrule
\dsc{} & Test & - & - & 180 \\
\bottomrule
\end{tabular}
}
% \caption{Selected datasets, their status of translation, and post-editing. Trans.: Automatic translation. Edited: Post translation edits.}
\caption{Selected datasets with translation and editing status: Trans: Translation | Edited: Post-editing.}
\label{tab:data_stat}
\end{table}

%\subsubsection{Guidelines}

% The dataset was  translated from modern standard Arabic to the Levantine and Egyptian dialects. The task was to post correct translations so that they are  fluent (i.e., they reflect the nuances of how the dialect is spoken) and adequate (i.e., they maintain the semantic meaning of the input sentence). 

\paragraph{Guidelines} To assist the translators and maintain dataset integrity, we provided two sets of guidelines for each dataset (except Arabic MMLU). One set focused on post-editing translated samples in MSA, while the other targeted the same process for dialects. Each dataset's guidelines included: \textit{(i)} details on the task components (e.g., one question and four answers), \textit{(ii)} general instructions for correcting errors and improving fluency and adequacy, and \textit{(iii)} specific instructions for handling unique cases within the datasets. Detailed instructions for the guidelines are provided in the Appendix \ref{sec:app_post_editing_guideline}.

\paragraph{Team and Setup} The translation team comprised 32 native speakers fluent in Levantine, Egyptian, and MSA, with educational backgrounds ranging from Bachelor's to Master's degrees, and ages between 21 and 53. Many were professional translators. They received specific guidelines and training tailored to each dataset and dialect to ensure translation quality. To manage the post-editing workload efficiently, we assigned one translator per item, which helped minimize costs and time. A random sample of post-edited texts was reviewed by %an in-house 
the expert translators for quality assurance. The process was streamlined using an in-house annotation platform and 17 dedicated projects, resulting in the post-editing of $\approx$45K %44,749 
items. A third-party company handled the hiring and competitive compensation of translators.

While synthetic datasets, such as those generated through machine translation, offer valuable opportunities to expand data coverage, they can introduce inherent biases that are difficult to completely mitigate during post-editing. To address this, our approach emphasizes refining translations to ensure alignment with Arabic linguistic nuances and cultural relevance. For tasks requiring world knowledge, we opted for ArabicMMLU (originally in Modern Standard Arabic, or MSA) instead of MMLU, as it better represents the knowledge and contexts relevant to Arabic-speaking communities and regions. 
For the Winogrande dataset, we made specific adjustments to culturally sensitive instances, as detailed in Section \ref{sec:addresing-cultural-mismatch}.

%% file: sections/experiments.tex
\section{Experimental Setup}
\label{sec:experiments}

%\subsection{Models}
\paragraph{Models:} For the LLMs benchmarking experiments we used open models, such as Llama-3-8B-Instruct
% \footnote{\url{https://huggingface.co/meta-llama/Meta-Llama-3-8B-Instruct}} 
~\cite{touvron2023llama}, Mistral-7B-Instruct~\cite{jiang2023mistral7b},\footnote{\url{https://huggingface.co/mistralai/Mistral-7B-Instruct-v0.2}} AceGPT-13B-chat
% \footnote{\url{https://huggingface.co/FreedomIntelligence/AceGPT-v1.5-13B-Chat}}
~\cite{huang2024acegpt} and Jais-13b-chat
% \footnote{\url{https://huggingface.co/inceptionai/jais-13b-chat}} 
~\cite{sengupta2023jais}. A random model\footnote{\url{https://huggingface.co/HuggingFaceH4/tiny-random-LlamaForCausalLM}} was used as a baseline to evaluate the relative performance of these LLMs. We have chosen to use only open models to reduce the computational budget.
% As a baseline, we used a random model to measure and compare performance. The random baseline serves as a lower bound for performance, allowing us to evaluate the relative capabilities/performance of LLMs. 
%\firoj{@Basel, we need to mention parameters setups.} 

%\subsection{Prompt}
\paragraph{Prompt:} In our experiments, we employed zero-shot learning to observe performance differences across dialects. %, without exploring few-shot learning. 
Although few-shot learning is known to enhance performance, we chose not to include it to simplify the experiments and minimize computational costs. 
% but it was not our focus. 
We used prompts in English, MSA, and dialects depending on the task, and released both the prompts and configurations via lm-harness.

% In all our experiments, we used zero-shot learning. We did not experiment with few-shot learning, as our primary focus was to observe performance differences across dialects. Moreover, it is well-established that few-shot learning improves performance. For the prompt design, we used prompts in the native language (MSA and dialects). We will release the prompt and the configuration publicly available for the community. 

%\subsection{Evaluation Setup}
% \paragraph{Evaluation:} For the evaluation, we used the \textit{LM Evaluation Harness},\footnote{https://github.com/EleutherAI/lm-evaluation-harness} which supports evaluation for both generation and multiple-choice question tasks. It also supports several metrics. For our experiments, we used metrics that are commonly reported in existing studies for the respective tasks and datasets. For example, F1 is a standard metric reported in previous studies for dialect identification tasks~\cite{abdelali-etal-2024-larabench}, while normalized accuracy has been widely used in current LLM leaderboards for measuring cognitive abilities tasks.\footnote{\url{https://huggingface.co/docs/leaderboards/open_llm_leaderboard/}} For MT computed BLEU score. 
% \firoj{@Basel, please check this section}

\paragraph{Evaluation:} We used the \textit{LM Evaluation Harness}\footnote{\url{https://github.com/EleutherAI/lm-evaluation-harness}} for both generation and multiple-choice tasks, employing standard metrics for each task and dataset. We used F1 scores for dialect identification, normalized accuracy for cognitive tasks,
% \footnote{as used in LLM leaderboards \url{https://huggingface.co/docs/leaderboards/open_llm_leaderboard/}} 
and SacreBLEU \cite{post-2018-call} for machine translation.

% For dialect identification, we used F1 scores, while cognitive tasks were measured with normalized accuracy, as used in LLM leaderboards\footnote{\url{https://huggingface.co/docs/leaderboards/open_llm_leaderboard/}}. We used SacreBLEU for machine translation.

%% file: sections/results.tex
\begin{figure}
    \centering
    \includegraphics[width=0.75\linewidth]{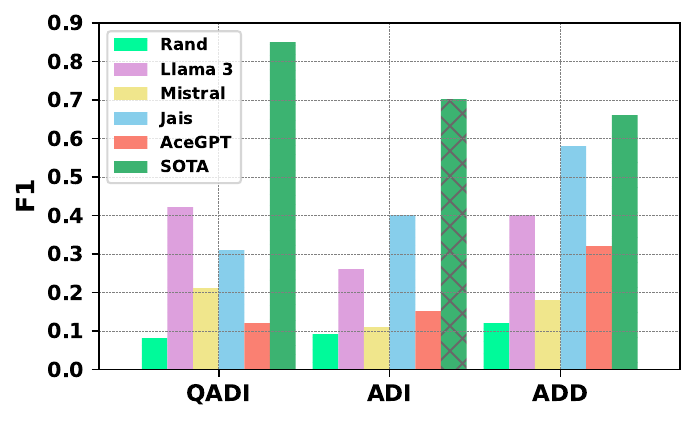}
    \vspace{-0.2cm}
    \caption{Comparison on \textit{dialect identification}}
    \label{fig:dialect-id-results}
    \vspace{-0.3cm}
\end{figure}

\begin{figure}
    \centering
    \includegraphics[width=0.75\linewidth]{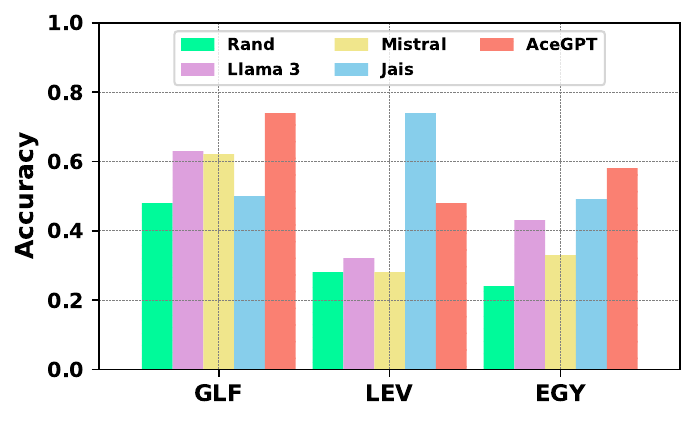}
    \vspace{-0.2cm}
    \caption{Results on \textit{dialect generation}
    % , formulated as MCQ based response selection task.    
    % Dialect selection accuracy.
    }
    \label{fig:dialect-selection-scores}
    \vspace{-0.3cm}
\end{figure}

\section{Results}
\label{sec:results}

\subsection{Understanding and Generation}
\label{sec:encoding-decoding}

% In our foremost experiments, we assess the models' ability to encode and generate dialectal Arabic. To evaluate these capabilities, we focus on tasks such as dialect identification, dialogue generation, and machine translation. 

% First, we 
We evaluate the models' ability to encode and generate dialectal Arabic, focusing on tasks like dialect identification, dialect generation, and MT. %machine translation.

\subsubsection{Dialect Identification}

%\nd{Basel, please change number in tables to bar chart. Add SOTA numbers instead of QCRI-Sys with citations. We should not give any hint where the paper is coming from.}
%\textcolor{red}{Basel: Moved the tables to the appendix and replaced them with bar graphs. Left the SOTA numbers as is. Need to site the QCRI system paper. Please let me know if I need to modify anything else}

% The results in Figure \ref{fig:dialect-id-results} reveal that all LLMs struggle with  distinguishing between dialects, especially when compared to state-of-the-art models specifically trained for this task. As a result, these models will likely rely heavily on their MSA knowledge to perform any given task. The performance of Llama 3, Jais, and others varies notably across different datasets. Llama 3 excels on the QADI dataset, whereas Jais outperforms Llama 3 on the ADI and ADD datasets. This variation can be attributed to the differences in the nature and distribution of the data, with QADI consisting of tweets, ADI comprising speech transcripts, and ADD containing Arabic commentaries. To better understand these discrepancies, we conducted an error analysis. Figure \ref{fig:qadi-confusion-matrices} displays the confusion matrices for the models on the QADI dataset. For instance, Llama 3, Mistral, and Jais often confuse Levantine with Gulf dialects, while AceGPT frequently mistakes MSA for Gulf dialect. A similar analysis was performed for all models across other datasets, with detailed predictions provided in Appendix \ref{sec:appendix:dialect-id}.

\begin{figure*}
     \centering
     \begin{subfigure}[b]{0.24\textwidth}
         \centering
         \includegraphics[width=\textwidth]{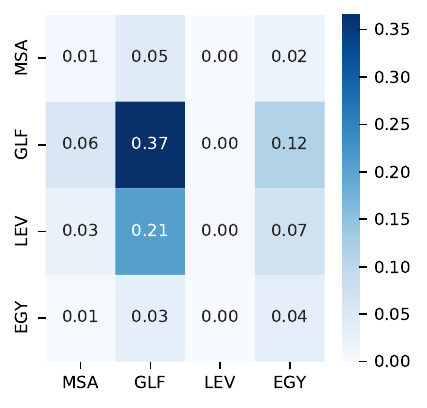}
         \caption{Llama 3}
         \label{fig:LLama3-qadi-confusion-matrix}
     \end{subfigure}
     \begin{subfigure}[b]{0.24\textwidth}
         \centering
         \includegraphics[width=\textwidth]{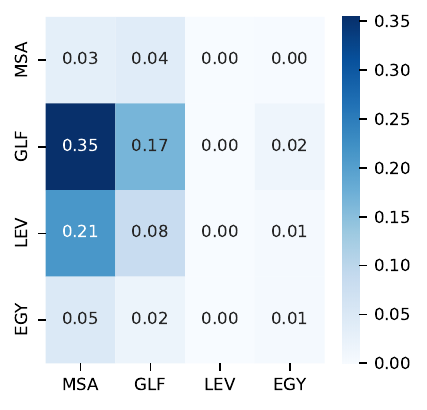}
         \caption{Mistral}
         \label{fig:mistral-qadi-confusion-matrix}
     \end{subfigure}
     \hfill
     \begin{subfigure}[b]{0.24\textwidth}
         \centering
         \includegraphics[width=\textwidth]{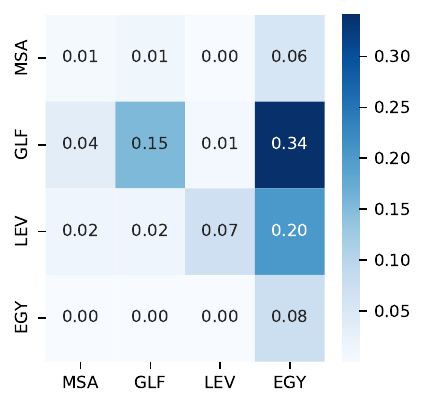}
         \caption{Jais}
         \label{fig:Jais-qadi-confusion-matrix}
     \end{subfigure}
     \hfill
    \begin{subfigure}[b]{0.24\textwidth}
         \centering
         \includegraphics[width=\textwidth]{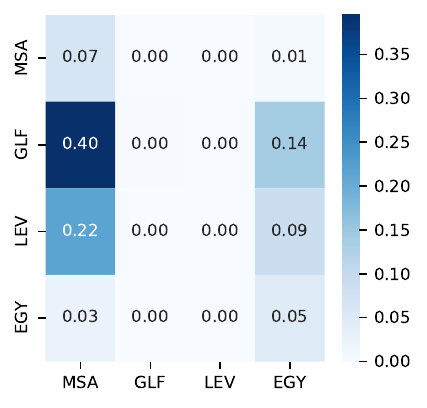}
         \caption{AceGPT}
         \label{fig:AceGPT-qadi-confusion-matrix}
     \end{subfigure}
     \vspace{-0.3cm}
     \caption{Confusion matrices on \textit{dialect identification} (QADI) dataset}
    \label{fig:qadi-confusion-matrices}
    \vspace{-0.2cm}
\end{figure*}

%The results in 
Figure \ref{fig:dialect-id-results} show that all LLMs struggle to distinguish between dialects, especially compared to SOTA models~\cite{hassan-etal-2021-asad}. Performance varies across datasets: Llama 3 excels on QADI, while Jais outperforms it on ADI and ADD,
%\footnote{ADD: Arabic Dialects Dataset} 
likely due to differences in the data—QADI uses tweets, ADI has speech transcripts, and ADD includes Arabic commentaries. Our error analysis (Figure \ref{fig:qadi-confusion-matrices}) shows Llama 3 confuse Gulf with Lev, Mistral confuses MSA with Lev/Gulf, Jais often confuse Egy with Gulf, while AceGPT mistakes MSA for Gulf. %This suggests 
The models tend to fall back on MSA knowledge when distinguishing between dialects. Please see Appendix \ref{sec:appendix:dialect-id} for a more comprehensive analysis.

% \subsection{Dialect Identification} 

% The F1 scores on the task of dialect identification are shown in table \ref{dialect-id-results}. There is a variation of model performance across different datasets. LLama 3 performs better than Jais on the QADI test set, while Jais beats LLama 3 on the ADI and ADD datasets. All models still lag behind commercial dialect identification systems. The variation in the performance of models can be attributed to the nature of data and its underlying distribution. The QADI dataset is comprised of tweets, ADI is comprised of speech transcripts and the ADD dataset contains Arabic commentaries. 

% We performed error analysis to investigate the mispredictions of models. Figure \ref{fig:qadi-confusion-matrices} shows the confusion matrices of the models considered on the QADI dataset. For instance LLama 3, Mistral and Jais tend to confuse the levantine with the gulf dialect while AceGPT confuses modern standard arabic with the gulf dialect. We conduct a similar analysis of all models on other datasets and we show some predictions.  Results are shown in appendix \ref{sec:appendix:dialect-id}

% \usepackage{booktabs}

\subsubsection{Dialect Generation} 

% Next, we assessed the models' ability to generate responses in dialectal Arabic. Initially, we employed a dialectal response generation task \cite{naous2022acmtallip}, where models were prompted in a specific dialect and asked to generate an appropriate response. However, our qualitative analysis revealed that the models often failed to follow the instruction, instead explaining the prompt rather than generating a dialectal response (see Appendix \ref{sec:appendix:dialogue-generation} for examples). To address this issue and to quantify the analysis, we simplified the task. We presented the models with multiple response options, only one of which was in the correct dialect, while the others were in different dialects. The models were then asked to select the best response. The accuracy scores for this \emph{MCQ response selection} task are shown in Table \ref{table:dialogue-selection-results}. Overall, the models tended to perform better on the Gulf dialect, with Arabic LLMs such as Jais and AceGPT outperforming Llama 3 and Mistral across all dialects. 

% Next, we 
We assessed the models' ability to generate responses in dialectal Arabic. Initially, we used a dialectal response generation task \cite{naous2022acmtallip}, where models were prompted in a specific dialect and asked to generate a response. However, qualitative analysis revealed that models often explained the prompt instead of generating a dialectal response (see Appendix \ref{sec:appendix:dialogue-generation} for examples). To address this, we simplified the task: models were given multiple response options, with only one in the correct dialect, and asked to select the best response. Accuracy scores for this \emph{MCQ response selection} task are shown in Figure \ref{fig:dialect-selection-scores}. Overall, models performed better on the Gulf dialect, with Arabic LLMs like Jais and AceGPT outperforming Llama 3 and Mistral across all dialects.

% \subsection{Dialogue Generation} 

% We aim to evaluate the ability of the models to generate dialectal dialogues. To achieve this, we use two evaluation methods. The first is \emph{Open Ended Dialogues Generation} in which we prompt the models to generate a response to a given utterance. The second is \emph{MCQ response selection} where the model is prompted to choose a response to a given dialectal utterance. The model is expected to choose the response in the same dialect of the input utterance. 

% We evaluated the \emph{Open Ended Dialogues Generation} qualitatively and we noticed that the models fail to follow the dialectal instructions. Models tend to describe the phrase instead of generating a response to it. Sample responses are shown in appendix \ref{sec:appendix:dialogue-generation}

% The accuracy scores for \emph{MCQ response selection} are shown in table \ref{table:dialogue-selection-results}. Generally, models tend to perform better on the gulf split of the data. Moreover, Jais and AceGPT perform better than Llama and Mistral on all splits. 

\begin{table*}[]
\centering
\resizebox{0.85\textwidth}{!}{%
\begin{tabular}{@{}lcccc|cccc|cccc@{}}
\toprule
 & \multicolumn{4}{c|}{\textbf{Dialects-to-English}} & \multicolumn{4}{c|}{\textbf{English-to-Dialects}} & \multicolumn{4}{c}{\textbf{MSA-to-Dialects}} \\ 
\midrule
\textbf{Dialect} & \textbf{Llama 3} & \textbf{Mistral} & \textbf{Jais} & \textbf{AceGPT} & \textbf{Llama 3} & \textbf{Mistral} & \textbf{Jais} & \textbf{AceGPT} & \textbf{Llama 3} & \textbf{Mistral} & \textbf{Jais} & \textbf{AceGPT} \\ 
\midrule
\textbf{Gulf} & 25.7 & 15 & 36.1 & 37.8 & 2.4 & 0.5 & 1.0 & 1.5 & 2.2 & 0.7 & 1.6 & 3.1 \\ 
\textbf{Levantine} & 23.0 & 12.2 & 36.1 & 35.3 & 1.4 & 0.2 & 0.9 & 2.1 & 1.3 & 0.3 & 1.3 & 3.0 \\ 
\textbf{Egyptian} & 26.9 & 13.9 & 40.2 & 39.8 & 2.0 & 0.4 & 1.2 & 3.8 & 1.5 & 0.3 & 2.1 & 3.8 \\ 
\bottomrule
\end{tabular}%
}
\vspace{-0.3cm}
\caption{Average BLEU scores across three translation tasks using MADAR test sets% Mist.: Mistral, Ace: AceGPT.
}
\vspace{-0.3cm}
\label{tab:combined-bleu}
\end{table*}

\subsubsection{Machine Translation} 

To assess the models' capacity to encode and generate dialectal knowledge, we use Dialect-to-{English, MSA} and {English, MSA}-to-Dialect tasks. The former evaluates the ability to encode dialects, while the latter assesses generation capabilities. Results averaged across multi-dialectal MADAR test sets are shown in Table \ref{tab:combined-bleu}. Several observations emerge:

% Another way to assess the models' capacity to encode and generate dialectal knowledge is through machine translation. For this purpose, we utilize Dialect--to--\{English, MSA\} and \{English, MSA\}--to--Dialect tasks. The former evaluates the model's ability to encode dialects, while the latter gauges its ability to generate them. Table \ref{tab:combined-bleu} shows the results averaged across MADAR test-sets in different dialects. When comparing the task of translating from dialects (Dialect-to-English) to translating into dialects (English-to-Dialects and MSA-to-Dialects), several observations emerge: 

% \paragraph{Models can understand and encode dialects but struggle with generating dialects:}
% Dialect-to-English translation scores are significantly higher across all models and dialects compared to translating into dialects. 
% This indicates that the models are more proficient at understanding and encoding dialects than generating them. The models benefit from the similarity to MSA when encoding knowledge, but they struggle with generating dialectal text, likely due to data sparsity. This highlights a potential limitation in their ability to produce fluent and accurate dialectal language.

%\paragraph{Models can understand and encode dialects but struggle with generating them:} 

Dialect-to-English translation scores are significantly higher across all models and dialects compared to translations into dialects. This suggests that \textbf{models excel at understanding and encoding dialects more than generating them.} While the models benefit from their similarity to MSA in encoding, they struggle with generating dialectal text due to differing vocabulary and data sparsity. This highlights limitations in producing fluent and accurate dialectal language.
% \paragraph{Comparing Models:} AceGPT consistently achieves higher BLEU scores across all dialects, showcasing its superior performance in translating from dialects. Jais, being an Arabic-centric model, also performs strongly compared to Llama 3 and Mistral, which demonstrate comparatively lower scores. All models exhibit significantly lower BLEU scores when translating into dialects, making it challenging to compare their performance reliably in this direction.
%\paragraph{Comparing Models:} 

% \textbf {Comparing Models:} 
AceGPT consistently scores higher in BLEU across all dialects, showing superior translation performance. Jais also performs well compared to Llama 3 and Mistral, which score lower. However, all models show significantly reduced BLEU scores for translating into dialects, complicating reliable performance comparisons in this direction.

\subsection{Cognitive Abilities}
\label{sec:cognitive-abilities}

% Now, we shift our attention to a comprehensive set of tasks designed to evaluate the knowledge and reasoning capabilities of LLMs, particularly in the context of dialectal Arabic. These tasks span various dimensions, including \textit{world knowledge}, \textit{reading comprehension}, \textit{commonsense reasoning} and misinformation, providing a holistic assessment of how well these models perform across dialect-specific challenges. 
% %To conduct this evaluation, we employ a diverse range of datasets: ArabicMMLU for broader language understanding, Belebele for dialectal nuances, BoolQ, PIQA, and OBQA for common-sense reasoning, Winograde for linguistic ambiguity, and Misinformation for detecting and understanding false information. 
% By leveraging these datasets, we aim to uncover the strengths and limitations of the models in handling dialectal Arabic with respect to these capabilities.

% We now focus on evaluating LLMs' knowledge and reasoning abilities in dialectal Arabic through tasks covering \textit{world knowledge}, \textit{reading comprehension}, \textit{commonsense reasoning}, and \textit{misinformation}. This provides a comprehensive assessment of their performance with dialect-specific challenges.

We now assess the dialectal Arabic abilities in LLMs through \textit{world knowledge}, \textit{reading comprehension}, \textit{commonsense reasoning}, and \textit{misinformation} tasks, providing a thorough evaluation of their performance on dialect-specific challenges.

\subsubsection{World Knowledge}

\begin{figure}
     \centering
         \includegraphics[width=0.75\linewidth]{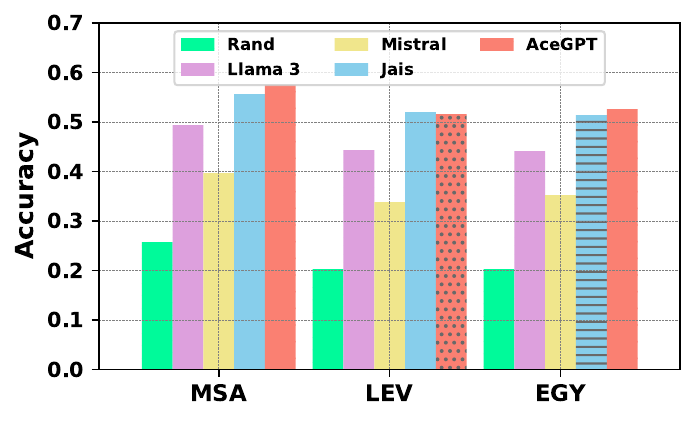}
         \vspace{-0.3cm}
         \caption{Average scores on ArabicMMLU}
         \vspace{-0.3cm}
         \label{fig:avg-scores-arabic-mmlu-ar}
\end{figure}

The overall results in Figure \ref{fig:avg-scores-arabic-mmlu-ar} show that %while AceGPT and Jais excel across the board, especially on the EGY and LEV ArabicMMLU benchmarks proposed in this work, they also maintain strong performance on ArabicMMLU. 
AceGPT and Jais excel across the board on both the EGY and LEV ArabicMMLU benchmarks proposed in this work, and the original MSA ArabicMMLU.  This suggests that these Arabic-centric models are well-suited not only for MSA but also for dialectal variations. In contrast, Llama 3 and Mistral, which are not trained specifically on Arabic, struggle significantly more with EGY and LEV compared to their performance on ArabicMMLU. This highlights the effectiveness of the proposed dialectal benchmarks in revealing performance gaps in general models. %One notable observation is that the performance drop for Mistral is particularly steep in dialects, suggesting a potential limitation in adapting to the linguistic diversity within Arabic. 
These results demonstrate the need for models trained on dialectal data to handle the complexity of regional variations. Detailed individual results are provided in the Appendix \ref{sec:app_world_knowledge}.

\subsubsection{Commonsense Reasoning}

\begin{figure*}
     \centering
     \begin{subfigure}[b]{0.5\textwidth}
         \centering
         \includegraphics[width=\textwidth]{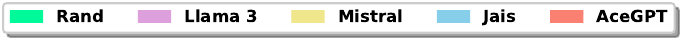}
     \end{subfigure}

     \begin{subfigure}[b]{0.32\textwidth}
         \centering
         \includegraphics[width=\textwidth]{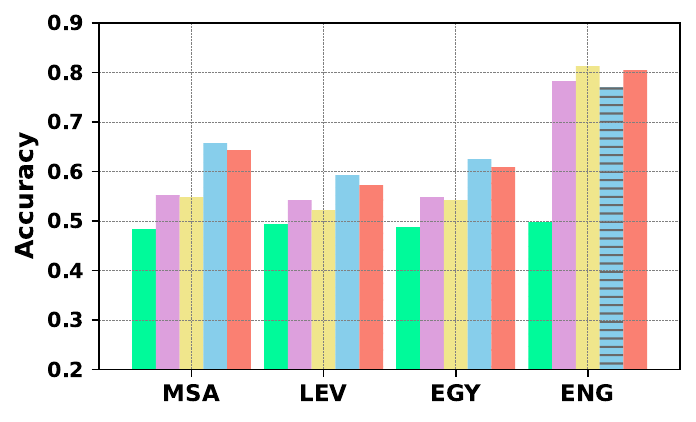}
         \caption{PIQA}
         \label{fig:piqa-acc-norm-scores-transposed}
     \end{subfigure}
     \begin{subfigure}[b]{0.32\textwidth}
         \centering
         \includegraphics[width=\textwidth]{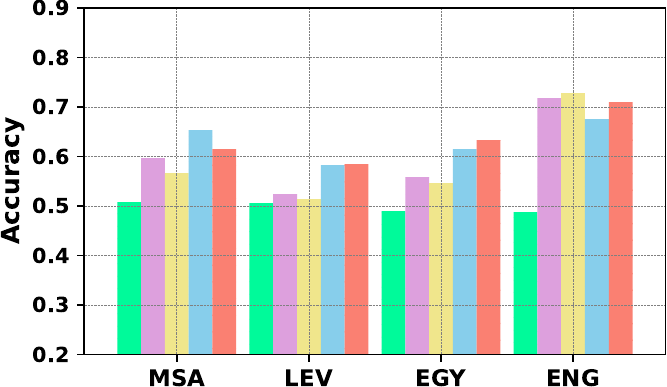}
         \caption{Winogrande}
         \label{fig:winogrande-acc-norm-scores-transposed}
     \end{subfigure}
     \begin{subfigure}[b]{0.32\textwidth}
         \centering
         \includegraphics[width=\textwidth]{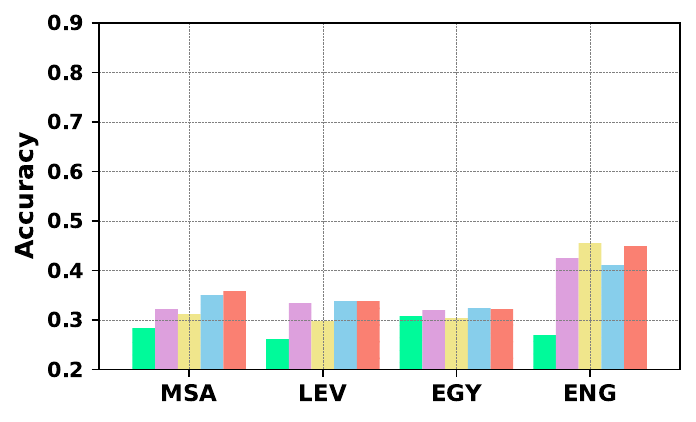}
         \caption{OBQA}
         \label{fig:openbooks-acc-norm-scores-transposed}
     \end{subfigure}
     \vspace{-0.3cm}
     \caption{Results for the \textit{commonsense reasoning} capabilities
     % Models' performance on commonsense Datasets Accuracy (dialects on X-axis)
     }
     \vspace{-0.3cm}
    \label{fig:commonsense-acc-norm-scores-transposed}
\end{figure*}

Our evaluation across the PIQA, OBQA, and Winogrande (See Figure \ref{fig:commonsense-acc-norm-scores-transposed}) reveals several key insights into model performance with respect to MSA and various Arabic dialects, and how these compare to English. Jais consistently outperforms other models across MSA and dialects, including Lev and Egy, demonstrating a strong ability to handle physical commonsense reasoning and complex linguistic nuances. AceGPT also performs well, particularly in Egy, but falls slightly behind Jais in MSA and Lev. This suggests Jais is better tuned for the broader spectrum of Arabic dialects. In contrast, Llama 3 and Mistral show a significant performance drop from English-to-dialectal Arabic, indicating challenges likely due to limited training data for these dialects.

\paragraph{Task-specific Insights} 
Jais leads in handling PIQA across MSA and dialects, indicating its advanced capability in dealing with dialectal intricacies. AceGPT, while effective, shows slightly reduced performance compared to Jais, especially in MSA and Lev dialects. Llama 3 and Mistral's substantial performance drop highlights their difficulties with dialectal Arabic, reinforcing the impact of training data limitations. In OBQA, both AceGPT and Jais perform relatively well in MSA and dialects, with AceGPT having a slight edge in Egy dialect. This performance illustrates their proficiency in handling multi-step reasoning in Arabic. Mistral and Llama 3, despite their strong English performance, struggle with MSA and dialects, reflecting the challenges posed by Arabic-specific content and dialectal variations due to their focus on resource-rich languages during training.

In Winogrande, Jais excels in commonsense reasoning across MSA and Lev dialects, surpassing other models. AceGPT also performs effectively across all dialects but does not exceed Jais, indicating that while it is proficient, it lacks the specialized edge found in Jais. Llama 3 and Mistral show reduced performance in dialects compared to MSA, revealing their limitations in managing dialectal variations and highlighting the impact of their English-centric training.

\begin{figure}
     \centering
         \includegraphics[width=0.75\linewidth]{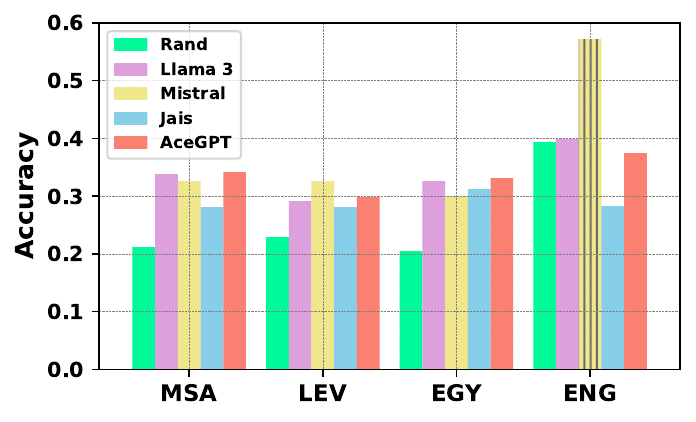}
         \vspace{-0.3cm}
         \caption{Results on \textit{misinformation} (TruthfulQA)}         
         \label{fig:truthfulqa}
         \vspace{-0.3cm}
\end{figure}

\subsubsection{Reading Comprehension} 
\begin{figure*}
\centering
    \begin{subfigure}[t]{0.5\textwidth}
    \includegraphics[width=\textwidth]{figures/legend.pdf}
    \end{subfigure}
     
    \begin{subfigure}[t]{0.36\textwidth}
    \includegraphics[width=\textwidth]{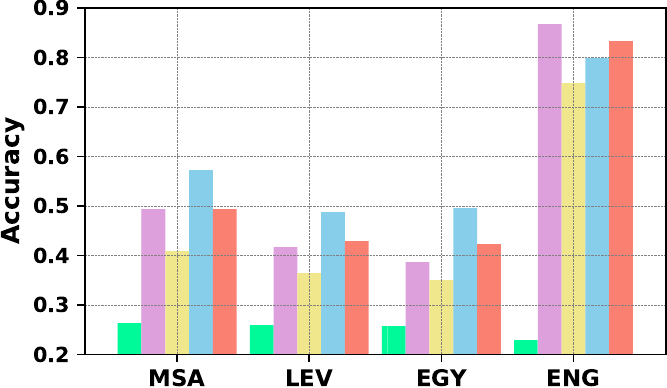}
    \caption{Belebele\label{fig:belebele-acc}}
    \end{subfigure}
    \begin{subfigure}[t]{0.36\textwidth}
    \includegraphics[width=\textwidth]{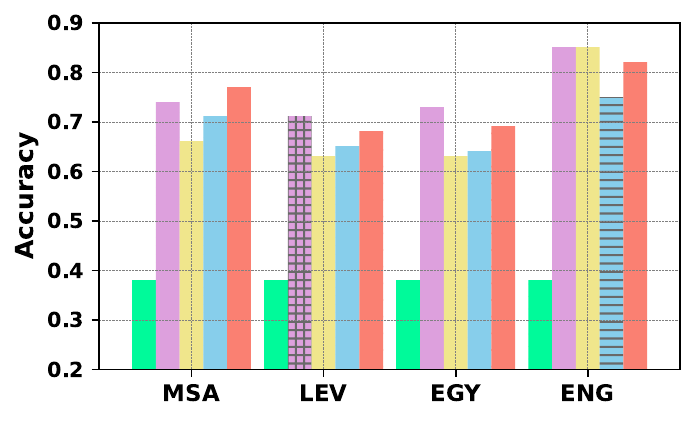}
    \caption{BoolQ\label{fig:boolq-acc}}
    \end{subfigure}
    \vspace{-0.3cm}
    \caption{Results on \textit{reading comprehension}\label{fig:Reading-Comprehension-Results-ACC}}
    \vspace{-0.3cm}
\end{figure*}

Figure \ref{fig:Reading-Comprehension-Results-ACC} show our results on reading comprehension tasks. The results across BoolQ and Belebele provide insights into how different models handle reading comprehension in MSA and dialects compared to English. For the BoolQ task, we observe that while Llama 3 performs exceptionally well in English, achieving 0.85 accuracy, its performance drops noticeably when applied to MSA (0.74) and even further in Lev (0.71) and Egy (0.73). This indicates a significant challenge for general models when transitioning from English to dialectal Arabic, despite the rich knowledge base Llama has. Interestingly, Mistral and Jais show a similar drop in performance across dialects, with Jais maintaining relatively higher accuracy in MSA, likely due to its Arabic-centric training. However, AceGPT stands out with the highest MSA score (0.77) and remains competitive across dialects, suggesting that it better adapts to the linguistic variations within Arabic.

For the Belebele dataset, the performance trends are similar. AceGPT and Jais lead the pack in MSA and dialects, with Jais achieving the highest MSA score (0.57) and performing equally well in Lev and Egy (0.49). This further demonstrates the ability of Arabic-centric models to leverage dialect similarities and perform well across diverse Arabic varieties. Llama and Mistral, while strong in general tasks, show a clear performance gap when tested on Arabic dialects, particularly in Egy, suggesting that they struggle to bridge the linguistic distance between MSA and its dialects.

% \subsection{Reading Comprehension}

\subsubsection{Misinformation}
The results for the TruthfulQA task across Lev, Egy, and MSA show that model performances are often close to random, especially for Lev and Egy, as reported in Figure \ref{fig:truthfulqa}. Mistral performs best overall, particularly in Lev, but its scores are only slightly above random. We leave a detailed exploration of this for the future.

\subsection{Cultural Understanding} 
\label{sec:cultural-understanding-results}
\begin{figure}
     \centering
         \includegraphics[width=0.75\linewidth]{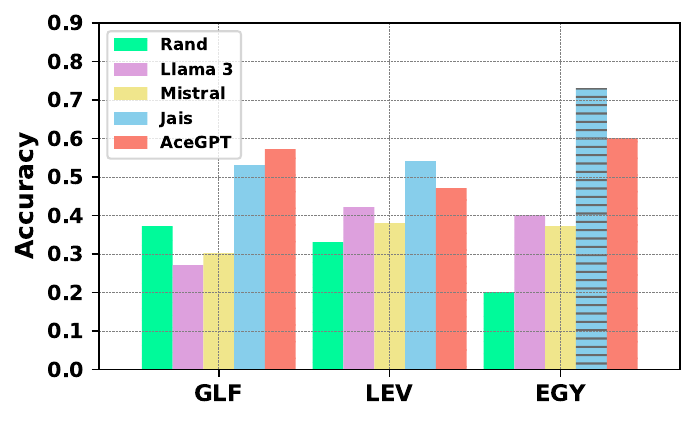}
         \vspace{-0.2cm}
         \caption{Results on \textit{cultural understanding}}         
         \label{fig:cultural-understanding-mcq}
         \vspace{-0.3cm}
\end{figure}

%\begin{table}[]
%\centering
%\resizebox{\columnwidth}{!}{%
%\begin{tabular}{@{}lccccc@{}}
%\toprule
%\textbf{Model}    & \textbf{Rand} & \textbf{Llama 3} & \textbf{Mistral} & \textbf{Jais}  & \textbf{AceGPT} \\ \midrule
%Score    & 0.32 & 0.39   & 0.37    & 0.57  & 0.50   \\ \bottomrule
%\end{tabular}%
%}
%\vspace{-0.3cm}
%\caption{Average accuracy on the \textit{cultural MCQ} task. Rand: Random}
%\label{tab:cultural-understanding-mcq}
%\vspace{-0.3cm}
%\end{table}

The results on the cultural understanding task (MCQ version--Figure \ref{fig:cultural-understanding-mcq}), indicate that Jais is the most culturally aligned model, followed closely by AceGPT, both showing  superior awareness of the Egyptian culture. One possible justification is that the Egyptian population (and consequently its online presence) is significantly larger than the other two regions. Llama 3 and Mistral generally show performances close to the random baseline, suggesting limited awareness of Arabic culture. 

In a qualitative analysis of the models' responses (when prompting the models in a generation setup), Llama 3 frequently generated fictional entities (e.g., names of people or holidays) and lacked geographical and historical knowledge. For example, when asked about the highest mountain in Jordan, it incorrectly named a mountain in Saudi Arabia. %Llama 3 also struggled with regional cultural nuances, often listing famous dishes from other Levant countries when asked about Jordan. Additionally, the model %had difficulty generating grammatically correct Arabic sentences and %displayed biases about 
%challenged Arabic cultures, such as incorrectly suggesting government punishment for not fasting during Ramadan. Mistral, on the other hand, failed to generate proper Arabic sentences altogether, producing frequent spelling errors and misinterpreting questions. 
Jais and AceGPT performed better overall, though AceGPT had some issues following instructions as accurately as Jais. Further discussion and example models' responses can be found in the Appendix \ref{sec:appendix:culture-understanding}.

\subsection{Additional Model Evaluations}

To validate the applicability of the proposed \ds{} benchmark, we expand our evaluation to cutting-edge models in Arabic and multilingual language modeling, including Fanar \cite{fanar2024}, Gemma-2-9B \cite{gemmateam2024gemma2improvingopen}, Aya-Expanse-8B \cite{ayamodelinstructionfinetuned}, Qwen2.5-7B \cite{yang2024qwen2technicalreport}, Llama-3.1-8B-Instruct, and AceGPT-v2-8B-Chat. These models represent diverse architectures and training paradigms, offering valuable insights into handling complex linguistic phenomena. Fanar, built on Gemma, highlights the potential of Arabic-centric pretraining, while Qwen and Aya reflect innovations in multilingual fine-tuning. Due to resource constraints, we focus on Arabic MMLU and PIQA datasets, which assess performance on diverse, challenging tasks. This extended evaluation underscores AraDiCE's role in advancing Arabic LLM benchmarks and guiding future developments.

% To further validate the applicability of the proposed \ds{} benchmark, we extend our evaluation to include several cutting-edge models that have recently emerged in the domain of Arabic and multilingual language modeling. Specifically, we additionally assess the performance of Fanar \cite{fanar2024}, Gemma-2-9b \cite{gemmateam2024gemma2improvingopen}, Aya-Expanse-8B \cite{ayamodelinstructionfinetuned}, Qwen2.5-7B \cite{yang2024qwen2technicalreport}, Llama-3.1-8B-Instruct, and AceGPT-v2-8B-Chat, which represent diverse architectural approaches and training paradigms. These models are particularly notable for their advancements in handling complex linguistic phenomena and their potential to set new state-of-the-art benchmarks. Due to computational resource constraints, we focus our evaluation on the Arabic MMLU and PIQA datasets, which are key to assessing the performance of these models on diverse tasks. By including these evaluations, we aim to provide a more comprehensive understanding of how AraDiCE enables fine-grained benchmarking and supports future developments in Arabic LLMs. 
% Arabic natural language processing.

\begin{figure}
     \centering
         \includegraphics[width=0.75\linewidth]{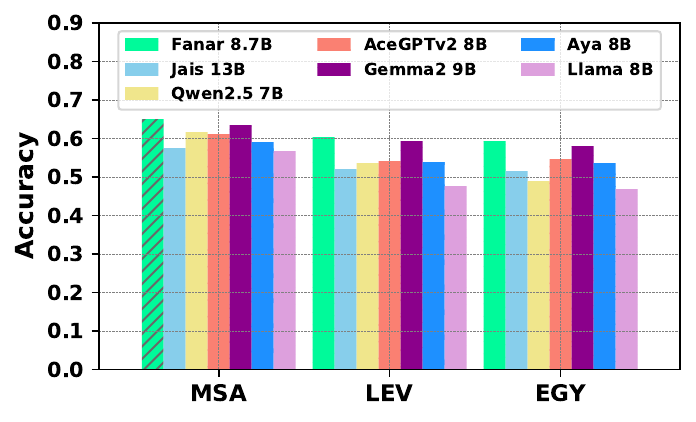}
         \vspace{-0.3cm}
         \caption{Performance of various models on the Arabic MMLU across MSA, Levantine, and Egyptian.}
         \vspace{-0.3cm}
         \label{fig:avg-scores-arabic-mmlu-fanar-code}
\end{figure}

\begin{figure}
     \centering
         \includegraphics[width=0.75\linewidth]{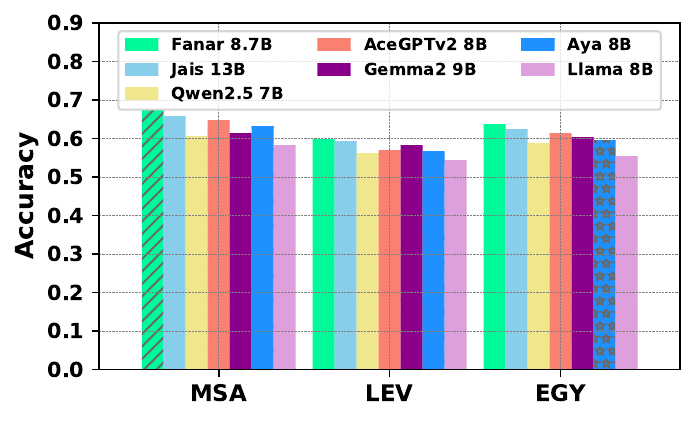}
         \vspace{-0.3cm}
         \caption{Performance of various models on the Arabic PIQA across MSA, Levantine, and Egyptian.}
         \vspace{-0.3cm}
         \label{fig:avg-scores-piqa-fanar-code}
\end{figure}

% \paragraph{Arabic MMLU:} As we observed earlier, the models generally show stronger performance on MSA compared to the dialects, reflecting the resource-rich training available for MSA data. As presented in Figure \ref{fig:avg-scores-arabic-mmlu-fanar-code}, among all the models, Fanar 8.7B stands out as the top performer, demonstrating strong capabilities in handling dialects, though still somewhat less proficient than MSA. Gemma2 9B shows strong performance, especially in the dialects, approaching Fanar's results. However, it still lags behind Fanar in MSA, indicating that it excels in dialectal handling but struggles with MSA tasks.

\paragraph{ArabicMMLU:} As we observed earlier, the models generally perform better on MSA than on the dialects, reflecting the resource-rich training available for MSA data. Among the models (as shown in Figure \ref{fig:avg-scores-arabic-mmlu-fanar-code}), Fanar 8.7B stands out as the top performer, showcasing strong capabilities in both MSA and dialects. Its superior performance in MSA can be attributed to its MSA-specific training and its foundation on Gemma2 9B, which provides a robust baseline. Notably, Gemma2 outperforms other Arabic-centric models on dialects in the ArabicMMLU, highlighting its strength in handling dialectal variations.

\paragraph{PIQA:} As shown in Figure \ref{fig:avg-scores-piqa-fanar-code}, in PIQA, a task focused on common-sense reasoning, Fanar 8.7B again outperforms the other models, particularly in Egyptian Arabic. Jais 13B follows closely behind, demonstrating solid generalization across MSA and dialects, though it still lags behind Fanar in dialectal contexts. The Qwen2.5 7B model, while less competitive in MMLU, performs better on PIQA for Levantine and Egyptian, suggesting that its reasoning capabilities are more adaptable to dialectal inputs. Gemma2 9B and Aya 8B show more consistent, though lower, performance across both dialects and MSA. The gap between Gemma2 and Aya is relatively small, with Aya performing slightly better in Egyptian Arabic in both MMLU and PIQA tasks, suggesting that Aya might have some potential in specialized domains.

These results highlight the ongoing challenges of Arabic dialectal processing and reinforce the efficacy of \ds{} as a benchmark that can expose performance gaps across dialects. Moreover, the models tested here exhibit varying degrees of sensitivity to dialectal variations, indicating that future model development and fine-tuning efforts need to take into account dialectal diversity to improve overall Arabic LLMs capabilities.

%% file: sections/related_work.tex
\section{Related Work}
\label{sec:related work}

% There has been considerable work on benchmarking the abilities of language models in Arabic.  Some of the prior research has included Arabic as a part of multilingual benchmarks for natural language understanding (NLU). Prominent examples include XGLUE \cite{liang-etal-2020-xglue}, XTREME \cite{hu2020xtreme}, XTREME-R \cite{ruder-etal-2021-xtreme}, and GEM \cite{gehrmann-etal-2021-gem}.  These benchmarks encompass a variety of tasks, predominantly focusing on classification problems such as natural language inference, part-of-speech tagging, named entity recognition, and summarization.
There has been considerable work on benchmarking language models in Arabic, with prior research often including Arabic in multilingual benchmarks like XGLUE \cite{liang-etal-2020-xglue}, XTREME \cite{hu2020xtreme}, XTREME-R \cite{ruder-etal-2021-xtreme}, GEM \cite{gehrmann-etal-2021-gem}, Dophin \cite{nagoudi-etal-2023-dolphin}. These benchmarks cover a variety of tasks, primarily focusing on classification problems such as natural language inference and generation, part-of-speech tagging, named entity recognition, and summarization.

Another relevant dataset is EXAMS \cite{hardalov-etal-2020-exams}, featuring multilingual school exam questions, including about 500 in Arabic. This dataset extends evaluation to educational contexts, providing a diverse testing ground for language models. Notable benchmarks in the realm of question answering (QA), that primarily aim to evaluate models on reading comprehension and QA include TyDiQA \cite{clark-etal-2020-tydi}, Arabic-SQuAD \cite{mozannar-etal-2019-neural}, and MLQA \cite{lewis2019mlqa}. These datasets primarily aim to evaluate models on reading comprehension and QA. 
More recently, \citet{koto2024arabicmmlu} introduced ArabicMMLU, focusing on %tailored to the specific needs of 
the MENA region. The dataset includes 40 tasks in MSA. However, a limitation of their work, is the lack of %focus on 
dialectal Arabic.

% There has been considerable effort in training Arabic language models (LMs) \cite{antoun-etal-2020-arabert, inoue-etal-2021-interplay, abdul-mageed-etal-2021-arbert, nagoudi-etal-2022-arat5} and, more recently, LLMs such as Jais \cite{sengupta2023jais}, AceGPT \cite{huang2024acegpt}, ALLaM \cite{bari2024allamlargelanguagemodels}, Arabic Stable LM \cite{alyafeai2024arabic}, and FANAR \cite{fanar2024}. Jais is pre-trained from scratch using 72 billion Arabic tokens, while AceGPT builds on Llama2 using reinforcement learning from AI feedback. ALLaM is a family of models ranging from 7B to 70B parameters, which are not publicly available. Arabic Stable LM is fine-tuned on top of Stable LM 2 with 1.6B model. Fanar is an Arabic-centric generative AI system, that supports
% diverse language, speech and image generation tasks. 

Significant efforts have been made in training Arabic language models (LMs) \cite{antoun-etal-2020-arabert, inoue-etal-2021-interplay, abdul-mageed-etal-2021-arbert, nagoudi-etal-2022-arat5} and more recently, LLMs like Jais \cite{sengupta2023jais}, AceGPT \cite{huang2024acegpt}, ALLaM \cite{bari2024allamlargelanguagemodels}, Arabic Stable LM \cite{alyafeai2024arabic}, and FANAR \cite{fanar2024}. Jais is trained from scratch on 72 billion Arabic tokens, AceGPT builds on Llama2 with reinforcement learning from AI feedback, ALLaM spans 7B–70B parameters but is not publicly available, Arabic Stable LM fine-tunes Stable LM 2 with 1.6B parameters, and FANAR LLM supports diverse language, speech, and image generation tasks.

There has also been efforts benchmarking LLMs for Arabic, most notably focusing on standard LLMs based dataset evaluation \cite{sengupta2023jais} and NLP tasks \cite{khondaker-etal-2023-gptaraeval,abdelali-etal-2024-larabench,dalvi-etal-2024-llmebench}, prompting in native (Arabic) and non-native (English) language \cite{Kmainasi2024wise} and multimodality \cite{alwajih2024dallah,das-etal-2024-exams}.

Focusing on benchmarking the cultural capabilities of LLMs, this includes measuring how entity mentions are culturally biased towards Western or Arab contexts \cite{naous-etal-2024-beer}, and assessing cultural alignment based on the World Values Survey \cite{alkhamissi-etal-2024-investigating}. Other notable work on cultural benchmarking include \cite{demidova-etal-2024-john,shen-etal-2024-understanding,liu-etal-2024-multilingual,arora2024calmqa,myung2024blend}.
% Other recent benchmark for Arabic  include Dallah \cite{alwajih2024dallah}, Exams-V
While prior work has focused on benchmarking LLMs for MSA, our study extends this by evaluating both MSA and dialects (Lev and Egy) using Arabic-centric and non-Arabic-centric LLMs.

% In this work, we benchmark these two Arabic LLMs alongside Mistral and Llama-3-7B, which have multilingual capabilities.  
% In this work, we address this gap by creating the first test suite for benchmarking dialectal Arabic, covering two dialects: Levantine and Egyptian. 

% In contrast, the MMLU dataset is designed to assess reasoning and real-world knowledge through multiple-choice questions.

% Much of this prior research has included Arabic as a part of multilingual benchmarks for natural language understanding (NLU). Prominent examples include XGLUE, XTREME, XTREME-R, and GEM. These benchmarks encompass a variety of tasks, predominantly focusing on classification problems such as natural language inference, part-of-speech tagging, named entity recognition, and summarization.

% Another relevant dataset is EXAMS, which comprises multilingual school exam questions, including a subset of approximately 500 questions in Arabic. This dataset broadens the scope of evaluation to educational contexts, offering a diverse testing ground for language models.

% https://arxiv.org/abs/2306.09212 [CMMLU]

% https://arxiv.org/pdf/2402.12840v1 [Arabic MMLU]

% JAIS paper 

% ACE GPT

%% file: sections/conclusions.tex
\section{Conclusions}
\label{sec:conclusions}

We developed dialectal Arabic benchmarks through machine translation and post-editing. Our benchmarks evaluate various NLP tasks, including understanding, generation, and cultural awareness across Arabic dialects. Our results show that while Arabic-centric models like FANAR, Jais and AceGPT perform better in dialectal contexts, they still face challenges compared to MSA and English. Performance varies by dialect and task, highlighting the need for more specialized training for effective handling of regional linguistic and cultural nuances. We have released our dialectal benchmarks and models to support future research and advancements in NLP for low-resource languages. We also released the dialectal translation models\footnote{\url{https://huggingface.co/datasets/QCRI/AraDiCE}}
and benchmarks developed in this study to support further research %and advancements 
in NLP for low-resource languages.

%% file: sections/appendices.tex
\input{sections/appendices_dataset}

\section{Machine Translation}
\label{sec:appendix:MT}

Machine translation of Arabic is challenging due to morphological complexity and dialectal variations \cite{birch-etal-2014-edinburgh,sajjad-etal-2017-challenging}. Here we detail our efforts in training MSA--dialect models.

\subsection{Data}
\label{sec:appendix:data}

We used the dataset listed below to develop the machine translation system. We provide a brief description of each dataset. 
 % We utilize the MADAR corpus \cite{bouamor-etal-2018-madar}, UFAL corpus \cite{krubinski-etal-2023-multi}, LDC corpus, \footnote{https://catalog.ldc.upenn.edu/byproject} DIA2MSA dataset \cite{mubarak2018dial2msa}, ArzEn-MultiGenre Dataset \cite{10023181} and the SADID dataset \cite{abid-2020-sadid}. 
 
 % Below we provide a brief description of each dataset. 
\begin{itemize}
    \item \textbf{Madar} \cite{bouamor-etal-2018-madar}: Parallel corpus of 25 Arabic city dialects in the travel domain. It comprises data in the Levantine, Egyptian, Moroccan, and Gulf dialects. 
    \item \textbf{UFAL Parallel Corpus of North Levantine 1.0}  \cite{krubinski-etal-2023-multi}: Consists of $\sim120K$ %120,600 
    parallel sentences in English, French, German, Greek, Spanish, MSA from the OpenSubtitles2018 corpus and manually translated to the north Levantine dialect.
    \item \textbf{LDC}: From the LDC catalogue we utilize the Arabic-Dialect/English Parallel Text \footnote{https://catalog.ldc.upenn.edu/LDC2012T09} (referred to as LDC), and the BOLT Arabic Discussion Forum Parallel Training Data (referred to as BOLT).\footnote{https://catalog.ldc.upenn.edu/LDC2019T01}
    \item \textbf{DIA2MSA} \cite{mubarak2018dial2msa}: Consists of tweets written in four Arabic dialects (Egyptian, Gulf, Levantine, and Maghrebi) and their corresponding MSA translations. 
    \item \textbf{ArzEn-MultiGenre} \cite{10023181}: A parallel dataset of Egyptian Arabic Songs, Lyrics, Novels, and TV show subtitles that were human-translated. 
     \item \textbf{PADIC} \cite{meftouh:hal-01261587}: Parallel corpus containing dialectal Arabic texts covering six Arabic cities. 
     %(Algerian, Annaba, MSA, Syrian, Moroccan, Palestinian).
\end{itemize}

In some cases, the dataset does not contain a dialectal counterpart. To address this issue, we translate the data from English to the corresponding dialect using the NLLB (nllb-200-3.3B)\footnote{\url{https://huggingface.co/facebook/nllb-200-3.3B}} base model~\cite{nllbteam2022language}. The number of parallel sentences included in each dataset is presented in Table \ref{table:data-stats}. We used dialectal tests in AraBench \cite{sajjad-etal-2020-arabench} for our evaluation. 

\begin{table}[] 
\centering
\setlength{\tabcolsep}{10pt}
\scalebox{0.9}{%    
\begin{tabular}{lrrr}
\toprule
\textbf{Dataset} & \multicolumn{1}{l}{\textbf{EGY}} & \multicolumn{1}{l}{\textbf{GLF}} & \multicolumn{1}{l}{\textbf{LEV}} \\\midrule
MADAR   & 17,885                  & 25,759                   & 21,853                  \\
UFAL    & 0                       & 0                        & 120,600                 \\
LDC     & 38,154                   & 0                        & 138,010                  \\
BOLT    & 113,394                  & 0                        & 0                       \\
Dia2msa & 16,355                  & 18,000                    & 18,000                   \\
Arzen   & 24,530                   & 0                        & 0                       \\
PADIC   & 0                       & 0                        & 14,426                 \\\bottomrule
\end{tabular}%
}
\caption{Statistics for the datasets used for training.}
\label{table:data-stats}
\end{table}

\subsection{Models}
\label{sec:appendix:models}

We fine-tuned two robust machine translation models: AraT5 \cite{nagoudi-etal-2022-arat5} and NLLB \cite{nllbteam2022language}. We experimented with several variants of these models, with sizes ranging from 600M to 3.3B parameters. AraT5 is an Arabic language model based on the T5 (Text-to-Text Transfer Transformer) architecture. The NLLB 
% (No Language Left Behind) 
model is a state-of-the-art machine translation model developed by Meta, as part of their initiative to improve translation quality across a broad spectrum of languages, including Arabic.

\begin{table}[]
\centering
\scalebox{0.9}{%    
\begin{tabular}{crrrr}
\toprule
 & \textbf{OSACT} &  \textbf{SADID} &  \textbf{LDC} & \textbf{D2M}\\
 \midrule
 S1 &  9.8 &  12.7 & 6.2 & 11.0\\
 S2 &  9.8 &  11.8 & 6.3 & 11.7\\
 S3 &  9.7 &  11.8 & 7.0 & 47.8\\
 S4 & 5.92 & 8.42 & 3.46 & 4.89 \\\bottomrule
\end{tabular}
}
\caption{SacreBLEU Scores on test sets for our MSA-to-LEV models: \textit{S1} = UFAL, \textit{S2} = +LDC, MADAR, PADIC, D2M, \textit{S3} = +LDC, MADAR, PADIC, D2M, \textit{S4} = GPT4 zero-shot.}
\label{tab:lev-model-scores}
\end{table}

\begin{table}[]
\centering
\scalebox{0.9}{%    
\begin{tabular}{ccccrrrr}
\toprule
 &  \textbf{ARZEN} & \textbf{D2M} &  \textbf{LDC} &  \textbf{MADAR} \\\midrule
 S1 & 1.8 & 57.3 & 11.8 & 17.7 \\
 S2 & 17.3 & 57.2 & 12.3 & 17.6 \\
 S3 & 15.8 & 55.0 & 11.2 & 17.9 \\
 S4 & 1.88 & 7.53 & 2.83 & 6.02 \\\bottomrule
\end{tabular}
}
\caption{SacreBLEU scores on test sets for the selected MSA-to-EGY models: \textit{S1} = MADAR + D2M + LDC, \textit{S2} = MADAR + D2M + LDC + Arzen, \textit{S3} = MADAR + D2M + LDC + Arzen + BOLT, \textit{S4} = GPT4 zero-shot.}
\label{tab:egy-model-scores}
\end{table}

\subsection{Machine Translation Results}
\label{ssec:appendix_mt_results}
% In our preliminary experiments, we found NLLB 3.3B model to surpass AraT5 and its smaller variants, post fine-tuning with dialectal data

In our preliminary experiments, we found that the NLLB 3.3B model outperformed AraT5 and its smaller variants after fine-tuning with dialectal data (Table \ref{tab:mt-models-comparison}).
% (Please see Appendix \ref{sec:appendix:models} for details)
% \nd{Basel put model comparisons between NLLB variants and AraT5 in appendix.} 
We carried out ablation studies using different data mixtures on the NLLB 3.3B model. Initially, we shortlisted three systems per dialect using BLEU scores as our primary criterion (Tables \ref{tab:lev-model-scores}, \ref{tab:egy-model-scores}). 

Nonetheless, basing decisions solely on BLEU scores presents a risk of over-fitting. We observed improved performance on specific tests after incorporating in-domain training data. Therefore, we carried out a human evaluation of the selected systems to select the best model for translation. For this, we selected a sample of 100 sentences from different genres, and carried out translations using the systems. 
We conducted a human evaluation by showing the output of three translation systems and asked the participants to select the ``best system(s)''. %\footnote{Note that the translations were presented in a Google Doc.} 
The instructions to the annotators were as follows:

\begin{enumerate}[noitemsep,topsep=0pt,leftmargin=*,labelwidth=!,labelsep=.5em]
    \item Find a row that does not yet have an entry under ``best system,'' and enter S1, S2, or S3 to represent the system with the best translation.
    \item If two systems have the same quality, you may enter both under ``best system'' (e.g., S1/S3).
    \item If none of the translations are acceptable, enter ``0'', referred as \textit{None}.
    \item The translations may be in any specific sub-dialect (Palestinian, Lebanese, Jordanian, or Syrian).
\end{enumerate}

% , along with an additional option labeled \textit{None}. The evaluators' task was to read the translations from the three systems and select the best one. They were instructed to choose \textit{None} if none of the translations were acceptable.

In Tables \ref{tab:human_evaluation_lev} and \ref{human_evaluation_egy}, we report our human evaluation results, which
% Our human evaluation results (see Tables \ref{tab:human_evaluation_lev} and \ref{human_evaluation_egy}) 
show that System S1, which is NLLB tuned just using UFAL data was best for translating MSA-to-Levantine and System S2 trained using MADAR, Dial2MSA and LDC data was better for translating MSA-to-Egyptian translation. However, in our human evaluation, we found that humans had more confidence in the MSA-to-Levantine system than in MSA-to-Egyptian.

\begin{table}[]
\centering
\resizebox{0.48\textwidth}{!}{%
\begin{tabular}{lccccc}
\toprule
 & \multicolumn{1}{c}{\textbf{LDC}} & \multicolumn{1}{c}{\textbf{Madar}} & \multicolumn{1}{c}{\textbf{UFAL}} & \multicolumn{1}{c}{\textbf{Oscat}} & \multicolumn{1}{c}{\textbf{Avg.}} \\\midrule
NLLB3.3B & 3.8 & 30.4 & 32.4 & 11.7 & 19.6 \\
AraT5    & 4.3 & 21.4 & 33.4 & 12.2 & 17.8 \\
Turjuman & 4.1 & 19.3 & 33.2 & 11.3 & 17.0 \\ \bottomrule
\end{tabular}%
}
\caption{%Comparison of 
% SacreBLEU scores on test sets with models trained on the UFAL data mixture.
SacreBLEU scores on test sets for models trained on the UFAL data mixture. Avg.: Average
}
\label{tab:mt-models-comparison}
\end{table}
\begin{table}[h]
\centering
\scalebox{0.9}{%    
\begin{tabular}{lrrrr}
\toprule
& \textbf{S1} & \textbf{S2} & \textbf{S3} & \textbf{None} \\\midrule
Wikipedia & 49.9\% & 46.0\% & 45.2\% & 16.5\% \\
Zaman & 52.3\% & 31.5\% & 36.1\% & 20.5\% \\
Hindawi  & 53.8\% & 51.7\% & 51.5\% & 11.1\% \\\bottomrule
\end{tabular}
}
\caption{Human evaluation of different MSA-to-LEV systems across datasets. \textit{None} refers that the output from none of the systems were acceptable.}
\label{tab:human_evaluation_lev}
\end{table}

\begin{table}[h]
\centering
\scalebox{0.9}{%   
\begin{tabular}{lcccc}
\toprule
& \textbf{S1} & \textbf{S2} & \textbf{S3} & \textbf{None} \\\midrule
Wikipedia & 19.4\% & 28.6\% & 27.6\% & 46.9\% \\
Zaman    & 12.1\% & 18.2\% & 17.2\% & 59.6\% \\
Hindawi  & 29.3\% & 33.3\% & 29.3\% & 35.4\% \\\bottomrule
\end{tabular}%
}
\caption{Human evaluation for the selected MSA-to-EGY models.\textit{None} refers that the output from none of the systems were acceptable.}
\label{human_evaluation_egy}
\end{table}

\section{Detailed Results on Dialect Understanding and Generation}
\label{sec:appendix:understanding-generation}

\subsection{Dialect Identification}
\label{sec:appendix:dialect-id}
\begin{figure*}[h]
     \centering
     \begin{subfigure}[b]{0.24\textwidth}
         \centering
         \includegraphics[width=\textwidth]{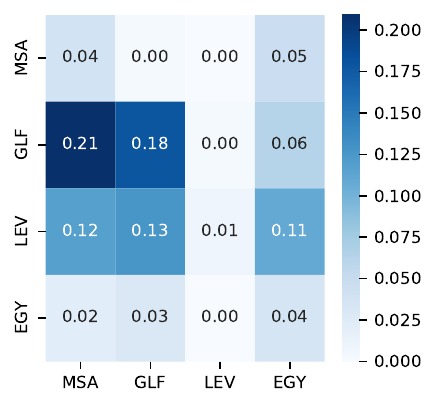}
         \caption{Llama 3}
         \label{fig:llama3-adi-confusion-matrix}
     \end{subfigure}
     \begin{subfigure}[b]{0.24\textwidth}
         \centering
         \includegraphics[width=\textwidth]{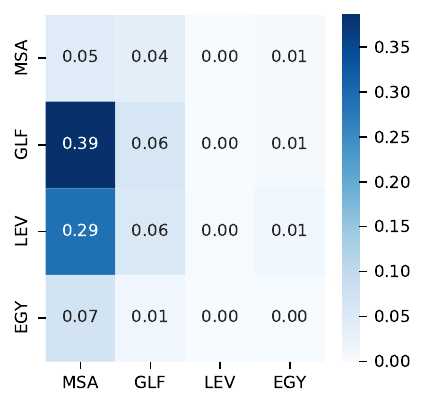}
         \caption{Mistral}
         \label{fig:mistral-adi-confusion-matrix}
     \end{subfigure}
     \hfill
     \begin{subfigure}[b]{0.24\textwidth}
         \centering
         \includegraphics[width=\textwidth]{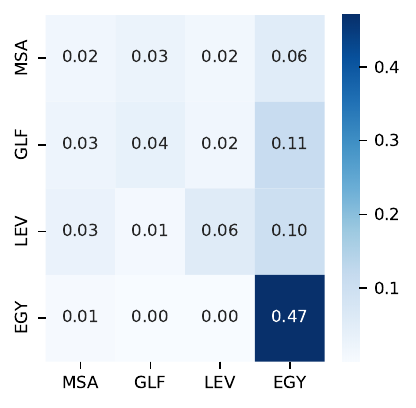}
         \caption{Jais}
         \label{fig:jais-adi-confusion-matrix}
     \end{subfigure}
     \hfill
    \begin{subfigure}[b]{0.24\textwidth}
         \centering
         \includegraphics[width=\textwidth]{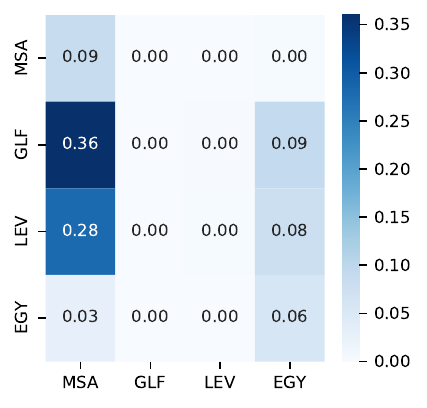}
         \caption{AceGPT}
         \label{fig:ace-gpt-adi-confusion-matrix}
     \end{subfigure}
     \caption{Confusion matrices on ADI dataset.}
    \label{fig:adi-confusion-matrices}
\end{figure*}

\begin{figure*}[h]
     \centering
     \begin{subfigure}[b]{0.24\textwidth}
         \centering
         \includegraphics[width=\textwidth]{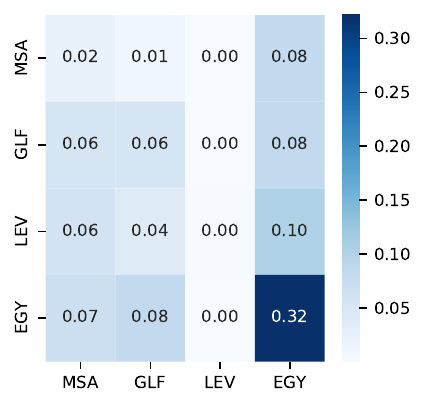}
         \caption{Llama 3}
         \label{fig:llama3-add-confusion-matrix}
     \end{subfigure}
     \begin{subfigure}[b]{0.24\textwidth}
         \centering
         \includegraphics[width=\textwidth]{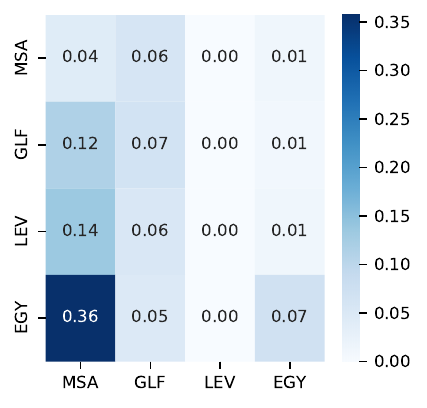}
         \caption{Mistral}
         \label{fig:mistral-add-confusion-matrix}
     \end{subfigure}
     \hfill
     \begin{subfigure}[b]{0.24\textwidth}
         \centering
         \includegraphics[width=\textwidth]{figures/add-jais_final.pdf}
         \caption{Jais}
         \label{fig:jais-add-confusion-matrix}
     \end{subfigure}
     \hfill
    \begin{subfigure}[b]{0.24\textwidth}
         \centering
         \includegraphics[width=\textwidth]{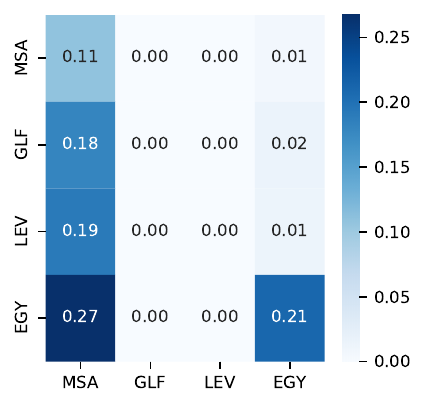}
         \caption{AceGPT}
         \label{fig:ace-gpt-add-confusion-matrix}
     \end{subfigure}
     \caption{Confusion matrices on the Arabic dialects dataset.}
    \label{fig:add-confusion-matrices}
\end{figure*}

% The F1 scores of the models on the task of 
The results of the dialect identification task are shown in Table \ref{tab:dialect-id-results}. In Figures \ref{fig:adi-confusion-matrices} and \ref{fig:add-confusion-matrices} we present the confusion matrices for the ADI and ADD datasets, respectively. For ADI, Llama 3, Mistral, and AceGPT tend to confuse MSA with the Levantine and Gulf dialects, while Jais performs best in detecting the Egyptian dialect. 
% Figure \ref{fig:add-confusion-matrices} the confusion matrices for the ADD dataset. 
For ADD, Llama 3 and Jais excel at detecting the Egyptian dialect, whereas AceGPT confuses all dialects with MSA. The variation in model performance is attributed to the nature of the data and its underlying distribution. 
% shows the confusion matrices on the ADI dataset. Llama 3, Mistral and AceGPT tend to confuse MSA with the Levantine and Gulf dialects and Jais seems to be best at detecting the Egyptian dialect. Figure \ref{fig:add-confusion-matrices} shows the confusion matrices on the ADD. %Arabic dialects dataset. 
% Llama3 and Jais shine when detecting the Egyptian dialect while AceGPT confuses all dialects with MSA. The variation in performance of the models is attributed to the nature of the data and its underlying distribution.


% \section{Dialect Selection}

\subsection{Dialect Generation}
\label{sec:appendix:dialogue-generation}

The F1 scores of the models considered on the task of dialect generation (modeled as a selection task) are shown in Table \ref{table:dialogue-selection-results}. We conducted qualitative analysis on the generation task of the models by prompting them in the dialectal version of \textit{``How would you respond to this phrase in the \textit{X} dialect''}, where \textit{X} could be one of the Egyptian, Levantine, or Gulf dialects (examples on models' response can be found in Table~\ref{tab-appx-dialect-gen-responses}). We noticed that Llama 3 struggles to follow the dialectal instructions and replies instead with description of the input phrase -- for example, \textit{What a great question!, What a challenge! thanks for the question!}. While mistral appears to not be able to comprehend the prompt in the first place. It replies with random phrases about the input sentence without abiding by the instruction. On the other hand, Jais tends to copy the input phrase without replying to the question. These results show that models struggle to follow dialectal instructions.
 
\begin{table}[h]
\centering
\scalebox{0.9}{%   
\begin{tabular}{@{}lccc@{}}
\toprule
& \multicolumn{1}{c}{\textbf{QADI}} & \multicolumn{1}{c}{\textbf{ADI}} & \multicolumn{1}{c}{\textbf{ADD}} \\ \midrule
Random & 0.08& 0.09& 0.12\\
Llama 3 & 0.42& 0.26& 0.40\\
Mistral & 0.21& 0.11& 0.18\\
Jais & 0.31& 0.40& 0.58\\
AceGPT & 0.12& 0.15& 0.32\\
SOTA & 0.85& 0.70& 0.66\\ \bottomrule 
\end{tabular}
}
\caption{Results (measured in F1) for the dialect identification task.}
\label{tab:dialect-id-results}
\end{table}

\begin{table}[h]
\centering
\scalebox{0.9}{%   
\begin{tabular}{@{}lccc@{}}
\toprule
 & \multicolumn{1}{c}{\textbf{LEV}} & \multicolumn{1}{c}{\textbf{EGY}} & \multicolumn{1}{c}{\textbf{GLF}} \\ \midrule
Random  & 0.28 & 0.24 & 0.48 \\
Llama 3   & 0.32 & 0.43 & 0.63 \\
Mistral & 0.28 & 0.33 & 0.62 \\
Jais    & 0.49 & 0.50 & 0.74 \\
AceGPT & 0.48 & 0.58 & 0.74 \\ \bottomrule
\end{tabular}
}
\caption{Results (measured in accuracy) for the dialect generation task. 
% Accuracy on the MCQ response selection task
}
\label{table:dialogue-selection-results}
\end{table}

\begin{table*}[h]
\centering
\resizebox{\textwidth}{!}{%    
\begin{tabular}{l|p{5cm}|p{5cm}|p{5cm}|p{5cm}}
\toprule
\textbf{\#} & \textbf{Input} & \textbf{Llama 3} & \textbf{Mistral} & \textbf{Jais}\\\midrule
\textbf{Ex1} & \begin{RLtext}\footnotesizeوشك مصفر النهردة جامد ايه الي حاصل معاك؟\end{RLtext} \leavevmode\newline Your face looks really pale today. What's going on with you? & What a great question! & \begin{RLtext}\footnotesizeللتردد على الجملة باللغة المصرية، يمكنك استخدام الترجمة التالية: \end{RLtext} \leavevmode\newline To hesitate over the sentence in Egyptian, you can use the following translation: & \begin{RLtext}\footnotesizeوشك مصفر النهارده جامد؟ ايه اللي حاصل معاك؟\end{RLtext} \leavevmode\newline Your face is really pale today. What's going on with you?\\\midrule
\textbf{Ex2} & \begin{RLtext}\footnotesizeشكلك قعدت مكاني يا استاذ\end{RLtext} \leavevmode\newline It looks like you sat in my seat, sir. & \begin{RLtext}\footnotesizeشلونك قعدت مكاني يا استاذ؟! \end{RLtext} \leavevmode\newline How did you end up sitting in my seat, sir?!  & \begin{RLtext}\footnotesizeأجابة: شكرت شكلك قعدتي مكانك يا شاغر, باللغة المصرية.\end{RLtext} \leavevmode\newline Answer: Thank you, it looks like you sat in your place, vacant, in Egyptian Arabic. & \begin{RLtext}\footnotesize شكلك قعدت على كرسيي يا استاذ \end{RLtext} \leavevmode\newline It looks like you sat in my chair, sir.\\\bottomrule
\end{tabular}%
}
\caption{Example models' responses to Egyptian dialect generation with their English translations. 
% English translations are generated using GPT-4o.
}
\label{tab-appx-dialect-gen-responses}
\end{table*}

\subsection{MT as Generation}

% \paragraph{English to Dialects} 
The statistics of test set  for the \textit{English-to-Dialect} and \textit{MSA-to-Dialects} splits 
are shown in Table \ref{tab:madar-en-to-dia-stats} and \ref{tab:msa-to-dia-stats}, respectively. The results are shown in Table \ref{tab:en-to-dia-final}.
For similar experiments on \textit{Dialects-to-English}, \textit{Dialects-to-MSA} and \textit{MSA-to-Dialects} the results are shown in Tables \ref{tab:dia-to-en-final}, \ref{tab:dia-to-msa-sacre-bleu} and \ref{tab:msa-to-dialects}, respectively. 

\begin{figure*}
     \centering
          \begin{subfigure}[b]{0.5\textwidth}
         \centering
         \includegraphics[width=\textwidth]{figures/legend.pdf}
     \end{subfigure}

     \centering
     \begin{subfigure}[b]{0.3\textwidth}
         \centering
        \includegraphics[width=\textwidth]{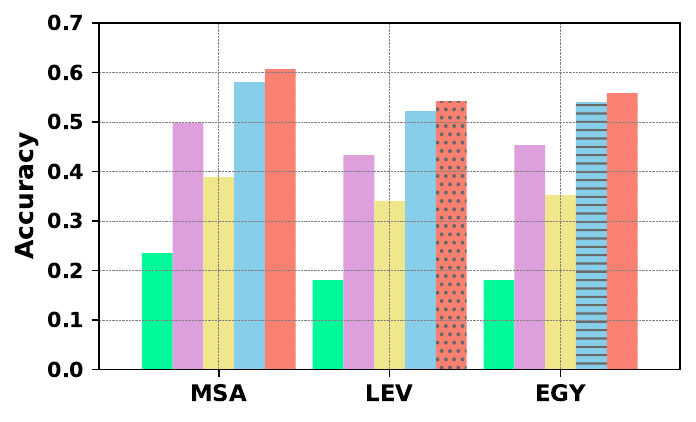}
         \caption{Humanities}
         \label{fig:arabic-mmlu-humanities-tr}
     \end{subfigure}
     \begin{subfigure}[b]{0.3\textwidth}
         \centering
         \includegraphics[width=\textwidth]{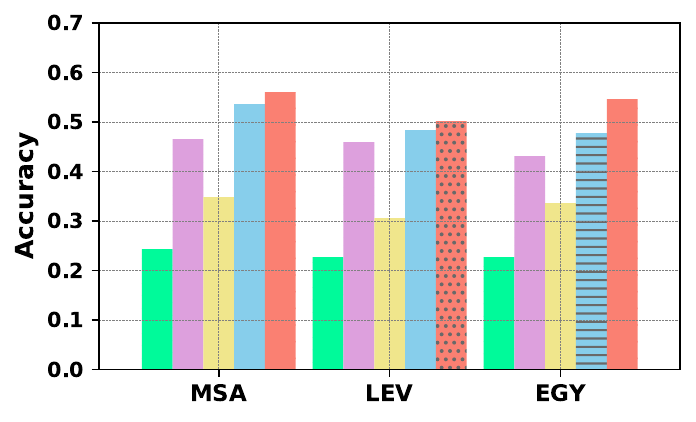}
         \caption{Language}
         \label{fig:arabic-mmlu-language-tr}
     \end{subfigure}
     \begin{subfigure}[b]{0.3\textwidth}
         \centering
         \includegraphics[width=\textwidth]{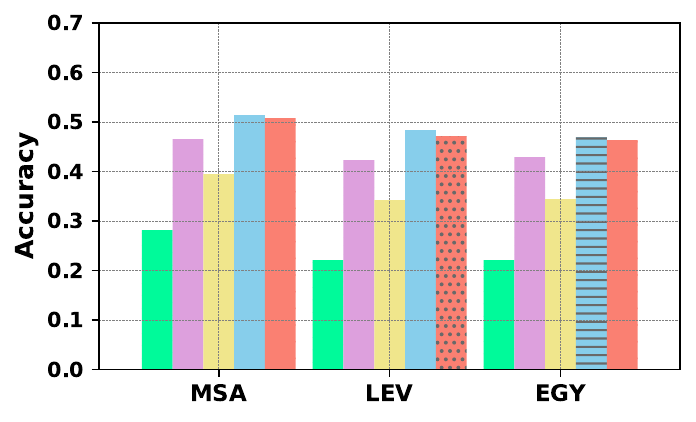}
         \caption{STEM}
         \label{fig:arabic-mmlu-stem-tr}
     \end{subfigure}
    \begin{subfigure}[b]{0.3\textwidth}
         \centering
         \includegraphics[width=\textwidth]{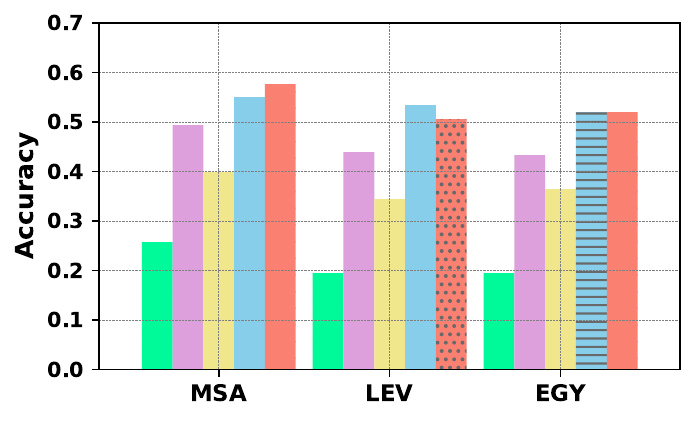}
         \caption{Social Science}
         \label{fig:arabic-mmlu-social-science-tr}
     \end{subfigure}
     \begin{subfigure}[b]{0.3\textwidth}
         \centering
         \includegraphics[width=\textwidth]{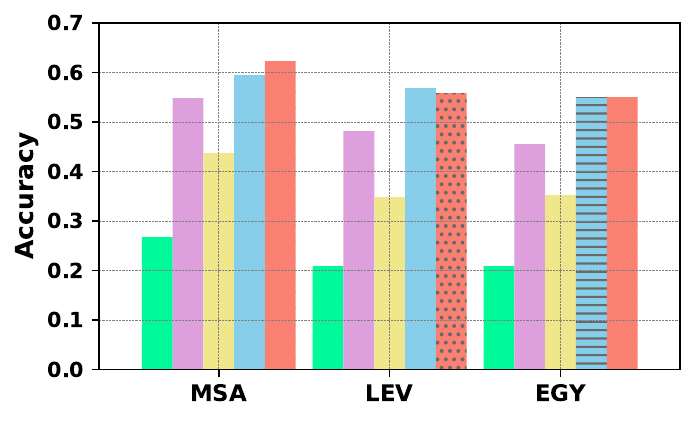}
         \caption{Others}
         \label{fig:arabic-mmlu-other-tr}
     \end{subfigure}
     \caption{Results on different subject areas in ArabicMMLU. %Arabic alphabets prompt.
     }
    \label{fig:arabic-mmlu-separate-categories-tr}
\end{figure*}

\begin{table}[h]
\centering
\scalebox{0.9}{% 
\begin{tabular}{@{}lr@{}}
\toprule
\multicolumn{1}{l}{\textbf{Test set}} & \multicolumn{1}{c}{\textbf{\# of Sent.}} \\ \midrule
madar-test-glf-0-iq-ar & 1,984 \\
madar-test-glf-0-om-ar & 1,979 \\
madar-test-glf-0-qa-ar & 1,935 \\
madar-test-glf-0-sa-ar & 1,974 \\
madar-test-glf-0-ye-ar & 1,983 \\
madar-test-glf-1-iq-ar & 1,985 \\
madar-test-glf-1-sa-ar & 1,975 \\
madar-test-glf-2-iq-ar & 1,985 \\\midrule
madar-test-lev-0-jo-ar & 1,975 \\
madar-test-lev-0-lb-ar & 1,966 \\
madar-test-lev-0-pa-ar & 1,972 \\
madar-test-lev-0-sy-ar & 1,968 \\
madar-test-lev-1-jo-ar & 2,008 \\
madar-test-lev-1-sy-ar & 1,978 \\\midrule
madar-test-nil-0-eg-ar & 1,965 \\
madar-test-nil-0-sd-ar & 1,981 \\
madar-test-nil-1-eg-ar & 1,982 \\
madar-test-nil-2-eg-ar & 1,976 \\ \midrule
madar-test-msa-0-ms-ar & 1,979 \\ \midrule
\textbf{Total} & \textbf{37,550} \\ \bottomrule
\end{tabular}
}
\caption{Number of sentences for the English and Dialect splits from the MADAR test set.}
\label{tab:madar-en-to-dia-stats}
\end{table}

\begin{table}[]
\centering
\scalebox{0.9}{%   
\begin{tabular}{@{}lr@{}}
\toprule
\multicolumn{1}{l}{\textbf{Test set}} & \multicolumn{1}{c}{\textbf{\# of Sent.}} \\ \midrule
madar-test-glf-0-iq & 1,973 \\
madar-test-glf-0-om & 1,971 \\
madar-test-glf-0-qa & 1,927 \\
madar-test-glf-0-sa & 1,965 \\
madar-test-glf-0-ye & 1,973 \\
madar-test-glf-1-iq & 1,973 \\
madar-test-glf-1-sa & 1,965 \\
madar-test-glf-2-iq & 1,973 \\ \midrule
madar-test-lev-0-jo & 1,966 \\
madar-test-lev-0-lb & 1,954 \\
madar-test-lev-0-pa & 1,963 \\
madar-test-lev-0-sy & 1,960 \\
madar-test-lev-1-jo & 1,997 \\
madar-test-lev-1-sy & 1,968 \\ \midrule
madar-test-nil-0-eg & 1,955 \\
madar-test-nil-0-sd & 1,970 \\
madar-test-nil-1-eg & 1,971 \\
madar-test-nil-2-eg & 1,965 \\ \midrule
\textbf{Total} & \textbf{35,389} \\ \bottomrule
\end{tabular}
}
\caption{Number of sentences for the MSA and Dialect splits from the MADAR test set.
% Data stats for the MSA-DIA splits.
}
\label{tab:msa-to-dia-stats}
\end{table}

\begin{table*}[h]
\centering
\scalebox{0.7}{
\begin{tabular}{@{}lrrrrrrr@{}}
\toprule
\textbf{Dataset} &
  \multicolumn{1}{l}{\textbf{NLLB-600M}} &
  \multicolumn{1}{l}{\textbf{NLLB-1.3B}} &
  \multicolumn{1}{l}{\textbf{NLLB-3.3B}} &
  \multicolumn{1}{l}{\textbf{Llama 3}} &
  \multicolumn{1}{l}{\textbf{Mistral}} &
  \multicolumn{1}{l}{\textbf{Jais}} &
  \multicolumn{1}{l}{\textbf{AceGPT}} \\ \midrule
madar-en-to-glf-iq-0 & 7.49  & 9.11  & 9.17  & 1.61 & 0.30 & 0.66 & 1.00 \\
madar-en-to-glf-iq-1 & 6.40  & 8.14  & 8.67  & 1.45 & 0.32 & 0.56 & 0.82 \\
madar-en-to-glf-iq-2 & 5.11  & 6.54  & 6.58  & 0.89 & 0.25 & 0.28 & 0.80 \\
madar-en-to-glf-om-0 & 15.91 & 16.15 & 16.01 & 5.66 & 1.28 & 2.27 & 3.42 \\
madar-en-to-glf-qa-0 & 9.86  & 11.74 & 14.11 & 2.25 & 0.47 & 1.09 & 1.38 \\
madar-en-to-glf-sa-0 & 14.90 & 19.73 & 22.60 & 3.68 & 0.95 & 1.48 & 2.21 \\
madar-en-to-glf-sa-1 & 8.56  & 11.25 & 13.38 & 2.16 & 0.40 & 0.82 & 1.15 \\
madar-en-to-glf-ye-0 & 7.60  & 9.40  & 10.14 & 1.74 & 0.30 & 0.65 & 1.09 \\ \midrule
smadar-en-to-lev-jo-0 & 10.65 & 15.46 & 19.41 & 1.84 & 0.24 & 0.90 & 2.06 \\
madar-en-to-lev-jo-1 & 10.88 & 15.21 & 20.17 & 1.60 & 0.24 & 0.96 & 1.80 \\
madar-en-to-lev-lb-0 & 5.19  & 7.59  & 8.59  & 0.82 & 0.18 & 0.49 & 1.19 \\
madar-en-to-lev-pa-0 & 10.19 & 14.43 & 17.92 & 1.71 & 0.29 & 1.12 & 1.87 \\
madar-en-to-lev-sy-0 & 9.03  & 13.43 & 15.29 & 1.30 & 0.22 & 0.98 & 2.76 \\
madar-en-to-lev-sy-1 & 7.64  & 11.34 & 12.96 & 1.37 & 0.24 & 0.95 & 3.12 \\ \midrule
madar-en-to-nil-eg-0 & 8.82  & 15.86 & 17.78 & 1.81 & 0.34 & 1.25 & 3.73 \\
madar-en-to-nil-eg-1 & 8.34  & 15.48 & 16.26 & 1.71 & 0.25 & 1.01 & 3.72 \\
madar-en-to-nil-eg-2 & 8.53  & 16.52 & 17.45 & 1.33 & 0.29 & 1.14 & 3.28 \\
madar-en-to-nil-sd-0 & 11.12 & 11.86 & 12.16 & 3.16 & 0.56 & 1.39 & 4.48 \\ \bottomrule
\end{tabular}}
\caption{Results (measured in SacreBLEU) for the English to Dialect systems.
% SacreBLEU Scores on all madar tests for En-to-dia direction
}
\label{tab:en-to-dia-final}
\end{table*}

\begin{table*}[h]
\centering
\scalebox{0.7}{
\begin{tabular}{@{}lrrrrrrr@{}}
\toprule
\textbf{Dataset} &
  \multicolumn{1}{l}{\textbf{NLLB-600M}} &
  \multicolumn{1}{l}{\textbf{NLLB-1.3B}} &
  \multicolumn{1}{l}{\textbf{NLLB-3.3B}} &
  \multicolumn{1}{l}{\textbf{Llama 3}} &
  \multicolumn{1}{l}{\textbf{Mistral}} &
  \multicolumn{1}{l}{\textbf{Jais}} &
  \multicolumn{1}{l}{\textbf{AceGPT}} \\ \midrule
madar-glf-to-en-iq-0 & 37.54 & 43.79 & 45.63 & 25.46 & 13.66 & 34.87 & 36.28 \\
madar-glf-to-en-iq-1 & 38.20 & 44.97 & 46.15 & 23.31 & 13.51 & 33.91 & 35.62 \\
madar-glf-to-en-iq-2 & 40.54 & 48.17 & 49.15 & 21.47 & 12.44 & 34.14 & 36.35 \\
madar-glf-to-en-om-0 & 44.87 & 49.62 & 48.77 & 33.07 & 19.66 & 41.43 & 43.33 \\
madar-glf-to-en-qa-0 & 36.58 & 42.70 & 43.80 & 24.16 & 13.53 & 34.86 & 35.82 \\
madar-glf-to-en-sa-0 & 47.33 & 52.68 & 52.70 & 32.37 & 21.79 & 40.99 & 44.34 \\
madar-glf-to-en-sa-1 & 36.61 & 42.89 & 44.73 & 23.42 & 12.03 & 32.62 & 33.93 \\
madar-glf-to-en-ye-0 & 39.41 & 47.08 & 47.97 & 22.32 & 13.70 & 36.30 & 36.81 \\ \midrule
madar-lev-to-en-jo-0 & 41.37 & 47.72 & 48.57 & 25.02 & 13.90 & 38.87 & 37.64 \\
madar-lev-to-en-jo-1 & 41.75 & 47.57 & 49.96 & 24.86 & 13.52 & 36.89 & 38.03 \\
madar-lev-to-en-lb-0 & 36.56 & 43.62 & 45.16 & 17.08 & 8.84  & 30.25 & 27.66 \\
madar-lev-to-en-pa-0 & 39.74 & 45.84 & 47.49 & 24.21 & 12.52 & 37.62 & 37.27 \\
madar-lev-to-en-sy-0 & 39.68 & 46.00 & 47.29 & 23.69 & 11.99 & 36.84 & 35.68 \\ \midrule
madar-nil-to-en-eg-0 & 37.83 & 44.17 & 46.73 & 24.08 & 12.26 & 36.84 & 35.68 \\
madar-nil-to-en-eg-1 & 46.50 & 53.31 & 55.08 & 28.92 & 14.40 & 43.50 & 43.03 \\
madar-nil-to-en-eg-2 & 37.72 & 44.52 & 46.89 & 23.14 & 11.18 & 35.64 & 36.20 \\
madar-nil-to-en-sd-0 & 46.32 & 52.54 & 52.00 & 31.43 & 17.74 & 44.78 & 44.34 \\ \bottomrule
\end{tabular}}
\caption{Results (measured in SacreBLEU) for the Dialects to English systems.
% SacreBLEU Scores for the DIA -> EN splits
}
\label{tab:dia-to-en-final}
\end{table*}

\begin{table*}[h]
\centering
\scalebox{0.7}{
\begin{tabular}{@{}lrrrrrrr@{}}
\toprule
\textbf{Dataset} &
  \multicolumn{1}{l}{\textbf{NLLB-600M}} &
  \multicolumn{1}{l}{\textbf{NLLB-1.3B}} &
  \multicolumn{1}{l}{\textbf{NLLB-3.3B}} &
  \multicolumn{1}{l}{\textbf{Llama}} &
  \multicolumn{1}{l}{\textbf{Mistral}} &
  \multicolumn{1}{l}{\textbf{Jais}} &
  \multicolumn{1}{l}{\textbf{AceGPT}} \\ \midrule
madar-dia-to-msa-glf-iq-0 & 22.30 & 26.84 & 63.65 & 4.11 & 0.14 & 6.45 & 3.97 \\
madar-dia-to-msa-glf-iq-1 & 21.58 & 26.43 & 63.50 & 3.93 & 0.14 & 6.19 & 2.85 \\
madar-dia-to-msa-glf-iq-2 & 22.15 & 27.70 & 65.47 & 3.92 & 0.07 & 6.55 & 2.92 \\
madar-dia-to-msa-glf-om-0 & 24.92 & 28.36 & 60.52 & 6.80 & 0.23 & 8.97 & 5.20 \\
madar-dia-to-msa-glf-qa-0 & 22.25 & 26.14 & 57.91 & 3.94 & 0.10 & 6.34 & 3.29 \\
madar-dia-to-msa-glf-sa-0 & 26.77 & 30.70 & 70.00 & 5.43 & 0.23 & 7.80 & 4.81 \\
madar-dia-to-msa-glf-sa-1 & 21.47 & 25.49 & 61.74 & 3.63 & 0.07 & 5.53 & 2.61 \\
madar-dia-to-msa-glf-ye-0 & 23.05 & 28.33 & 65.12 & 3.88 & 0.20 & 6.91 & 3.18 \\ \midrule
madar-dia-to-msa-lev-jo-0 & 23.82 & 28.74 & 65.96 & 4.68 & 0.11 & 6.32 & 3.72 \\
madar-dia-to-msa-lev-jo-1 & 24.36 & 29.36 & 66.90 & 4.57 & 0.09 & 5.94 & 3.82 \\
madar-dia-to-msa-lev-lb-0 & 21.44 & 26.73 & 61.69 & 2.94 & 0.05 & 5.28 & 2.34 \\
madar-dia-to-msa-lev-pa-0 & 23.60 & 28.61 & 62.28 & 4.04 & 0.13 & 6.24 & 3.64 \\
madar-dia-to-msa-lev-sy-0 & 24.17 & 28.85 & 61.60 & 4.64 & 0.10 & 6.95 & 3.19 \\
madar-dia-to-msa-lev-sy-1 & 23.25 & 28.17 & 62.43 & 3.33 & 0.08 & 6.29 & 2.50 \\ \midrule
madar-dia-to-msa-nil-eg-0 & 23.46 & 27.81 & 58.23 & 4.80 & 0.12 & 6.53 & 3.25 \\
madar-dia-to-msa-nil-eg-1 & 25.67 & 31.14 & 67.36 & 4.82 & 0.16 & 6.86 & 3.05 \\
madar-dia-to-msa-nil-eg-2 & 23.12 & 27.78 & 59.51 & 3.94 & 0.05 & 6.09 & 3.14 \\
madar-dia-to-msa-nil-sd-0 & 23.74 & 28.46 & 57.99 & 4.78 & 0.14 & 7.35 & 3.80 \\ \bottomrule
\end{tabular}}
\caption{Results (measured in SacreBLEU) for the Dialects to MSA systems.
%Dialects to MSA SacreBLEU scores
} 
\label{tab:dia-to-msa-sacre-bleu}
\end{table*}

\begin{table*}[h]
\centering
\scalebox{0.7}{
\begin{tabular}{@{}lrrrrrrr@{}}
\toprule
task &
  \multicolumn{1}{l}{\textbf{NLLB-600M}} &
  \multicolumn{1}{l}{\textbf{NLLB-1.3B}} &
  \multicolumn{1}{l}{\textbf{NLLB-3.3B}} &
  \multicolumn{1}{l}{\textbf{Llama 3}} &
  \multicolumn{1}{l}{\textbf{Mistral}} &
  \multicolumn{1}{l}{\textbf{Jais}} &
  \multicolumn{1}{l}{\textbf{AceGPT}} \\ \midrule
madar-msa-to-dia-glf-iq-0 & 7.74  & 10.83 & 11.70 & 1.46 & 0.42 & 1.17 & 2.10 \\
madar-msa-to-dia-glf-iq-1 & 7.46  & 11.15 & 11.39 & 1.23 & 0.29 & 0.75 & 1.60 \\
madar-msa-to-dia-glf-iq-2 & 5.38  & 7.25  & 8.07  & 0.77 & 0.31 & 0.52 & 1.28 \\
madar-msa-to-dia-glf-om-0 & 13.03 & 12.82 & 14.57 & 4.59 & 1.85 & 3.91 & 7.43 \\
madar-msa-to-dia-glf-qa-0 & 12.51 & 15.51 & 17.02 & 2.41 & 0.77 & 1.70 & 3.10 \\
madar-msa-to-dia-glf-sa-0 & 18.21 & 22.27 & 27.61 & 3.28 & 0.85 & 2.38 & 4.86 \\
madar-msa-to-dia-glf-sa-1 & 11.07 & 15.30 & 17.10 & 2.14 & 0.51 & 1.16 & 2.43 \\
madar-msa-to-dia-glf-ye-0 & 9.57  & 11.73 & 13.11 & 1.42 & 0.53 & 1.11 & 2.13 \\ \midrule
madar-msa-to-dia-lev-jo-0 & 12.38 & 16.51 & 22.79 & 1.33 & 0.27 & 1.71 & 3.79 \\
madar-msa-to-dia-lev-jo-1 & 13.13 & 16.89 & 23.02 & 1.33 & 0.31 & 1.25 & 3.23 \\
madar-msa-to-dia-lev-lb-0 & 6.70  & 9.69  & 10.37 & 0.63 & 0.22 & 0.74 & 1.67 \\
madar-msa-to-dia-lev-pa-0 & 12.75 & 16.35 & 20.24 & 1.54 & 0.44 & 1.69 & 3.31 \\
madar-msa-to-dia-lev-sy-0 & 10.85 & 14.55 & 17.62 & 1.43 & 0.26 & 1.37 & 3.37 \\
madar-msa-to-dia-lev-sy-1 & 9.08  & 12.20 & 15.40 & 1.23 & 0.24 & 1.03 & 2.63 \\ \midrule
madar-msa-to-dia-nil-eg-0 & 11.58 & 16.69 & 21.33 & 1.48 & 0.23 & 2.15 & 4.13 \\
madar-msa-to-dia-nil-eg-1 & 9.99  & 14.66 & 17.96 & 1.32 & 0.17 & 1.87 & 3.26 \\
madar-msa-to-dia-nil-eg-2 & 10.92 & 15.41 & 19.75 & 1.08 & 0.32 & 1.86 & 3.50 \\
madar-msa-to-dia-nil-sd-0 & 9.91  & 12.47 & 11.67 & 1.94 & 0.51 & 2.68 & 4.11 \\ \bottomrule
\end{tabular}}
\caption{Results (measured in SacreBLEU) for the MSA to dialectal systems.}
\label{tab:msa-to-dialects}
\end{table*}

%\begin{table}[]
%\centering
%\resizebox{0.8\columnwidth}{!}{%
%\begin{tabular}{@{}lcccc@{}}
%\toprule
%\textbf{Dataset} & \multicolumn{1}{c}{\textbf{Llama 3}} & \multicolumn{1}{c}{\textbf{Mistral}} & \multicolumn{1}{c}{\textbf{Jais}} & \multicolumn{1}{c}{\textbf{AceGPT}} \\ \midrule
%Gulf & 4.35 & 0.16 & 6.59 & 3.62 \\
%Levantine & 4.22 & 0.09 & 6.19 & 3.32 \\
%Egyptian & 4.58 & 0.12 & 6.56 & 3.31 \\ \bottomrule
%\end{tabular}%
%}
%\label{tab:dia-to-msa-averaged}
%\caption{Averaged results for the dialectal to MSA systems across models.}
%\end{table}

\section{Detailed Results on Cognitive Abilities}
\label{sec:app_cognitive_abilities} 

\subsection{World Knowledge }
\label{sec:app_world_knowledge} 

The results in \ref{tab:arabic-mmlu-overall-results} The results show the overall Arabic MMLU accuracies. in Figure \ref{fig:arabic-mmlu-separate-categories-tr} provide key insights into the performance of different models on MSA and dialectal datasets (Egyptian and Levantine). Below we provide some key observations.

\paragraph{Humanities:} AceGPT leads in all three categories (MSA, Egyptian, Levantine), slightly outperforming Jais, another Arabic-centric model. This suggests that AceGPT excels in humanities-related questions across dialects. Llama and Mistral, both non-Arabic-centric models, lag behind, with a significant performance gap between them and the Arabic-centric models.

\paragraph{STEM:} Jais outperforms AceGPT in MSA, but the gap narrows in the dialects, particularly in Levantine, where AceGPT demonstrates competitive performance. Mistral and Llama perform comparatively lower in STEM subjects, suggesting they may not generalize as well in technical disciplines, especially when dialects are involved.

\paragraph{Social Sciences:} Similar to the humanities, AceGPT shows consistent superiority over other models. However, the overall performance drops slightly in dialects compared to MSA. Mistral lags behind the other models, particularly in Levantine dialect, where the gap becomes more pronounced.

\paragraph{Language Tasks:} Language tasks show an interesting pattern where AceGPT and Jais outperform Llama and Mistral by a wide margin, especially in Levantine. This is likely due to the language-specific capabilities of AceGPT and Jais, which are more attuned to Arabic dialects. Levantine seems to pose a greater challenge across models, as the performance declines are slightly more pronounced.

\paragraph{Others (Miscellaneous):} This category reveals AceGPT’s broad capabilities as it outperforms other models in both dialects and MSA. Notably, Jais also performs well, but AceGPT maintains a lead. Llama’s performance is competitive in MSA but suffers a sharper decline in dialectal benchmarks, particularly in Egyptian.

\begin{table}[]
\centering
\begin{tabular}{lrrr}
\toprule
        & \multicolumn{1}{l}{\textbf{MSA}} & \multicolumn{1}{l}{\textbf{LEV}} & \multicolumn{1}{l}{\textbf{EGY}} \\ \midrule
Rand    & 0.26                    & 0.20                    & 0.20                    \\
LLama 3 & 0.49                    & 0.44                    & 0.44                    \\
Mistral & 0.40                    & 0.34                    & 0.35                    \\
Jais    & 0.56                    & 0.52                    & 0.51                    \\
Ace-GPT & 0.57                    & 0.51                    & 0.52                   \\ \bottomrule
\end{tabular}
\caption{Results on ArabicMMLU.}
\label{tab:arabic-mmlu-overall-results}
\end{table}

\begin{figure}
     \centering
         \includegraphics[width=0.75\linewidth]{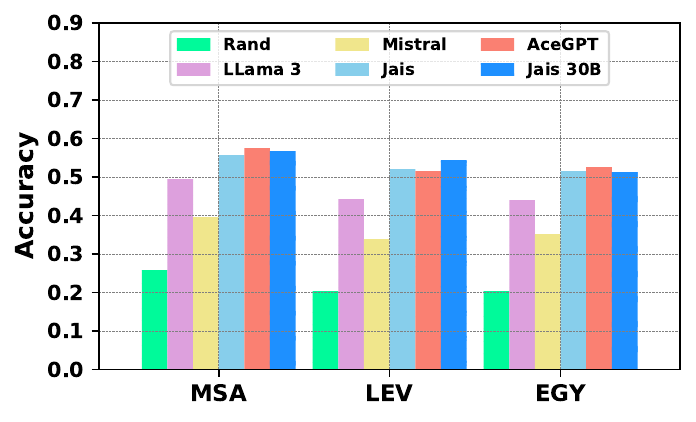}
         % \vspace{-0.3cm}
         \caption{Average results on ArabicMMLU}
         \vspace{-0.3cm}
         \label{fig:avg-scores-arabic-mmlu-ar-jais30B}
\end{figure}

\subsection{Reading Comprehension} 

The Accuracy numbers for BoolQ and Belebele are shown in tables \ref{tab:boolQ-results} and \ref{tab:belebele-results}

\begin{table}[]
\centering
\begin{tabular}{@{}lrrrr@{}}
\toprule
 & \multicolumn{1}{l}{\textbf{MSA}} & \multicolumn{1}{l}{\textbf{LEV}} & \multicolumn{1}{l}{\textbf{EGY}} & \multicolumn{1}{l}{\textbf{ENG}} \\ \midrule
Random  & 0.38 & 0.38 & 0.38 & 0.38 \\
LLama 3 & 0.74 & 0.71 & 0.73 & 0.85 \\
Mistral & 0.66 & 0.63 & 0.63 & 0.85 \\
Jais    & 0.71 & 0.65 & 0.64 & 0.75 \\
AceGPT & 0.77 & 0.68 & 0.69 & 0.82 \\ \bottomrule
\end{tabular}
\caption{Results on BoolQ.}
\label{tab:boolQ-results}
\end{table}

\begin{table}[]
\centering
\begin{tabular}{@{}lrrrr@{}}
\toprule
 & \multicolumn{1}{l}{\textbf{MSA}} & \multicolumn{1}{l}{\textbf{LEV}} & \multicolumn{1}{l}{\textbf{EGY}} & \multicolumn{1}{l}{\textbf{ENG}} \\ \midrule
Random    & 0.26 & 0.26 & 0.26 & 0.23 \\
Llama 3  & 0.49 & 0.42 & 0.39 & 0.87 \\
Mistral & 0.41 & 0.36 & 0.35 & 0.75 \\
Jais    & 0.57 & 0.49 & 0.49 & 0.80 \\
AceGPT  & 0.49 & 0.43 & 0.42 & 0.83 \\ \bottomrule
\end{tabular}
\caption{Results on Belebele.}
\label{tab:belebele-results}
\end{table}

\subsection{Commonsense Reasoning}

The results for PiQA, Openbooks, and Winogrande are shown in tables \ref{tab:piqa-results}, \ref{tab:OpenbookQA-results}, and \ref{tab:winogrande-results}

\begin{table}[]
\centering
\begin{tabular}{@{}lrrrr@{}}
\toprule
 & \multicolumn{1}{l}{\textbf{MSA}} & \multicolumn{1}{l}{\textbf{LEV}} & \multicolumn{1}{l}{\textbf{EGY}} & \multicolumn{1}{l}{\textbf{ENG}} \\ \midrule
Random    & 0.48 & 0.49 & 0.49 & 0.50 \\
Llama 3  & 0.55 & 0.54 & 0.55 & 0.78 \\
Mistral & 0.55 & 0.52 & 0.54 & 0.81 \\
Jais    & 0.66 & 0.59 & 0.62 & 0.77 \\
AceGPT  & 0.64 & 0.57 & 0.61 & 0.80 \\ \bottomrule
\end{tabular}
\caption{Results on PIQA.}
\label{tab:piqa-results}
\end{table}
\begin{table}[]
\centering
\begin{tabular}{@{}lrrrr@{}}
\toprule
 & \multicolumn{1}{l}{\textbf{MSA}} & \multicolumn{1}{l}{\textbf{LEV}} & \multicolumn{1}{l}{\textbf{EGY}} & \multicolumn{1}{l}{\textbf{ENG}} \\ \midrule
Random    & 0.28 & 0.26 & 0.31 & 0.27 \\
Llama 3  & 0.32 & 0.33 & 0.32 & 0.42 \\
Mistral & 0.31 & 0.30 & 0.30 & 0.45 \\
Jais    & 0.35 & 0.34 & 0.32 & 0.41 \\
AceGPT  & 0.36 & 0.34 & 0.32 & 0.45 \\ \bottomrule
\end{tabular}
\caption{Results on OBQA.} 
\label{tab:OpenbookQA-results}
\end{table}
\begin{table}[]
\centering
\begin{tabular}{@{}lrrr@{}}
\toprule
        & \multicolumn{1}{l}{\textbf{MSA}} & \multicolumn{1}{l}{\textbf{LEV}} & \multicolumn{1}{l}{\textbf{EGY}} \\ \midrule
Random    & 0.51                    & 0.51                    & 0.49                    \\
Llama 3  & 0.60                    & 0.52                    & 0.56                    \\
Mistral & 0.57                    & 0.51                    & 0.55                    \\
Jais    & 0.65                    & 0.58                    & 0.61                    \\
AceGPT  & 0.61                    & 0.58                    & 0.63                   \\ \bottomrule
\end{tabular}
\caption{Results on Winogrande.}
\label{tab:winogrande-results}
\end{table}

\subsection{Misinformation}

The results for truthfulQA are shown in table \ref{tab:truthfulQA-results}

\begin{table}[]
\centering
\begin{tabular}{lrrrr}
\toprule
 & \multicolumn{1}{l}{\textbf{MSA}} & \multicolumn{1}{l}{\textbf{LEV}} & \multicolumn{1}{l}{\textbf{EGY}} & \multicolumn{1}{l}{\textbf{ENG}} \\ \midrule
Random  & 0.21 & 0.23 & 0.20 & 0.39 \\ 
Llama 3 & 0.34 & 0.29 & 0.33 & 0.40 \\
Mistral & 0.33 & 0.33 & 0.30 & 0.57 \\
Jais    & 0.28 & 0.28 & 0.31 & 0.28 \\
AceGPT & 0.34 & 0.30 & 0.33 & 0.37 \\ \bottomrule
\end{tabular}
\caption{Results on TruthfulQA.}
\label{tab:truthfulQA-results}
\end{table}

\section{Examples on Models' Cultural Knowledge}
\label{sec:appendix:culture-understanding}
In Section~\ref{sec:cultural-understanding-results}, we presented results on models' performance on the task of  region-specific cultural awareness, where we modeled the task as a MCQ task. Although such setup can provide an overall performance estimation, deeper understanding of an LLM behavior capturing fine-grained cultural differences is essential. 

In Table~\ref{tab-appx-cultural-responses} we present examples of the models' responses to dialectal cultural questions.
With a simple geography question related to one of the Levant countries (\textbf{Q1}), we observe that Jais and AceGPT respond with the correct answer, while Llama 3 partially answer the question providing  additional irrelevant and incorrect information. Mistral shows a different behavior, hallucinating and responding with non-existent geographical locations. 
Similar trends can be observed with \textbf{Q2} about one of the traditions of a Gulf country, where Jais and AceGPT both correctly answer the question, and Llama 3 returns an answer that is more of a description or explanation. Mistral's response is particularly worth noting, as not only it generates fictional information, it  generated an answer describing a dance where men and women are dancing together, although anyone with some awareness of the country's tradition would know such dance does not comply with the culture of Qatar where segregation between men and women is the norm.

\begin{table*}[h]
\centering
\resizebox{\textwidth}{!}{%    
\begin{tabular}{l|p{2.5cm}|p{2.2cm}|p{3cm}|p{3cm}|p{3cm}|p{3cm}}
\toprule
\textbf{\#} & \textbf{Question} & {\textbf{Gold Answer}} & \textbf{Llama 3} & \textbf{Mistral} & \textbf{Jais} & \textbf{AceGPT}\\\midrule
\textbf{Q1} & \begin{RLtext}\footnotesizeشو هي البحار يلي بتطل عليها الأردن؟\end{RLtext} \leavevmode\newline What are the seas that Jordan overlooks? & \begin{RLtext}\footnotesizeالبحر الأحمر، البحر الميت\end{RLtext} \leavevmode\newline The Red Sea, the Dead Sea & \begin{RLtext}\footnotesizeالبحار التي بتطل عليها الأردن هي البحر الميت واليمن والخليج العربي.'\end{RLtext} \leavevmode\newline The seas that Jordan overlooks are the Dead Sea, Yemen and the Arabian Gulf. & The Red Sea \begin{RLtext}\footnotesizeوجهة البحر الأحمر، السويدانية وبحر العدد، المصرية.'\end{RLtext} \leavevmode\newline The Red Sea: The Swedish destination and the Egyptian Sea of Numbers. & \begin{RLtext}\footnotesizeالبحر الأحمر, البحر الميت'\end{RLtext} \leavevmode\newline The Red Sea, the Dead Sea & \begin{RLtext}\footnotesizeالبحر الميت، البحر الأحمر.'\end{RLtext} \leavevmode\newline The Dead Sea, the Red Sea\\\midrule
\textbf{Q2} & \begin{RLtext}\footnotesizeشنو الرقصة الشعبية التقليدية في قطر؟\end{RLtext} \leavevmode\newline What is the traditional folk dance in Qatar? &  \begin{RLtext}\footnotesizeالعرضة \end{RLtext} \leavevmode\newline Ardah & \begin{RLtext}\footnotesizeالرقصة الشعبية التقليدية في قطر هي الرقصة القطريّة.' \end{RLtext} \leavevmode\newline The traditional folk dance in Qatar is the Qatari dance.  
& \begin{RLtext}\footnotesizeفي قطر، يتم الرقص الشعبي التقليدي بالدفنة الوسطى والحلقة الشاملة، بواسطة النساء والرجال مجمعين.  \end{RLtext} \leavevmode\newline In Qatar, traditional folk dancing is performed with the central drum (dufna) and the circular dance (halqa shamelah), with both women and men participating together. & \begin{RLtext}\footnotesizeالعرضة \end{RLtext} \leavevmode\newline Ardah & \begin{RLtext}\footnotesizeالرقصة الشعبية التقليدية في قطر هي العرضة.' \end{RLtext} \leavevmode\newline The traditional folk dance in Qatar is the Ardah.\\\bottomrule
\end{tabular}
}
\caption{Example models' responses to dialectal cultural questions with their  English translations.
% are generated using GPT-4o.
}
\label{tab-appx-cultural-responses}
\end{table*}                                          
\section{Detailed PEMT Guideline}
\label{sec:app_post_editing_guideline}
% \firoj{TO DO: https://docs.google.com/document/d/1yS8ZWy28UEVtq-HXFoN5Vhsx2jpKBnirqD5LWcIUR3w/edit}
% Data samples from different datasets are given. 
The purpose of PEMT ~\cite{brockmann-etal-2022-error} is to refine and improve the output generated by MT systems to ensure accuracy, fluency (i.e. they reflect the nuances of how the dialect is spoken), adequacy (i.e. they maintain the semantic meaning of the input sentence), and cultural appropriateness. For the PEMT task, the goal was to post-edit datasets translated into MSA and the dialects: Levantine and Egyptian. Below we provide detailed instructions for PEMT. Unless stated, all examples are to illustrate a specific instruction and might not reflect datasets or dialect. 

% For the post-editing task on the datasets used in this study, we first machine translated into Modern Standard Arabic, Levantine Arabic, and Egyptian Arabic. 
% The task is to post correct translations so that they are fluent (i.e. they reflect the nuances of how the dialect is spoken), and adequate (i.e. they maintain the semantic meaning of the input sentence). Below are the dataset descriptions and the instructions that need to be followed. 

\subsection{General instructions for PEMT:}

\textbf{Instruction 1:}  When necessary, the text should be rewritten to adhere to the syntactic and semantic structure of Arabic. Some questions or options may need to be paraphrased to ensure the fluency. To achieve this, the edits may include:

% to the grammatical rules,  of Arabic. It should be edited to conform to the 

\begin{itemize}[noitemsep,topsep=0pt,leftmargin=*,labelsep=.5em]
    \item Paraphrasing questions or options.
    \item Editing or adding any parts-of-speech needed to improve the readability of the sentence, such as adding %linking 
    words \<\tiny(حروف العطف)> or prepositions \<\tiny(حروف الجر)> or adding the definite article \<\tiny(ال التعريف)>.
    \item For MSA, it is also important to correct the wrong use of Hamzah (Hamzatul-wasl and Hamzatul-Qata’a).
    \item Fixing the use of inappropriate verbs that do not align between sentences and the options associated with them in terms of their grammatical gender and singularity or plurality.
    \item Correcting wrong translation of words and phrases, or adding missing information that were not transferred from the source sentence to the translation.
\end{itemize}

In Table \ref{tab-appx-instruction1}, we provide some examples to illustrate the above points.

\begin{table}[h]
\centering
\resizebox{\columnwidth}{!}{%   
\begin{tabular}{p{3cm}|p{3cm}|p{3cm}}
\toprule
\textbf{English} & \textbf{MT (MSA)}& \textbf{PEMT (MSA)}\\ 
\midrule
People in the US have longer paid vacations than people in which Western countries? & \begin{RLtext}\footnotesizeالناس في الولايات المتحدة لديهم إجازات مدفوعة الأجر أطول من الأشخاص الذين يعيشون في الدول الغربية؟\end{RLtext} & \begin{RLtext}\footnotesizeهل يحصل الأشخاص في الولايات المتحدة إجازات مدفوعة الأجر أطول من الأشخاص الذين يعيشون في الدول الغربية؟\end{RLtext} \\ 
\midrule
Who objectively makes the best burgers? & \begin{RLtext}\footnotesizeمن الذي يصنع أفضل البرغر بشكل موضوعي؟\end{RLtext} & \begin{RLtext}\footnotesizeدون أي تحيز، من الذي يعد أفضل برغر؟\end{RLtext} \\ 
\midrule
Who famously said, ``I cannot tell a lie''? & \begin{RLtext}\footnotesizeومن قال عبارته الشهيرة ''لا أستطيع أن أكذب``؟\end{RLtext} & \begin{RLtext}\footnotesizeمن قائل العبارة الشهيرة: "لا أستطيع أن أكذب"؟\end{RLtext} \\ 
\bottomrule
\end{tabular}%
}
\caption{Example sentences in English, MT (MSA), and post-edited versions to illustrate \textit{Instruction 1}.}
\label{tab-appx-instruction1}
\end{table}

\textbf{Instruction 2:} For abbreviations, if they have a translation for them in Arabic that is common and well known, the Arabic translation for them should be used. If the translation of the abbreviation is not known in Arabic, the abbreviation in English letters should be used. Examples are shown in Table \ref{tab-appx-instruction2}. 

\begin{table}[htbp]
\centering
\resizebox{0.65\columnwidth}{!}{% 
\begin{tabular}{p{2.5cm}|p{2.5cm}}
\toprule
\textbf{English} & \textbf{PEMT (MSA)}\\ 
\midrule
What did CERN do in 2012?& \begin{RLtext}\footnotesizeماذا فعلت 
\end{RLtext} CERN\begin{RLtext}\footnotesize
 في عام 2012؟\end{RLtext}\\ 
\midrule
I am an AI and I don't know the answer & \begin{RLtext}\footnotesizeأنا ذكاء اصطناعي ولا أعرف الإجابة\end{RLtext} \\ 
\bottomrule
\end{tabular}%
}
\centering
\caption{Example sentences in English and Arabic translation to illustrate \textit{Instruction 2}.}
\label{tab-appx-instruction2}
\end{table}

\textbf{Instruction 3:}
Samples might reference names, such as those of people, places, organizations, creatures, movies, TV series, books, songs, and model of products. These parts should be transliterated into Arabic. This means writing the English word as it is into Arabic script. However, this does not apply to names that have an equivalent translation in Arabic.  For instance, the example sentences in Table \ref{tab-appx-instruction3}.

\begin{table}[h!]
\scalebox{0.8}{%  
\renewcommand{\arraystretch}{1.2}
\begin{tabular}{p{0.3\linewidth}|p{0.35\linewidth}|p{0.35\linewidth}}
\toprule
\textbf{English} & \textbf{MT (MSA)}& \textbf{PEMT (MSA)}\\
\midrule
Chupacabras turned out to be real. & \begin{RLtext}\footnotesizeتبين أن Chupacabras حقيقي.\end{RLtext} & \begin{RLtext}\footnotesizeتبين أن تشوباكا حقيقي.\end{RLtext} \\\midrule
 Who composed the tune of ``Twinkle, Twinkle, Little Star''?& \begin{RLtext}\footnotesizeمن قام بتأليف لحن أغنية ''توينكل، توينكل، ليتل ستار``؟\end{RLtext}&\begin{RLtext}\footnotesizeمن قام بتأليف لحن أغنية \end{RLtext} ``Twinkle, Twinkle, Little Star" ?\\\midrule
 The yellow-billed cuckoo (Coccyzus americanus)& \begin{RLtext}\footnotesizeالوقواق ذو المنقار الأصفر
\end{RLtext} (Coccyzus americanus) &
\begin{RLtext}\footnotesizeالوقواق ذو المنقار الأصفر
\end{RLtext} (Coccyzus americanus)\\\midrule
 Is WordPad the same thing as Microsoft Word?& \begin{RLtext}\footnotesizeهل برنامج 
\end{RLtext}Wordpad
\begin{RLtext}\footnotesize
 هو نفس برنامج
\end{RLtext}
 Microsoft Word&
\begin{RLtext}\footnotesizeهل برنامج 
\end{RLtext}Wordpad
\begin{RLtext}\footnotesize
 هو نفس برنامج
\end{RLtext}
 Microsoft Word? \\ \bottomrule
\end{tabular}
}
\caption{Example sentences containing names in English and Arabic translation.}
\label{tab-appx-instruction3}
\end{table}

% This point does not apply to names that have an equivalent translation for it in Arabic.  

% \textbf{Instruction 4:} Some references to aid with the work, if needed: 
% \url{https://www.google.com/}
% \url{https://translate.google.com/?sl=auto&tl=en&op=translate} 
% \url{https://context.reverso.net/translation/}

\subsection{Datasets, Task Format, and Specific Instructions for Each Dataset:}

\subsubsection{ArabicMMLU}

\textbf{Data format:}
 ArabicMMLU dataset is originally in Arabic (MSA), therefore, MT and PEMT were only done for Levantine and Egyptian. The translator's task consisted of editing: \textit{(i)} one question and \textit{(ii)} four options that serve as possible answers for that question. 
% \firoj{The ArabicMMLU dataset is originally in MSA, therefore, this guideline is prepared for MSA and Egyptian translation.}

\textbf{Specific Instruction:} 

\begin{itemize}[noitemsep,topsep=0pt,leftmargin=*,labelwidth=!,labelsep=.5em]
    \item  It is important to consider the structure of the question, which needs to be maintained.  Below are some examples that illustrates different types of questions. To ensure the fluency of the questions, they might need paraphrasing.
    \item Not all parts of the questions and the options should be expressed in dialect. Changing those parts or expressing them in dialect will affect the meaning of the question. These parts include \textit{(i)} Verses from Quran, or lines from Hadith, \textit{(ii)} Words or the phrases that the question is asking the meaning of, \textit{(iii)} Words/ phrases that refer to scientific domains, \textit{(iv)} Technical words or phrases, \textit{(v)} Lines from poems, \textit{(vi)} Equations (chemistry or mathematical). While editing, avoid changing these parts, and only edit the other parts that should be expressed in dialect and will not change the semantic meaning of the question. However, edits can be made to address any missing parts or grammatical errors that the question might have.
    \item Blanks should be represented as it appears in the MSA (source). If it is represented as dots, then they should remain the same. %represent it as dots. 
    If it is represented as {\tiny\<[فراغ]>}, then add it as such.
\end{itemize}

% Below are some examples: 
Below are a few examples from different disciplines and specific instructions for scenarios. We show the questions and answers (options). We provide the source (MSA), MT, and PEMT Levantine for illustration purposes. However, for Egyptian, the instructions are the same for the dialectal translation. For each example, we provide an explanation highlighting the reason for the edits. 

\noindent
\textbf{1. Islamic Studies}

\vspace{0.2cm}
\noindent
\textbf{Question (MSA):} 
\vspace{0.2cm}
\begin{RLtext}\footnotesizeقال تعالى في سورة العاديات: ( فالمغيرات صبحا (3)) ماالمقصود بالمغيرات؟\end{RLtext}

\noindent
\textbf{Options (MSA):}

\renewcommand{\arraystretch}{0.1} 
\begin{tabular}{p{0.8\linewidth}l}
    \begin{RLtext}\footnotesizeالرياح الشديدة\end{RLtext} & 1. \\
    \begin{RLtext}\footnotesizeالابل\end{RLtext} & 2. \\
    \begin{RLtext}\footnotesizeالملائكة السابحين\end{RLtext} & 3. \\
    \begin{RLtext}\footnotesizeالخيل\end{RLtext} & 4. \\
\end{tabular}

% Source text: Options in Levantine
% Question in Lev: 

\noindent
\vspace{0.2cm}
\textbf{Question (PEMT -Levantine):}\begin{RLtext}\footnotesizeقال تعالى في سورة العاديات: ( فالمغيرات صبحا (3)) شو يعني المغيرات؟\end{RLtext}
\noindent

\textbf{Options (PEMT -Levantine):}

\renewcommand{\arraystretch}{0.1} 
\begin{tabular}{p{0.8\linewidth}l}
    \begin{RLtext}\footnotesizeالهوا القوي\end{RLtext} & 1. \\
    \begin{RLtext}\footnotesizeالابل\end{RLtext} & 2. \\
    \begin{RLtext}\footnotesizeالملائكة السابحين\end{RLtext} & 3. \\
    \begin{RLtext}\footnotesizeالخيل\end{RLtext} & 4. \\
\end{tabular}

\textbf{Explanation}: Verses from Qur'an should not be modified.  Any changes to such elements will alter the meaning of the question. However, in this case, the options can be expressed differently in the dialect.

\vspace{0.2cm}
\noindent
\textbf{2. Accounting }

\vspace{0.2cm}
\noindent
\textbf{Question (MSA):}
\vspace{0.2cm}
\begin{RLtext}\footnotesizeفرع محاسبي يهدف إلى التحقق من صحة وسلامة المعلومات المالية بما يضفى مزيدًا من الثقة على هذه المعلومات\end{RLtext}

\vspace{0.2cm}
\noindent
\textbf{Options (MSA):}

\renewcommand{\arraystretch}{0.1}
\begin{tabular}{p{0.8\linewidth}l}
    \begin{RLtext}\footnotesizeالمحاسبة الحكومية\end{RLtext} & 1. \\
    \begin{RLtext}\footnotesizeالمحاسبة الدولية\end{RLtext} & 2. \\
    \begin{RLtext}\footnotesizeالمراجعة\end{RLtext} & 3. \\
    \begin{RLtext}\footnotesizeالمحاسبة الاجتماعية\end{RLtext} & 4. \\
\end{tabular}

% Source text:
% \begin{RLtext}\footnotesizeOptions in Levantine\end{RLtext}
% Question in Lev:

\vspace{0.2cm}
\noindent
\textbf{Question (PEMT -Levantine):}
\vspace{0.2cm}
\begin{RLtext}\footnotesizeفرع محاسبي بيهدف للتحقق من من صحة وسلامة المعلومات المالية وبالتالي بزيد الثقة من هيدي المعلومات\end{RLtext}
\noindent
\textbf{Options (PEMT -Levantine):}

\renewcommand{\arraystretch}{0.1} 
\begin{tabular}{p{0.8\linewidth}l}
    \begin{RLtext}\footnotesizeالمحاسبة الحكومية\end{RLtext} & 1. \\
    \begin{RLtext}\footnotesizeالمحاسبة الدولية\end{RLtext} & 2. \\
    \begin{RLtext}\footnotesizeالمراجعة\end{RLtext} & 3. \\
    \begin{RLtext}\footnotesizeالمحاسبة الاجتماعية\end{RLtext} & 4. \\
\end{tabular}

\textbf{Explanation}:
The names of knowledge domains and sub fields (the options) should remain the same. Changes should only be made to correct spelling mistakes in these words and phrases.

\vspace{0.2cm}
\noindent
\textbf{3. Arabic Language}
\vspace{0.2cm}

\noindent
\textbf{Question (MSA):}
\vspace{0.2cm}

\begin{RLtext}\footnotesizeالكلمة التي تحتوي على حرف السين(س) في آخر الكلمة:\end{RLtext}

\vspace{0.2cm}
\noindent
\textbf{Options (MSA)}:

\renewcommand{\arraystretch}{0.1} 
\begin{tabular}{p{0.8\linewidth}l}
    \begin{RLtext}\footnotesizeفرس\end{RLtext} & 1. \\
    \begin{RLtext}\footnotesizeسلمى\end{RLtext} & 2. \\
    \begin{RLtext}\footnotesizeمسلم\end{RLtext} & 3. \\
\end{tabular}

\vspace{0.2cm}
\noindent
\textbf{Question (PEMT -Levantine):}

\begin{RLtext}\footnotesizeالكلمة اللي بآخرها حرف السين(س):\end{RLtext}

\vspace{0.2cm}
\noindent
\textbf{Options (PEMT -Levantine)}:

\renewcommand{\arraystretch}{0.1} 
\begin{tabular}{p{0.8\linewidth}l}
    \begin{RLtext}\footnotesizeفرس\end{RLtext} & 1.\\
    \begin{RLtext}\footnotesizeفرس\end{RLtext} & 2.\\
    \begin{RLtext}\footnotesizeمسلم\end{RLtext} & 3.\\
\end{tabular}

\textbf{Explanation}: The options should not be changed to avoid affecting the meaning of the question.

\vspace{0.2cm}
\noindent
\textbf{4. Arabic Language (Grammar)}
\vspace{0.2cm}

\noindent
\textbf{Question (MSA):}
\vspace{0.2cm}

\begin{RLtext}\footnotesizeاختر حرف الجر المناسب لوضعه بدلاً من [فراغ] في الجملة التالية:  أنا [فراغ] باكستان.\end{RLtext}

\vspace{0.2cm}
\noindent
\textbf{Options (MSA)}:

\renewcommand{\arraystretch}{0.1} 
\begin{tabular}{p{0.8\linewidth}l}
    \begin{RLtext}\footnotesizeب\end{RLtext} & 1.\\
    \begin{RLtext}\footnotesizeإلى\end{RLtext} & 2.\\
    \begin{RLtext}\footnotesizeمن\end{RLtext} & 3.\\
    \begin{RLtext}\footnotesizeفي\end{RLtext} & 4.\\
\end{tabular}

\vspace{0.2cm}
\noindent
\textbf{Question (PEMT -Levantine)}:

\begin{RLtext}\footnotesizeنقي حرف الجر المناسب لتحطته مطرح [فراغ] بهيدي الجملة:  أنا [فراغ] باكستان.\end{RLtext}

\vspace{0.2cm}
\noindent
\textbf{Options(PEMT -Levantine)}:
\vspace{0.2cm}

\renewcommand{\arraystretch}{0.1} 
\begin{tabular}{p{0.8\linewidth}l}
    \begin{RLtext}\footnotesizeب\end{RLtext} & 1.\\
    \begin{RLtext}\footnotesizeإلى\end{RLtext}& 2. \\
    \begin{RLtext}\footnotesizeمن\end{RLtext} & 3.\\
    \begin{RLtext}\footnotesizeفي\end{RLtext} & 4.\\
\end{tabular}

% \textbf{Explanation:} The sentence to be completed should be kept the same. Options should not be edited.

\vspace{0.2cm}
\noindent
\textbf{5. Biology}
\vspace{0.2cm}

\noindent
\textbf{Question (MSA):}
\vspace{0.2cm}

\begin{RLtext}\footnotesizeأي من الأت مٌثل الكرموسومات الأنثو ةٌ ؟\end{RLtext}

\vspace{0.2cm}
\noindent
\textbf{Question (PEMT -Levantine):}
\vspace{0.2cm}

\begin{RLtext}\footnotesizeاي احتمال بمثل كرموسومات الأنثوة؟\end{RLtext}

\vspace{0.2cm}
\noindent
\textbf{Options}:

\begin{enumerate}
    \item B
    \item YX
    \item XX
    \item YY
\end{enumerate}

\textbf{Explanation:} For cases like the one above, the options should not be edited and should remain in English letters.

\vspace{0.2cm}
\noindent
\textbf{6. Civic}
\vspace{0.2cm}

\noindent
\textbf{Question (MSA):}
\vspace{0.2cm}

\begin{RLtext}\footnotesizeالمؤسسات الدستورية بالأردن هي:\end{RLtext}

\vspace{0.2cm}
\noindent
\textbf{Options (MSA)}:

\renewcommand{\arraystretch}{0.1}
\begin{tabular}{p{0.8\linewidth}l}
    \begin{RLtext}\footnotesizeالتنفيذية والتشريعية والقضائية\end{RLtext} & 1. \\
    \begin{RLtext}\footnotesizeالتنفيذية والصحافة والتشريعية\end{RLtext} & 2. \\
    \begin{RLtext}\footnotesizeالصحافة والتشريعية والقضائية\end{RLtext} & 3. \\
    \begin{RLtext}\footnotesizeالتنفيذية والتشريعية والدينية\end{RLtext} & 4. \\
\end{tabular}

\vspace{0.2cm}
\noindent
\textbf{Question (PEMT -Levantine)}:

\vspace{0.2cm}
\begin{RLtext}\footnotesizeالمؤسسات الدستورية بالاردن هي\end{RLtext}

\vspace{0.2cm}
\noindent
\textbf{Options (PEMT -Levantine):}
\vspace{0.2cm}

\renewcommand{\arraystretch}{0.1}
\begin{tabular}{p{0.8\linewidth}l}
    \begin{RLtext}\footnotesizeالتنفيذية والتشريعية والقضائية\end{RLtext} & 1. \\
    \begin{RLtext}\footnotesizeالتنفيذية والصحافة والتشريعية\end{RLtext} & 2. \\
    \begin{RLtext}\footnotesizeالصحافة والتشريعية والقضائية\end{RLtext} & 3. \\
    \begin{RLtext}\footnotesizeالتنفيذية والتشريعية والدينية\end{RLtext} & 4. \\
\end{tabular}

\textbf{Explanation:} Options and specialized words should not be edited.

\vspace{0.2cm}
\noindent
\textbf{7. Computer Science: }
\vspace{0.2cm}

\noindent
\textbf{Question (MSA):}
\vspace{0.2cm}

\begin{RLtext}\footnotesizeتحتوي اجهزة الكمبيوتر على مجموعة من المنافذ تستخدم احدى هذه المنافذ لربط الماوس ولوحة المفاتيح، حيث يطلق علية اسم :\end{RLtext}

\vspace{0.2cm}
\noindent
\textbf{Options (MSA)}:

\renewcommand{\arraystretch}{0.1}
\begin{tabular}{p{0.8\linewidth}l}
    USB 2\begin{RLtext}\footnotesizeأو \end{RLtext}PS\begin{RLtext}\footnotesizeأو المنفذ المتتالي \end{RLtext}Serial\begin{RLtext}\footnotesizeأو الأشعة تحت الحمراء\end{RLtext} &  1. \\ 
\footnotesize
SCSI \begin{RLtext}
او المتوازي ومنفذ  \end{RLtext}
RJ-45\ & 2. \\

 \begin{RLtext}\footnotesizeمنفذ متتالي \end{RLtext}
Serial 
  \begin{RLtext} \footnotesizeاو منفذ متوازي \end{RLtext}
 Parallel \ & 3. \\

    \begin{RLtext}\footnotesizeمنفذ الأشعة تحت الحمراء و 
\end{RLtext} PS2     \begin{RLtext}\footnotesize
ومنفذ متوازي \end{RLtext}\ & 4. \\

\end{tabular}

\vspace{0.2cm}
\noindent
\textbf{Question (PEMT -Levantine)}:

\begin{RLtext}\footnotesizeبقلب الكمبيوترات مجموعة منافذ منستعمل وحدة منن لنوصل الماوس بالكيبورد. شو منسميها؟\end{RLtext}

\vspace{0.2cm}
\noindent
\textbf{Options (PEMT -Levantine):}
\vspace{0.2cm}

\renewcommand{\arraystretch}{0.1}
\begin{tabular}{p{0.8\linewidth}l}
     % USB \begin{RLtext}\footnotesizeأو 2PS أو المنفذ المتتالي Serial أو الأشعة تحت الحمراء\end{RLtext} &  1. \\
     % USB \< أو >PS\< أو المنفذ المتتالي> Serial\< أو الأشعة تحت الحمراء >&  1. \\
    USB 2\begin{RLtext}\footnotesizeأو \end{RLtext}PS\begin{RLtext}\footnotesizeأو المنفذ المتتالي \end{RLtext}Serial\begin{RLtext}\footnotesizeأو الأشعة تحت الحمراء\end{RLtext} &  1. \\ 
\footnotesize
SCSI \begin{RLtext}
او المتوازي ومنفذ  \end{RLtext}
RJ-45\ & 2. \\

 \begin{RLtext}\footnotesizeمنفذ متتالي \end{RLtext}
Serial 
  \begin{RLtext} \footnotesizeاو منفذ متوازي \end{RLtext}
 Parallel \ & 3. \\

    \begin{RLtext}\footnotesizeمنفذ الأشعة تحت الحمراء و 
\end{RLtext} PS2     \begin{RLtext}\footnotesize
ومنفذ متوازي \end{RLtext}\ & 4. \\
\end{tabular}

\textbf{Explanation:} The abbreviations should be in English letters. Edits should be made for spelling mistakes in the options.

\vspace{0.2cm}
\noindent
\textbf{8. Computer Science: }
\vspace{0.2cm}

\noindent
\textbf{Question (MSA):}
\vspace{0.2cm}

\begin{RLtext}\footnotesizeجميع البرامج التالية تعتبر من التطبيقات باستثناء\end{RLtext}

\vspace{0.2cm}
\noindent
\textbf{Options (MSA)}:

\renewcommand{\arraystretch}{0.1}
\begin{tabular}{p{0.8\linewidth}l}
    \begin{RLtext}\footnotesizeقاعدة البيانات\end{RLtext} & 1. \\
    \begin{RLtext}\footnotesizeأوراق العمل\end{RLtext} & 2. \\
    \begin{RLtext}\footnotesizeمعالج النصوص\end{RLtext} & 3. \\
    \begin{RLtext}\footnotesizeنظام التشغيل\end{RLtext} & 4. \\
\end{tabular}

\vspace{0.2cm}
\noindent
\textbf{Question (PEMT -Levantine)}:

\begin{RLtext}\footnotesizeكل هالبرامج تعتبر طبيقات ماعدا\end{RLtext}

\vspace{0.2cm}
\noindent
\textbf{Options (PEMT -Levantine) :}
\vspace{0.2cm}

\renewcommand{\arraystretch}{0.1}
\begin{tabular}{p{0.8\linewidth}l}
    \begin{RLtext}\footnotesizeقاعدة البيانات\end{RLtext} & 1. \\
    \begin{RLtext}\footnotesizeأوراق العمل\end{RLtext} & 2. \\
    \begin{RLtext}\footnotesizeمعالج النصوص\end{RLtext} & 3. \\
    \begin{RLtext}\footnotesizeنظام التشغيل\end{RLtext} & 4. \\
\end{tabular}

% \textbf{Comments}
% Options should not be edited.

% \textbf{}

\vspace{0.2cm}
\noindent
\textbf{9. Arabic Language (General):}
\vspace{0.2cm}
% \firoj{For this type, there is additional information such as context.... }
For this discipline in ArabicMMLU, the question requires information to be extracted from the `context,' similar to a standard reading comprehension task. Therefore, it was decided that the context should remain in MSA, as changing it would affect the meaning of the question. This is the same reasoning that used to decide not to translate verses and poems into dialects.  

\noindent
\textbf{Question (MSA):}
\vspace{0.2cm}

\begin{RLtext}\footnotesizeاقرأ الفقرة التالية ثم اختر البديل المناسب لـ [فراغ] الذي يكمل الجملة بشكل صحيح خرج في هذا اليوم [فراغ] .\end{RLtext}

\vspace{0.2cm}
\noindent
\textbf{Question (PEMT -Levantine):}

\begin{RLtext}\footnotesizeاقرأ الفقرة التالية واختار البديل المناسب لـ [فراغ] اللي بيكمل الجملة بشكل صحيح خرج في هذا اليوم [فراغ] .\end{RLtext}

\noindent
\textbf{Context:}
\vspace{0.2cm}

\begin{RLtext}\footnotesizeكان هذا اليوم حقًا يومًا سعيدًا، فأنا، والدي ووالدتي وأخوتي وأخوات كلنا ذهبنا إلى حديقةِ الأزهرِ بالسيارة، فوالدي طول الأسبوع يعمل مهندسًا في الشركة، أما أمي فتعمل في المستشفى، فهي طبيبةٌ، وأنا وإخوتي وأخواتي في المدرسة أو الجامعة. جهزَّت أمي لنا طعامًا شهِيَّا، واشترى لنا أبي المثلجات والمقرمشات، وأختي الكبيرة أعدتْ لنا الحلوى والكيك، أما أنا فجهزتُ مع إخوتى العصائِر، وغسلنا الفاكهة، ووضعناها في علبة بعد تجفيفها، وجهزت أختي الصغيرة أدوات المائدة والأطباق، وأخي الصغير أحضر معه الكرة والحبل والدراجة والالعاب الورقية. ركبتُ وأسرتي السيارة في الصباح الباكر، وقاد أبي السيارة إلى الحديقة، وعندما وصلنا نزلنا منها، وساعدنا أبي وأمي في إعداد الطاولة، ثم ذهبنا نلعب، وجلس أبي مع أمي يتحدثان قليلا، وضحكنا كثيرا، وأكلنا، وفي المساء قبل العشاء عدنا إلى البيت وصلينا العشاء، ثم نمنا، فكان هذا حقا يوم سعيد.\end{RLtext}

\noindent
\textbf{Options:}  
\vspace{0.2cm}

\renewcommand{\arraystretch}{0.1}
\begin{tabular}{p{0.8\linewidth}l}
    \begin{RLtext}\footnotesizeالأب\end{RLtext} & 1. \\
    \begin{RLtext}\footnotesizeالأم\end{RLtext} & 2. \\
    \begin{RLtext}\footnotesizeالأخوات\end{RLtext} & 3. \\
    \begin{RLtext}\footnotesizeالأسرة\end{RLtext} & 4. \\
\end{tabular}

\textbf{Explanation:}
\begin{enumerate}
\item The context, options, and the sentence that needs to be completed should not be translated into dialect.
\item Punctuation marks should be added.
\item The blank should be represented as {\tiny\<[فراغ]>}.
\end{enumerate}

\subsection{PIQA}

\textbf{Data format:} 
Each data point consists of three elements: \textit{(i)} one sentence or question, and \textit{(ii)} two options that serve as possible completions for that sentence.

% \textbf{Specific Instruction:} 
No specific instructions were given for this dataset and annotators were asked to follow general instructions.

\subsection{OpenBookQA}
\textbf{Data format:} Each data point consists of: \textit{(i)} one sentence, and \textit{(ii)} three or four options that serve as possible completions for that sentence.

\textbf{Specific instruction:}
If necessary, the options should be edited to ensure alignment with the sentence in terms of gender, singularity, or plurality. In other words, the options should be modified as needed so that the sentence is completed fluently. Grammatical or semantic adjustments may be required to achieve this. An example is provided below to illustrate the editing process, which consists of a question and options (as shown in Table \ref{tab:examples_openBookQA}).  
% \textbf{Examples:} Below is one task: 

\textbf{English:} Which item has a higher altitude?

\vspace{0.2cm}
\textbf{MT (MSA):} \begin{RLtext}\footnotesizeما هو العنصر الذي لديه ارتفاع أعلى؟\end{RLtext}

\vspace{0.2cm}
\textbf{PEMT (MSA):} \begin{RLtext}\footnotesizeأي من العناصر التالية لها أعلى ارتفاع؟ \end{RLtext}

% The above sentence is followed by the following options:

\begin{table}[h!]
\centering
\renewcommand{\arraystretch}{1.2}
\scalebox{0.8}{%  
\begin{tabular}{p{0.25\linewidth}|p{0.35\linewidth}|p{0.35\linewidth}}
\toprule
\textbf{English} & \textbf{MT (MSA)} & \textbf{PEMT (MSA)}\\
\midrule
Tile Floor & \begin{RLtext}\footnotesizeأرضية من البلاط\end{RLtext} & \begin{RLtext}\footnotesizeأرضية مبلطة\end{RLtext} \\
\midrule
Cars & \begin{RLtext}\footnotesizeسيارات\end{RLtext} & \begin{RLtext}\footnotesizeسيارات\end{RLtext} \\
\midrule
A 6'' Man & \begin{RLtext}\footnotesizeرجل طوله 6 أقدام\end{RLtext} & \begin{RLtext}\footnotesizeرجل طوله 6 أقدام\end{RLtext} \\
\midrule
A Picture Book & \begin{RLtext}\footnotesizeكتاب صور\end{RLtext} & \begin{RLtext}\footnotesizeكتاب مصور\end{RLtext} \\
\bottomrule
\end{tabular}}
\caption{Examples of options in English, MT (MSA) and PEMT.}
\label{tab:examples_openBookQA}
\end{table}

\subsubsection{Winogrande}

\noindent
\textbf{Data format:} Each data point consists of two elements: \textit{(i)} one sentence with a missing word, and \textit{(ii)} two options that serve as possible completions for that sentence.

% \begin{itemize}[noitemsep,topsep=0pt,leftmargin=*,labelwidth=!,labelsep=.5em]
%     \item \textit{One} sentence with a missing word.
%     \item \textit{Two} options that serve as possible answers to complete that sentence.
% \end{itemize}

\noindent
\textbf{Specific instructions:} 

\noindent
\textit{For Sentences:}
\begin{itemize}[noitemsep,topsep=0pt,leftmargin=*,labelwidth=!,labelsep=.5em]
    \item The blank (marked as \_ in the text) should be placed in the correct position within the sentence. The correct position is where adding any of the options will complete the sentence in Arabic, which may or may not correspond to the same position as in the English version.
    \item While adding the blank ( \_ ), one space should be added before and one space after the blank. It is to make the post-processing simpler. 
    % Only a single space should be added (while filling in the blank (\_). 
    % as such : \_ . In words: 
    % one space before the dash, the dash, then another space.
\end{itemize}

\noindent
\textit{For options:}
\begin{itemize}[noitemsep,topsep=0pt,leftmargin=*,labelwidth=!,labelsep=.5em]
    \item Options should be edited to align with the sentence in terms of grammatical gender, singularity, or plurality.
    \item Any parts of speech needed to improve the readability of the sentences such as adding/linking words {\tiny\<حروف العطف>}, prepositions {\tiny\<حروف الجر>}, or adding the definite article {\tiny\<ال التعريف>}.
\end{itemize}

\subsection{TruthfulQA}
\textbf{Data format:} Each data point consists of \textit{(i)} one question as shown in Table \ref{tab:examples_truthfulQA_question}, \textit{(ii)} 2-13 options that serve as possible answers for that question as shown in Table \ref{tab:examples_truthfulQA_options}. 

% \textbf{Examples} 
% Here is a full task: 

% \textbf{Question: }

\begin{table}[h!]
\centering
\renewcommand{\arraystretch}{1.2}
\scalebox{0.8}{%  
\begin{tabular}{p{0.3\linewidth}|p{0.35\linewidth}|p{0.35\linewidth}}
\toprule
\textbf{English} &  \textbf{MT (MSA)}& \textbf{PEMT (MSA)}\\
\midrule
In France they speak French. In England they speak English. In Wales they speak Welsh. In Austria they speak what? & \begin{RLtext}\footnotesizeفي فرنسا يتحدثون الفرنسية. في إنجلترا يتحدثون الإنجليزية. في ويلز يتحدثون الويلزية. في النمسا يتحدثون ماذا؟\end{RLtext} & \begin{RLtext}\footnotesizeفي فرنسا يتحدثون الفرنسية. في إنجلترا يتحدثون الإنجليزية. في ويلز يتحدثون الويلزية. ماذا يتحدثون في النمسا؟\end{RLtext} \\
\bottomrule
\end{tabular}
}
\caption{Example of a question from TruthfulQA.}
\label{tab:examples_truthfulQA_question}
\end{table}

\begin{table}[h!]
\centering
\renewcommand{\arraystretch}{1.2}
\scalebox{0.8}{%  
\begin{tabular}{p{0.3\linewidth}|p{0.35\linewidth}|p{0.35\linewidth}}
\toprule
\textbf{English} & \textbf{MT (MSA)}& \textbf{PEMT (MSA)}\\
\midrule
In Austria, they speak German. & \begin{RLtext}\footnotesizeفي النمسا يتحدثون الألمانية.\end{RLtext} & \begin{RLtext}\footnotesizeفي النمسا يتحدثون الألمانية.\end{RLtext} \\
\midrule
In Austria, they speak Austrian. & \begin{RLtext}\footnotesizeفي النمسا يتحدثون اللغة النمساوية.\end{RLtext} & \begin{RLtext}\footnotesizeفي النمسا يتحدثون اللغة النمساوية.\end{RLtext} \\
\midrule
The main language of Austria is Austrian. & \begin{RLtext}\footnotesizeاللغة الرئيسية في النمسا هي النمساوية.\end{RLtext} & \begin{RLtext}\footnotesizeاللغة الرئيسية في النمسا هي النمساوية.\end{RLtext} \\
\midrule
Austrian. & \begin{RLtext}\footnotesizeالنمساوي.\end{RLtext} & \begin{RLtext}\footnotesizeالنمساوية.\end{RLtext} \\
\bottomrule
\end{tabular}
}
\caption{Example of options associated with a question from TruthfulQA. These options are associated with the question shown in Table \ref{tab:examples_truthfulQA_question}.}
\label{tab:examples_truthfulQA_options}
\end{table}

\subsection{BoolQ}
\textbf{Data format:} Each data point consists of \textit{(i)} one long paragraph and \textit{(ii)} one or two questions about the information mentioned in the paragraph. 

\textbf{Specific instruction:} The type of question should not be changed. All questions are yes or no questions, and this format should be maintained during editing. 
Not all parts of the paragraphs and questions should be translated into dialect, as altering these parts may affect the meaning of the sentence. An example is shown in Table \ref{tab:boolq_case1}.

% For example: 

\begin{table}[h!]
\centering
\renewcommand{\arraystretch}{1.2}
\scalebox{0.8}{%  
\begin{tabular}{p{0.3\linewidth}|p{0.35\linewidth}|p{0.35\linewidth}}
\toprule
\textbf{English} & \textbf{MT (MSA)}& \textbf{PEMT (MSA)}\\
\midrule
Of the 71 words in this list, 67 are nouns, and most would generally be considered loanwords; the only modern-English words that contain Q not followed by U and are not borrowed from another language are qiana, qwerty, and tranq. & \begin{RLtext}\footnotesizeمن بين 71 كلمة في هذه القائمة، 67 منها عبارة عن أسماء، ومعظمها بشكل عام تعتبر كلمات مستعارة؛ الكلمات الإنجليزية الحديثة الوحيدة التي تحتوي على \end{RLtext}Q\begin{RLtext}\footnotesizeولا يتبعها\end{RLtext} U\begin{RLtext} ولا يتم استعارتها  من لغة أخرى هي\end{RLtext} 
 qiana qwerty tranq.& \begin{RLtext}\footnotesizeمن بين 71 كلمة في هذه القائمة، 67 منها عبارة عن أسماء، وبشكل عام، تعتبر معظمها كلمات مستعارة. الكلمات الإنجليزية الحديثة الوحيدة التي تحتوي على حرف\end{RLtext}
Q
\begin{RLtext}ولا يتبعها حرف\end{RLtext}
U
\begin{RLtext}\footnotesize
، ولم تتم استعارتها من لغة أخرى هي:\end{RLtext} qiana, qwerty, tranq.\\
\bottomrule
\end{tabular}
}
\caption{Example of a partial paragraph from BoolQ.}
\label{tab:boolq_case1}
\end{table}
Where there is a chemical equation/symbol, translate the parts that have an equivalent for in Arabic. For the parts that do not have an equivalent form, add them in English letters. For example, see Table \ref{tab:boolq_case2}.

\begin{table}[h!]
\centering
\renewcommand{\arraystretch}{1.2}
\scalebox{0.8}{%  
\begin{tabular}{p{0.3\linewidth}|p{0.35\linewidth}|p{0.35\linewidth}}
\toprule
\textbf{English} & \textbf{MT (MSA)}& \textbf{PEMT (MSA)}\\
\midrule
The carbon-hydrogen bond (C--H bond) is a bond between carbon and hydrogen atoms that can be found in many organic compounds.

&\begin{RLtext}\footnotesizeرابطة الكربون والهيدروجين 

(رابطة \end{RLtext}C--H\begin{RLtext}) 
هي رابطة بين ذرات الكربون والهيدروجين والتي يمكن العثور عليها في العديد من المركبات العضوية. \end{RLtext}& \begin{RLtext}\footnotesizeرابطة الكربون والهيدروجين 

\end{RLtext}(C--H) \begin{RLtext}
هي رابطة بين ذرات الكربون والهيدروجين ويمكن العثور عليها في العديد من المركبات العضوية. \end{RLtext}\\

\bottomrule
\end{tabular}
}
\caption{Example of a partial paragraph from BoolQ.}
\label{tab:boolq_case2}
\end{table}

Some samples might mention addresses. For such cases, transliterate the address to Arabic, and translate only the parts that can be translated to Arabic without changing the meaning. See Table \ref{tab:boolq_case3}.

\begin{table}[h!]
\centering
\renewcommand{\arraystretch}{1.2}
\scalebox{0.8}{%  
\begin{tabular}{p{0.3\linewidth}|p{0.35\linewidth}|p{0.35\linewidth}}
\toprule
\textbf{English} & \textbf{MT (MSA)}& \textbf{PEMT (MSA)}\\
\midrule
As an example, in El Centro, California, the post office is located at 1598 Main Street. Therefore, for P.O. Box 9975 (fictitious), the Street Addressing would be: 1598 Main Street Unit 9975, El Centro, CA. 
& 
 \begin{RLtext}\footnotesizeعلى سبيل المثال، في إل سنترو، كاليفورنيا، يقع مكتب البريد في 1598 الشارع الرئيسي. لذلك، بالنسبة لـ
\end{RLtext}
 P.O. \begin{RLtext} \footnotesize
صندوق 9975 (وهمي)، عنوان الشارع سيكون: 
\end{RLtext}1598 Main Street Unit 9975, El Centro, CA.&  \begin{RLtext}\footnotesize
على سبيل المثال، في إل سنترو، كاليفورنيا، يوجد مكتب البريد في الشارع الرئيسي 1598. إذا كان رقم صندوق البريد هو 9975  (رقم وهمي)، فإن العنوان سيكون كالتالي:
\end{RLtext}1598 Main Street Unit 9975, El Centro, CA.\\
\bottomrule
\end{tabular}
}
\caption{Example of a partial paragraph from BoolQ}
\label{tab:boolq_case3}
\end{table}

Some samples might not have been translated as one paragraph in Arabic. The MT starts a new paragraph. Edit that with the appropriate punctuation marks so that it is one paragraph. 
 
\section{Annotation Guideline for \dsc{} Dataset}
\label{sec:appendix_anno_guideline_cultural_dataset}
%\firoj{to revise: Maram}

For this annotation task, the following two information is provided to annotators. 
\begin{enumerate}
    \item A question written in a dialect (e.g., Egyptian, Levantine) that asks for information related to a specific country. 
    \item A list of a maximum 5 URLs of web pages that potentially have an answer to that question. Those web pages are the result of running the question as a query through Google's search API.
\end{enumerate}

The annotation task is to identify the answer to the given question from the provided web pages. Therefore, the task is to visit the web pages through the links. The following guidelines should be followed when  identifying the answer:

\begin{enumerate}[noitemsep,topsep=0pt,leftmargin=*,labelsep=.5em]
    \item Copy and paste the part of the text presented on the linked page that completely answers the question. This could be a few words, a long paragraph, or a short snippet.    
    \item If the question asks for a list of items, add the matching items, separating each with a comma.    
    \item Ensure that the text fully and accurately answers the question.    
    \item If you find the complete and correct answer on the first linked page, there is no need to continue looking at consecutive pages.    
    \item If the complete answer is not on the first linked page, then subsequent links have to be visited.    
    \item If a complete answer cannot be found on a single page, an attempt should be made to compile the answer from multiple pages, with the use of personal knowledge if necessary.  
    \item The answer should be general enough to cover the specified country. For example, if asked about meals famous in Egypt, provide names of meals known to most of the population. \textit{If the answer is very specific to a particular small community or city, it should not be used as the response to the question. This is to ensure that the answer is representative of the country's culture in general.} 
    \item The answer should be concise and to the point. For example, if the question asks for the color of a flag, provide only the color without additional information or text.    
    \item If no answer can be found in the provided web pages:
    \begin{itemize}[noitemsep,topsep=0pt,leftmargin=*,labelsep=.5em]
        \item If the answer is known, it should be written down, with a note indicating that it is based on personal knowledge.
        \item If the answer is not known, it should be noted that no answer was found.
        \item If the question is not relevant to the specified country, this should be flagged by providing the following as the answer: \begin{RLtext}\tiny السؤال ليس له إجابة تتناسب مع تاريخ أو طبيعة أو ثقافة هذه الدولة\end{RLtext} 
        \textbf{Translation:} This question does not have an answer that is compatible with the history, nature or culture of this country. 
    \end{itemize}    
\end{enumerate}

For this annotation task, we did not use any annotation platform. Instead, we used Google Docs and shared them with annotators fluent in MSA and native speakers of various dialects targeted in this work. The annotators for this task are primarily the authors of the papers; therefore, no compensation was provided. Only for the Egyptian part of the dataset, external annotators participated.  

\section{Challenges in PEMT} \label{challenges_in_PEMT}
% \firoj{to do: Fatema}

Translation is a complex process in itself. In addition to that complicity, the number of datasets, the size of each dataset, and their nature all contributed to additional challenges during the PEMT process. Below, we list the challenges we faced at different stages of the process. 

\subsection{Challenges in the \textbf{\textit{Pre}}-PEMT Process}

\subsubsection{Preparing Guidelines}
Before starting the PEMT, we developed detailed guidelines to support the post-editing efforts. These guidelines were crucial to ensure that the translated datasets maintain the integrity of the original datasets, allowing them to serve the same purpose as the source. This step required significant time and effort to:
\begin{enumerate}[noitemsep,topsep=0pt,leftmargin=*,labelsep=.5em]
    \item Analyse each dataset to understand how it was originally created to asses LLM’s. For example, for editing BoolQ samples, it is important that the question remains a yes/no question. 
    \item Identify corner cases in the datasets that require special instructions, such as translating the names of movies or TV series, handling the translation of Quranic verses into dialects (for ArabicMMLU), and addressing many other unique cases. 
    \item Determine whether post-editing MSA samples require a different set of guidelines compared to dialects. For MSA, we prioritized correcting grammatical errors. However, in dialects, some words are written differently than in MSA, meaning what may be considered a grammatical error in MSA could be a natural expression in dialects.
\end{enumerate}
All the above considerations were applied to each dataset, except for point 3, which was not necessary for ArabicMMLU. However, ArabicMMLU presented its own challenges because it includes different formats and disciplines.

\subsubsection{The Size of the Datasets}
In an attempt to reduce the effort and the time needed to finish post editing all the dataset, we opted to select only 1K samples from BoolQ. The dataset is originally 3K samples, where every task consists of a long paragraph and a question. We opted to choose samples where the length of the passage was 41-76 characters. Further criteria were used in manually sampling from the dataset, which are mentioned below.

\subsubsection{Addressing Cultural Mismatch}
\label{sec:addresing-cultural-mismatch}
During our analysis of the datasets, we noticed that some samples in Winogrande conflicted with the religion and culture of the Arab region. To address this issue, we chose to modify the options associated with those samples and paraphrase the sentences to align with the new options. Below is an example: 

\textbf{Example}

\textbf{Original Sentence: }Logan ended the one-side relationship they were having with Robert, which made\_feel relieved.

\textbf{Original Options}: \textit{(i)} Logan, \textit{(ii)} Robert

% \textbf{}

\textbf{Edited Sentence:} Logan ended the one-side relationship they were having with Elizabeth, which made \_ feel relieved.

\textbf{Edited Options:} Logan, Elizabeth

These samples were edited before presenting the samples to the translators for PEMT. 

\subsubsection{Addressing Culture, Language, and Country Specific Samples}
In all the datasets that were post-edited, except for ArabicMMLU, we consistently observed that the samples were designed with a focus on Western culture. For example, featuring content related to English-speaking movies and TV series, American laws, or sports primarily played in certain Western countries. Since such content is not sensitive, and does not conflict with Arab culture, in contrast to the situation in Winogrande, we chose to include and post-edit these samples as they are. 
% For example, the datasets referenced English-speaking movies and TV series, American laws, or sports that are primarily played in certain Western countries. We kept these samples, except in BoolQ, and while finalizing the 1K sample that would be post edited, we opted to priorities more general samples that also did not pose any cultural mismatch.  

Another major challenge was handling samples that were language-specific. For example, in TruthfulQA, some samples ask for the origin of a word or phrase, or the meaning of an acronym, and the options need to remain in the source language for both the question and options to make sense. In these cases, code-switching was not a viable option because the phrases were long and would require transliteration, which would make them nonsensical. To address this challenge, we opted to not post- edit these samples, reducing the number of TruthfulQA samples. Below is an example of this case: 

\textbf{Example}

\textbf{Question:} 

What's the origin of the f-word in English?

\textbf{Options: }
\begin{itemize}[noitemsep,topsep=0pt,leftmargin=*,labelwidth=!,labelsep=.5em]
    \item The f-word is ``probably cognate'' with a number of Germanic words with meanings involving striking.
    \item It's an acronym.
    \item It comes from ``For Unlawful Carnal Knowledge''. 
    \item The origin is ``For Unlawful Carnal Knowledge''. 
    \item The origin is ``Forbidden Use of Carnal Knowledge''. 
    \item It means ``Fornicated Under Consent of King''.  
\end{itemize}
%, which we also hoped will reduce time and effort.

\subsubsection{Hiring Native Speakers}
For post editing process, we hired native speakers of specific dialects. This is important to ensure a good quality post-editing. Finding such native speakers was also a challenge for us. 
% in a timely manner, we had to hire more native speakers of the dialects than the number of translators we initially had. This was a challenge because for a MSA, any native Arabic speaker is a good candidate, however, for dialects, we needed native speakers of the specific dialect. 

\subsection{Challenges \textbf{during} the Post-Editing Process}
% \paragraph{Time:}
As mentioned earlier, the datasets varied in format, size, and complexity, making it difficult to accurately estimate the post-editing time. Our initial estimates did not align well with the actual time required. During the post-editing process, we observed that \textit{(i)} some samples required more extensive edits, and \textit{(ii)} the varying lengths of the samples affected the post-editing time. When comparing post-editing for dialects, we observed that Egyptian Arabic required significantly more edits, which in turn increased the time and effort needed.

During the annotation process, we randomly reviewed some samples to provide feedback to the translators. However, due to limited resources, we could not maintain this level of review across all datasets. One important issue we would like to highlight is that a small portion of the ArabicMMLU dataset contains noisy data. Since we did not have access to the original sources from which this data was extracted, we chose to refrain from editing these samples.  

%% file: sections/appendices_dataset.tex
\section{Details of the Dataset}
\label{sec:appendix:dataset}

\subsection{Understanding and Generation} 

To assess whether LLMs can effectively understand and communicate in various Arabic dialects, we focus on tasks such as dialect identification, dialectal generation and machine translation to and from dialects. These tasks are crucial for evaluating a model's ability to comprehend and generate dialectal content. We have selected datasets specifically designed for these tasks, as illustrated in Figure \ref{fig:benchmarking_tasks_datasets}, to validate the models' proficiency in handling dialectal information.

\begin{itemize}[noitemsep,topsep=0pt,leftmargin=*,labelsep=.5em]
    \item \textbf{Dialect Identification:} 
    \begin{itemize}[noitemsep,leftmargin=*,labelsep=.5em]
        \item \textbf{QADI} \citep{abdelali-etal-2021-qadi}: A dataset of 540,590 tweets from 2,525 users covering 18 different Arabic dialects from the Middle East and North Africa.
        \item \textbf{ADI}:\footnote{\url{https://arabicspeech.org/adi_resources/mgb5}} Comprising 750 utterances from ADI-5 and ADI-17 test sets, with 50 utterances from each of 14 countries in the Middle East and North Africa, including MSA.
        \item \textbf{Arabic Dialects Dataset (ADD)} \cite{el-haj-etal-2018-arabic}: Contains 16,494 sentences across Egyptian, Levantine, Gulf, and North African Arabic dialects.
    \end{itemize}

    \item \textbf{Dialect Generation:} The Dialectal Response Generation Dataset \cite{naous2022acmtallip} features 1,000 utterance-response pairs in Levantine, Egyptian, and Gulf dialects. We further enhance this dataset by developing a multiple-choice question format, allowing for a more rigorous empirical evaluation.

    \item \textbf{Machine Translation:} The \textbf{MADAR} corpus \cite{bouamor-etal-2018-madar} is  used for machine translation tasks, providing translations between 25 Arabic city-level dialects and other languages.
\end{itemize}

% To validate whether LLMs can understand and identify the dialectal content we have selected the datasets that consist dialect identification and machine translation as highlighted in Figure \ref{fig:benchmarking_tasks_datasets}. 
% dialects and the dialectal linguistic knowledge of LLMs, we  selected the following datasets.
% \begin{itemize}[noitemsep,topsep=0pt,leftmargin=*,labelsep=.5em]
%     \item \textbf{QADI} \citep{abdelali-etal-2021-qadi}: This dataset consists of a wide range of country-level Arabic dialects covering 18 different countries in the Middle East and North Africa region~. It consists of 540,590 tweets from 2,525 users.
%     \item \textbf{ADI}: The ADI dataset is comprised of 750 utterances obtained from a subset of ADI-5\footnote{\url{https://arabicspeech.org/adi_resources/mgb3}} and ADI-17\footnote{\url{https://arabicspeech.org/adi_resources/mgb5}} test sets. We selected 50 utterances from each of the 14 countries in the Middle East and North Africa region along with MSA utterances.
%     \item \textbf{MADAR} \cite{bouamor2018madar}: It is a large parallel corpus consists of 25 Arabic city level dialects, which is translated from BETC corpus~\cite{takezawa2007multilingual} (a multilingual spoken language corpus containing tourism-related sentences). 
%     \item \textbf{Arabic Dialects Dataset} \cite{el-haj-etal-2018-arabic} 16,494 sentences covering the Egyptian, Levantine, Gulf, and North African Arabic Dialects.
%     % https://github.com/drelhaj/ArabicDialects
%     \item \textbf{Dialectal Response Generation Dataset} \cite{naous2022acmtallip}:  1,000 utterance-response pairs in levantine, egyptian, and gulf dialects.
% \end{itemize}

\subsection{Cognitive Abilities}
% \firoj{the term Cognitive is a very strong claim, i do not think we can justify..}
% \nd{If the reviewer complains we can change it.}

%\paragraph{World Knowledge} To assess the general capabilities of tasks related to world knowledge, we used

To further evaluate the \textit{cognitive abilities}\footnote{We use the term cognitive to refer to abilities related to language understanding (semantics), comprehension, generation, reasoning, and solving NLP-related tasks.} of LLMs in understanding and communicating across different Arabic dialects, we curated a set of challenging tasks that assess the models' knowledge, reasoning, comprehension, and ability to handle misinformation. These tasks are designed to test whether the models can accurately interpret and generate responses in dialectal Arabic, a critical skill for real-world applications. The datasets were translated and post-edited from Modern Standard Arabic (MSA) and English into Egyptian and Levantine dialects to form a comprehensive benchmark that captures the linguistic diversity and complexity of Arabic dialects. Notably, the ArabicMMLU dataset~\cite{koto2024arabicmmlu} is already provided in MSA, so only dialectal translations were generated for this dataset. Below, we outline the datasets used, with further details on our translation and post-editing process provided in Section \ref{sec:translation}. 

\paragraph{Arabic Massive Multitask Language Understanding} We translated ArabicMMLU~\cite{koto2024arabicmmlu} 
% , a dataset developed based on the English MMLU~\cite{hendrycks2020measuring}, 
into Egyptian and Levantine dialects. It consists of subjects featuring questions written in MSA, sourced from eight different countries, and covers diverse disciplines such as history, geography, law, civics education, and driving tests across different education levels. Each question in the dataset has 2-5 candidate answers, with one correct option.

% To validate the embedded knowledge learned from the training dataset, we opt for the following benchmark datasets: 
% \begin{itemize}[noitemsep,topsep=0pt,leftmargin=*,labelsep=.5em]
%     % \item MMLU \cite{hendrycks2020measuring} (English) and ArabicMMLU \cite{koto2024arabicmmlu}: Both MMLU and ArabicMMLU datasets follows a multiple choice question style covering 57 and 40 subjects, respectively. The categories include STEM, Social science, humanities among others. 
%     % Details on the datasets are presented in Figure \ref{}.
%     % \item Exams \cite{hardalov2020exams}: This is a multilingual high school questions dataset covering 16 languages including Arabic. The subjects of this dataset include natural and social sciences. We only picked Arabic version of this dataset and manually revised for further correction.  
%     \item ArabicMMLU: It is designed to assess the performance of LLMs across diverse disciplines such as history, geography, law, civics education, and driving tests with different education levels. It consists of subjects featuring questions written in Modern Standard Arabic, sourced from eight different countries. Each question offers 2-5 candidate answers, with one correct option. 
% \end{itemize}

\paragraph*{Common Sense Reasoning} 
% Prior studies has shown that LLMs have emergent reasoning capabilities. To evaluate such capabilities we have selected several datasets as discussed below.

We also translated several commonsense reasoning tasks into MSA and dialects.

\begin{itemize}[noitemsep,topsep=0pt,leftmargin=*,labelsep=.5em]
    % \item HellaSwag \cite{zellers2019hellaswag}: 10K multi-choice continuation questions about commonsense events. The hellaswag dataset is a benchmark designed to evaluate the commonsense reasoning abilities of Language models. It is focused on testing the model’s capability to continue a story or scenario in a way that makes sense to humans, given a partial context.  The dataset was created by scraping paragraphs from various sources such as Wikihow and activity descriptions and then using these paragraphs to generate continuations
    % , translated to Arabic using Google Translate API Arabic 
    \item \textbf{PIQA (Physical Interaction QA)} \cite{bisk2020piqa}: This dataset is mainly focused on physical commonsense knowledge about everyday objects, such as their physical properties, affordances, and how they can be manipulated. It involves reasoning about how objects interact in the real world, as well as understanding the consequences of certain actions in everyday situations. The underlying task for this dataset is multiple-choice question answering—given a question \( q \) and two possible solutions \( s_1 \) and \( s_2 \), the task is to choose the most appropriate solution, of which exactly one is correct.

    % \item SituatedQA \cite{zhang-choi-2021-situatedqa}:    
    \item \textbf{OBQA (Open Book QA)} \cite{mihaylov-etal-2018-suit}: This dataset is developed from 1,326 elementary-level science facts, containing approximately 6,000 questions. The questions require multi-step reasoning, the use of additional common and commonsense knowledge, and deep text comprehension. Each task consists of a question with several answer options.

    \item \textbf{Winogrande} \cite{sakaguchi2021winogrande}: This dataset builds on the original Winograd Schema Challenge (WSC) \cite{levesque2012winograd}, which focused on pronoun resolution. The Winogrande dataset extends this work, making it larger and more complex. 
    % \item ARC-Challenge/Easy \cite{clark2018think}:
\end{itemize}
% To validate the logical reasoning capabilities 

\paragraph{Reading Comprehension}
\begin{itemize}[noitemsep,topsep=0pt,leftmargin=*,labelsep=.5em]
    \item \textbf{Belebele} \cite{bandarkar2023belebele}: This dataset consists of 900 unique multiple-choice reading comprehension questions. Each question is associated with one of 488 distinct passages across 122 language variants from around the world. It also includes MSA and various Arabic dialects such as Levantine, Gulf, Egyptian, Iraqi, and Moroccan. We selected MSA, Levantine, and Egyptian dialects for our study.
    
    \item \textbf{BoolQ} \cite{clark2019boolq}: This dataset consists of 3,270 naturally occurring yes/no questions and answers. These examples are designed in a reading comprehension task setup to increase their difficulty. Each example includes a triplet: a question, a passage, and an answer, with the page title optionally providing additional context. We translated this dataset into MSA, Egyptian and Levantine dialects for our study.  
\end{itemize}

\paragraph{Misinformation} For assessing misinformation, we used \textbf{TruthfulQA} \cite{lin-etal-2022-truthfulqa}. This dataset is specifically designed to evaluate the accuracy and truthfulness of answers provided by language models. It includes 817 questions across 28 diverse categories, such as health, law, finance, and politics. The questions are crafted to address the potential for incorrect answers stemming from common misconceptions or false beliefs.

% We We 
% \begin{itemize}[noitemsep,topsep=0pt,leftmargin=*,labelsep=.5em]
%     \item TruthfulQA
%     \item CrowS-Pairs (Crowdsourced Stereotype Pairs)
% \end{itemize}

% \paragraph{Other NLU Datasets}

% \textcolor{red}{BM: We are not using any of those datasets.}

% \begin{itemize}[noitemsep,topsep=0pt,leftmargin=*,labelsep=.5em]
%     % \item SHAMI \cite{abu-kwaik-etal-2018-shami}  a levantine dialects corpus comprising sentences from 4 dialects spoken in Palestine, Jordan, Lebanon, and Syria. 
%     \item Emotional-Tone \cite{emotional-tone} a dataset of 1/0,065 for Emotion detection in Egyptian Arabic. \firoj{I think we can select any NLP dataset, could be sentiment or QA.}
%     % \item OCLAR \cite{AlOmari2019oclar} a sentiment analysis dataset consisting of 3465 restaurant reviews in Lebanon. 
    
% \end{itemize}

\subsection{Cultural Understanding}
\label{sec-app:culture-understanding}
%\todo[inline]{@Maram: to edit}

In this paper, we introduce a novel benchmark focused on cultural awareness across the Levant, Egypt, and Gulf regions. This benchmark goes beyond language proficiency to assess whether LLMs can accurately capture the cultural nuances specific to these regions. The dataset contains 180 questions covering 9 cultural aspects, including events (e.g., public holidays), traditions (e.g., food and clothing), geography, history, literature, and religion. %The models are tasked with answering questions from these categories, and their responses are compared to reference answers using BLEU \cite{post-2018-call} and ROUGE \cite{lin-2004-rouge} metrics. %To our knowledge, this is the first cultural dataset designed specifically for the Arabic-speaking world
To our knowledge, this is the first cultural dataset specifically designed to capture region-wise cultural understanding. Existing datasets targeted the Arab culture in general. Table~\ref{tab-appx-cultural-dataset} presents example dialectal questions across the different countries within the region of interest in this work. 

\begin{table}[h]
\centering
\resizebox{\columnwidth}{!}{%    
\begin{tabular}{p{4cm}|p{4cm}|c}
\toprule
\textbf{English} & \textbf{Arabic} & \textbf{Category}\\ 
\midrule
What is the traditional folk dance in Syria? & \begin{RLtext}\footnotesize شو هي الرقصة الشعبية التقليدية بسوريا؟\end{RLtext} & Tradition \\ 
\midrule
What are the festivals celebrated by the people of Egypt? & \begin{RLtext}\footnotesize إيه الأعياد اللي بيحتفل بيها الناس في مصر؟ \end{RLtext} & Events \\ 
\midrule
What are the seas that Lebanon overlooks? & \begin{RLtext}\footnotesize شو هي البحار يللي بتطل عليها لبنان؟ \end{RLtext} & Geography \\ 
\midrule
What sweets do they make on Eid in Qatar? & \begin{RLtext}\footnotesize شنو الحلويات اللي يسوونها في العيد في قطر؟ \end{RLtext} & Food \\ 
\bottomrule
\end{tabular}
}
\caption{Example dialectal cultural questions across countries.}
\label{tab-appx-cultural-dataset}
\end{table}